\documentclass[Journal, 10pt]{IEEEtran}
\usepackage{dsfont}

\usepackage{hyperref}
\usepackage{mathrsfs}

\usepackage[usenames]{color}
\usepackage{colortbl}
\usepackage[color,final]{showkeys}
\usepackage[table,xcdraw]{xcolor}
\usepackage{graphicx,dblfloatfix}
\usepackage{makecell}

\usepackage{amsmath,amssymb,amsthm}
\usepackage{latexsym,amsfonts,amscd,amsxtra,amstext}
\usepackage{algorithm,algorithmic}
\usepackage{url,cite}
\usepackage{hyperref}
\usepackage{graphicx}
\usepackage{epstopdf}
\usepackage{epsfig} 
\usepackage{threeparttable}
\usepackage{subfigure}

\usepackage{mathtools}
\usepackage[normalem]{ulem} 
\usepackage{enumerate}
\usepackage{amssymb}
\usepackage{booktabs}
\usepackage{multirow,color}
\usepackage{amsfonts,dsfont}
\usepackage{hyperref}
\usepackage{cite}
\usepackage{pifont}
\usepackage{cases}

\usepackage[color]{showkeys}
\definecolor{refkey}{gray}{0.8}
\definecolor{labelkey}{gray}{0.8}

\newcommand{\remove}[1]{{}}
\newcommand{\cut}[1]{}

\def\bE{\mathbb{E}}
\def\filt{\boldsymbol{\mathcal{F}}}

\newcommand{\bm}[1]{\mbox{\boldmath $#1$}}

\newcommand{\nn}{\nonumber}
\newcommand{\ba}{\left[ \begin{array}}
	\newcommand{\ea}{\\ \end{array} \right]}
\newcommand{\qd}{\hfill{$\blacksquare$}}
\newcommand{\define}{\;\stackrel{\Delta}{=}\;}

\newcommand{\Tr}{\mbox{\rm {\small Tr}}}
\newcommand{\eq}[1]{\begin{align}#1\end{align}}
\def\tran{^{\mathsf{T}}}

\def\bpsi       {{\boldsymbol \psi}}
\def\bphi       {{\boldsymbol \phi}}

\def\bgamma {{\boldsymbol \gamma}}

\def\H{{\boldsymbol{H}}}

\def\d{{\boldsymbol{d}}}

\def\h{{\boldsymbol{h}}}

\def\s{{\boldsymbol{s}}}

\def\u{{\boldsymbol{u}}}
\def\w{{\boldsymbol{w}}}
\def\x{{\boldsymbol{x}}}

\def\z{{\boldsymbol{z}}}

\newcommand{\sw}{{\scriptstyle{\mathcal{W}}}}

\newcommand{\swb}{{\boldsymbol{\scriptstyle{\mathcal{W}}}}}
\newcommand{\syb}{{\boldsymbol{\scriptstyle{\mathcal{Y}}}}}
\newcommand{\szb}{{\boldsymbol{\scriptstyle{\mathcal{Z}}}}}

\newcommand{\sy}{{\scriptstyle{\mathcal{Y}}}}

\newcommand{\sxb}{\boldsymbol{\scriptstyle{\mathcal{X}}}}

\newcommand{\tw}{\widetilde{\boldsymbol{w}}}

\newcommand{\twb}{\widetilde{\boldsymbol{\scriptstyle{\mathcal{W}}}}}
\newcommand{\tyb}{\widetilde{\boldsymbol{\scriptstyle{\mathcal{Y}}}}}

\newcommand{\tcA}{\overline{{\mathcal{A}}}}

\newcommand{\nnb}{\nonumber \\}

\newcommand{\bvec}{\mathrm{bvec}}

\newcommand{\cA}{{\mathcal{A}}}
\newcommand{\cB}{{\mathcal{B}}}
\newcommand{\cC}{{\mathcal{C}}}
\newcommand{\cD}{{\mathcal{D}}}

\newcommand{\cF}{{\mathcal{F}}}
\newcommand{\cG}{{\mathcal{G}}}
\newcommand{\cH}{{\mathcal{H}}}
\newcommand{\cI}{{\mathcal{I}}}
\newcommand{\cJ}{{\mathcal{J}}}

\newcommand{\cL}{{\mathcal{L}}}

\newcommand{\cN}{{\mathcal{N}}}

\newcommand{\cQ}{{\mathcal{Q}}}
\newcommand{\cR}{{\mathcal{R}}}
\newcommand{\cS}{{\mathcal{S}}}
\newcommand{\cT}{{\mathcal{T}}}
\newcommand{\cU}{{\mathcal{U}}}
\newcommand{\cV}{{\mathcal{V}}}

\newcommand{\cX}{{\mathcal{X}}}
\newcommand{\cY}{{\mathcal{Y}}}

\newcommand{\RR}{\mathbb{R}}

\newcommand{\diag}{{\mathrm{diag}}} 
\newcommand{\col}{{\mathrm{col}}} 
\newcommand{\grad}{{\nabla}}    

\DeclareMathOperator*{\argmin}{arg\,min}

\newcommand{\bc}{\begin{center}}
	\newcommand{\ec}{\end{center}}

\newcommand{\bdm}{\begin{displaymath}}
	\newcommand{\edm}{\end{displaymath}}

\newcommand{\beq}{\begin{equation}}
	\newcommand{\eeq}{\end{equation}}

\newcommand{\bfl}{\begin{flushleft}}
	\newcommand{\efl}{\end{flushleft}}

\newcommand{\bt}{\begin{tabbing}}
	\newcommand{\et}{\end{tabbing}}

\newcommand{\beqn}{\begin{align}}
	\newcommand{\eeqn}{\end{align}}

\newcommand{\beqs}{\begin{align*}} 
	\newcommand{\eeqs}{\end{align*}}  

\newtheorem{assumption}{Assumption}

\newtheorem{lemma}{{Lemma}}
\newtheorem{theorem}{{Theorem}}

\def\tran{^{\mathsf{T}}}

\def\bpsi{{\boldsymbol \psi}}

\def\bgamma {{\boldsymbol \gamma}}

\def\H{{\boldsymbol{H}}}

\def\d{{\boldsymbol{d}}}

\def\h{{\boldsymbol{h}}}

\def\s{{\boldsymbol{s}}}

\def\u{{\boldsymbol{u}}}
\def\v{{\boldsymbol{v}}}
\def\w{{\boldsymbol{w}}}
\def\x{{\boldsymbol{x}}}

\def\z{{\boldsymbol{z}}}

\def\bE{\mathbb{E}}
\def\filt{\boldsymbol{\mathcal{F}}}

\newtheorem{remark}{Remark}

\def\Zint{{\mathchoice{\setbox1=\hbox{\sf Z}\copy1\kern-.75\wd1\box1}
{\setbox1=\hbox{\sf Z}\copy1\kern-.75\wd1\box1}
{\setbox1=\hbox{\scriptsize\sf Z}\copy1\kern-.75\wd1\box1}
{\setbox1=\hbox{\scriptsize\sf Z}\copy1\kern-.75\wd1\box1}}}

\makeatletter
\def\hlinewd#1{%
  \noalign{\ifnum0=`}\fi\hrule \@height #1 \futurelet
   \reserved@a\@xhline}
\makeatother

\title{{\color{black}On the Influence of Bias-Correction on Distributed Stochastic Optimization}}
\author{\IEEEauthorblockN{Kun Yuan, Sulaiman A. Alghunaim, Bicheng Ying,  and Ali H. Sayed \\}
	\vspace{0.4cm}
	
	\thanks{\scriptsize{K. Yuan, S. A. Alghunaim, and B. Ying are with the Department of Electrical and Computer Engineering, University of California, Los Angeles, CA 90095 USA. Email:\{kunyuan, salghunaim, ybc\}@ucla.edu. B. Ying is now with Google, LA. A. H. Sayed is  with School of Engineering, EPFL, CH-1015 Lausanne, Switzerland. Email:\{ali.sayed@epfl.ch\}. This work was supported in part by NSF grant CCF-1524250. }A short preliminary version of this work is under review for IEEE CDC 2019 now.}
}

\begin{document}
%
\maketitle
%
\begin{abstract}
 \color{black} Various bias-correction methods such as EXTRA\cite{shi2015extra}, gradient tracking methods\cite{lorenzo2016next,nedic2017achieving}, and exact diffusion\cite{yuan2017exact1} have been proposed recently to solve distributed {\em deterministic} optimization problems. These methods employ constant step-sizes and converge linearly to the {\em exact} solution under proper conditions. However, their performance under stochastic and adaptive settings is less explored. It is still unknown {\em whether}, {\em when} and {\em why} these bias-correction methods can outperform their traditional counterparts (such as consensus and diffusion) with noisy gradient and constant step-sizes.
 
 This work studies the performance of exact diffusion under the stochastic and adaptive setting, and provides conditions under which exact diffusion has superior steady-state mean-square deviation (MSD) performance than traditional algorithms without bias-correction. In particular, it is proven that this superiority is more evident over sparsely-connected network topologies such as lines, cycles, or grids. Conditions are also provided under which exact diffusion method match or may even degrade the performance of traditional methods. Simulations are provided to validate the theoretical findings.

\end{abstract}

\section{Introduction}
This work considers stochastic optimization problems where a collection of $K$ networked agents work cooperatively to solve an aggregate optimization problem of the form:
\eq{\label{general-prob}
	\hspace{-2mm}	w^\star &= \argmin_{w\in \RR^M}\ \sum_{k=1}^{K} J_k(w), \mbox{ where } J_k(w) = \bE\, Q(w; \x_k)
}
The local risk function $J_k(w)$ held by agent $k$ is assumed to be  differentiable and $\nu$-strongly convex, and it is constructed as the expectation of some loss function $Q(w; \x_k)$. The random variable $\x_k$ represents the streaming data received by agent $k$, and the expectation in $J_k(w)$ is over the distribution of $\x_k$. While the cost functions $J_k(w)$ may have {\em different} local minimizers, all agents seek to determine the {\em common} global solution $w^\star$ under the constraint that agents can only communicate with their {direct} neighbors. Problem \eqref{general-prob} can find applications in a wide range of areas including  wireless sensor networks \cite{rossi2004distributed,li2002detection},  distributed statistical learning \cite{duchi2012dual}, and distributed adaptation and learning \cite{chen2012diffusion,sayed2014adaptive,sayed2014adaptation},.

{There are several techniques that can be used to solve problems of the type \eqref{general-prob} such as  diffusion \cite{chen2012diffusion,tu2012diffusion,sayed2014adaptive,sayed2014adaptation}  and consensus ({also known as decentralized gradient descent}) \cite{nedic2009distributed,kar2013consensus+,yuan2016convergence,tu2012diffusion} strategies. The latter class of strategies has  been shown to be particularly well-suited for stochastic and adaptive learning scenarios from streaming data due to their enhanced stability range over other methods, as well as their ability to track drifts in the underlying models and statistics \cite{tu2012diffusion,sayed2014adaptive,sayed2014adaptation}. We therefore focus on this class of algorithms since we are mainly interested in methods that are able to learn and adapt from data. 
	For example, the adapt-then-combine (ATC) formulation \cite{sayed2014adaptive,sayed2014adaptation} of diffusion takes the following form:}
\eq{
	\bpsi_{k,i} &= \w_{k,i-1} - \mu \grad Q(\w_{k,i-1}; \x_{k,i}), \hspace{3mm}\mbox{{\bf (Adapt)}}\label{diffusion-1}\\
	\w_{k,i} &= \sum_{\ell \in \cN_k} a_{\ell k} \bpsi_{\ell, i}, \hspace{23.5mm}\mbox{{\bf (Combine)}} \label{diffusion-2}
}
where the subscript $k$ denotes the agent index and $i$ denotes the iteration index. The variable $\x_{k,i}$ is the data realization observed by agent $k$ at iteration $i$. The nonnegative scalar $a_{\ell k}$ is the weight used by agent $k$ to scale information received from agent $\ell$, $\cN_k$ is the set of neighbors of agent $k$ (including $k$ itself), {\color{black}and it is required that $\sum_{\ell\in \cN_k}a_{k\ell}=1$ for any $k$}. {In \eqref{diffusion-1}--\eqref{diffusion-2}, variable $\bm{\psi}_{k,i}$ is an intermediate estimate for $w^{\star}$ at agent $k$, while $\w_{k,i}$ is the updated estimate. Note that step \eqref{diffusion-1} uses the gradient of the loss function, $Q(\cdot)$, rather than the gradient of its  expected value $J_k(w)$. This is because the statistical properties of the data are not known beforehand.  
	If $J_k(w)$ were known, then we could use its gradient vector  in \eqref{diffusion-1}. In that case, we would refer to the resulting method as a {\em deterministic} rather than {\em stochastic} solution.} 
Throughout this paper, we employ a {\em constant} step-size $\mu$ to enable continuous adaptation and learning in response to drifts of the global minimizer due to changes in the statistical properties of the data. The adaptation and tracking abilities are crucial in many applications, as already explained in \cite{sayed2014adaptation}. 

{Previous studies have shown that both consensus and diffusion methods are able to solve problems of the type \eqref{general-prob} well for sufficiently small step-sizes. That is, the squared error $\bE\|\widetilde{\w}_{k,i}\|^2$ approaches a small neighborhood around zero for all agents, where $\widetilde{\w}_{k,i}=w^{\star}-\w_{k,i}$.  These methods do not converge to the {\em exact} minimizer $w^{\star}$ of \eqref{general-prob} but rather approach a small neighborhood around $w^{\star}$ with a small steady-state bias under {\em both} stochastic and deterministic optimization scenarios. For example, in deterministic settings where the individual costs $J_k(w)$ are known, it is shown in \cite{chen2013distributed,sayed2014adaptation} that the squared errors $\|\widetilde{\w}_{k,i}\|^2$ generated by the diffusion iterates converge to a $O(\mu^2)$-neighborhood.} Note that, in the deterministic  case, this inherent limiting bias is not due to any gradient noise arising from stochastic approximations; it is instead due to the update structure in diffusion and consensus implementations --- see the explanations in Sec. III.B in \cite{yuan2017exact1}.  For stochastic optimization problems, on the other hand, the size of the bias is $O(\mu)$ rather than $O(\mu^2)$ because of the gradient noise. 

{When high precision is desired, especially in deterministic optimization problems, it would be preferable to remove the $O(\mu^2)$  bias altogether. Motivated by these considerations, the works \cite{yuan2017exact1,yuan2017exact2} showed that a simple correction step  inserted between the adaptation and combination steps \eqref{diffusion-1} and \eqref{diffusion-2} is sufficient to ensure {\em exact} convergence of the algorithm to $w^{\star}$ by all agents --- see expression \eqref{correct} further ahead. In this way, the $O(\mu^2)$  bias is removed completely, and the convergence rate is also improved.}

{While the correction of the second order $O(\mu^2)$  bias is critical in the deterministic setting, it is not clear {\em whether} it can help in the stochastic and adaptive settings. This motivates us to study exact diffusion these settings in this paper and compare against standard diffusion. To this end, we carry out a {\em higher-order} analysis of the error dynamics for both methods, and derive their steady-state performance  as an expansion in the first two powers of the step-size parameter, i.e., $\mu$ and $\mu^2$.  In contrast, traditional analysis for diffusion and consensus focus mainly on performance expressions that depend on a first-order expansion in $\mu$ \cite{sayed2014adaptive,sayed2014adaptation}. Our analysis will reveal conditions under which bias correction improves the performance of diffusion.

\subsection{Main Results}
\label{sec:intro-results}
In particular, we will prove in Theorem \ref{ed-stability}, that, under sufficiently small step-sizes, the exact diffusion strategy will converge exponentially fast, at a rate $\rho = 1 - O(\mu\nu)$, to a neighborhood around $w^\star$. Moreover, the size of the neighborhood will be characterized as 
{\color{black}\eq{
	\hspace{-1mm}\limsup_{i\to \infty}\frac{1}{K}\sum_{k=1}^K\bE\|\tw_{k,i} \|^2_{\rm ed} = O\left(\frac{\mu\sigma^2}{K\nu} + \frac{\delta^2}{\nu^2} \hspace{-0.5mm}\cdot\hspace{-0.5mm} \frac{\mu^2\sigma^2}{1-\lambda}\right) \label{ed-steady-state}	
}}where $\delta$ and $\nu$ are the Lipschitz and strong convexity constants, the quantity $\sigma^2$ is a measure of the variance of the gradient noise, and $\lambda \in (0, 1)$ is the second largest magnitude of the eigenvalues of the combination matrix $A = [a_{\ell k}]$ which reflects the level of network connectivity.   The subscript ${\rm ed}$ indicates that $\w_{k,i}$ is generated by the exact diffusion method.  In comparison, we will show that the traditional diffusion strategy converges at a similar rate albeit to the following neighborhood:
{\color{black}\eq{
	&\hspace{-2mm} \limsup_{i\to \infty}\frac{1}{K}\sum_{k=1}^K\bE\|\tw_{k,i}\|^2_{\rm d} \nnb
	&\ = O\left(\frac{\mu\sigma^2}{K \nu} + \frac{\delta^2}{ \nu^2}\cdot\frac{\mu^2\lambda^2\sigma^2}{1-\lambda} + \frac{\delta^2}{\nu^2}\cdot\frac{\mu^2\lambda^2 b^2}{(1-\lambda)^2}\right) \label{d-steady-state}
}}where the subscript ${\rm d}$ indicates that $\w_{k,i}$ is generated by the diffusion method \eqref{diffusion-1}--\eqref{diffusion-2}, and $b^2 = (1/K)\sum_{k=1}^K \|\grad J_k(w^\star)\|^2$ is a bias constant independent of the gradient noise. Observe that the expressions on the right-hand side of \eqref{ed-steady-state} and \eqref{d-steady-state} depend on $\mu$ and $\mu^2$. These are therefore more refined performance expressions, which are more challenging to derive than earlier expressions that just depend on $\mu$ (see \cite{sayed2014adaptive,sayed2014adaptation,chen2012diffusion,nedic2009distributed,chen2013distributed}). The terms that depend on $\mu^2$ in \eqref{ed-steady-state} and \eqref{d-steady-state} help reveal the important insights that arise from using the exact diffusion strategy.


Expressions \eqref{ed-steady-state} and \eqref{d-steady-state} have the following important implications. First, it is obvious that diffusion suffers from an additional bias term $\mu^2 \lambda^2 b^2/(1-\lambda)^2$, which is independent of the gradient noise $\sigma^2$, while exact diffusion removes it completely. In the deterministic setting when the gradient noise $\sigma^2 = 0$, it is observed from \eqref{ed-steady-state} and \eqref{d-steady-state} that diffusion converges to an $O(\mu^2)$-neighborhood around the global solution $w^\star$ while exact diffusion converges exactly to $w^\star$. This result is consistent with \cite{chen2013distributed,sayed2014adaptation,yuan2017exact2}.

{\color{black}Second, it is further observed that the performance of diffusion and exact diffusion differs only on  the $O(\mu^2)$ terms inside \eqref{ed-steady-state} and \eqref{d-steady-state}. When the step-size is moderately small so that these $O(\mu^2)$ terms are non-negligible, then the superiority of exact diffusion or diffusion will highly depend on the network topology. In particular, when the network topology is sparsely-connected (in which case $\lambda$ approaches $1$), the bias term $\mu^2 \lambda^2 b^2/(1-\lambda)^2$ will be significantly large and the correction of this term will greatly improve the steady-state performance. It should be emphasized that the bias-correction property of exact diffusion is particularly critical for large-scale {\em linear} or {\em cyclic} networks where $1-\lambda = O(1 / K^2)$ and {\em grid} networks where $1-\lambda = O(1 / K)$ since the bias term will grow rapidly on these network topologies as the size $K$ increases. On the other hand, when the network is well-connected (in which case $\lambda$ approaches $0$), one can find that the $O(\mu^2)$ terms in diffusion \eqref{d-steady-state} diminishes while the $O(\mu^2)$ term  in exact diffusion \eqref{ed-steady-state} still exists. This implies that for well connected networks and moderatly-small step-sizes, diffusion is a better choice than  exact diffusion. The comparison between \eqref{ed-steady-state} and \eqref{d-steady-state} provides guidelines on the proper choice of diffusion or exact diffusion in various application scenarios.  }

\begin{table*}[t!]
	\centering \caption{ Performance of Exact Diffusion and Diffusion under different scenarios.} 
	\begin{tabular}{ | c | c  | c | c | c| }
		\hline   
	 \multicolumn{5}{ |c| }{\cellcolor{gray!45} \bf Moderately small step-size $\mu$} \\
	  \hline	
	 \cellcolor{gray!15} \bf Scenario &   \cellcolor{gray!15} \bf  Network 
		&     \cellcolor{gray!15} \bf  Diffusion &     \cellcolor{gray!15} \bf  Exact Diffusion &     \cellcolor{gray!15} \bf  Better Algorithm  \\
		 
\hline
		\multirow{2}{*}{$b^2=0, \sigma^2\neq0$}    & Sparse  ($\lambda \rightarrow 1$)  & $O(\mu\sigma^2 + \frac{\mu^2  \sigma^2}{1-\lambda})$ &  $O(\mu\sigma^2 + \frac{\mu^2\sigma^2}{1-\lambda})$   &                Similar performance       \\
	\cline{2-2}	& Dense ($\lambda \rightarrow 0$) & $O(\mu\sigma^2)$       &  $O(\mu\sigma^2+\mu^2 \sigma^2)$ &                Diffusion      \\ 
		\hline
		\multirow{2}{*}{$b^2 \neq 0, \sigma^2=0$}    & Sparse  ($\lambda \rightarrow 1$)  & $O({\mu^2 b^2  \over (1- \lambda)^2})$        & 0 &                Exact diffusion     \\
	\cline{2-2}	& Dense ($\lambda \rightarrow 0$) & $O(\mu^2 b^2 \lambda^2 )$        & 0 &                Exact diffusion$^\dagger$    \\
	\hline
		\multirow{2}{*}{$b^2 \neq 0, \sigma^2\neq0$}    & Sparse  ($\lambda \rightarrow 1$)  & $O\left(\mu\sigma^2 +\frac{\mu^2 \sigma^2}{1-\lambda} + \frac{\mu^2  b^2}{(1-\lambda)^2}\right)$       & $O\left(\mu\sigma^2 +\frac{\mu^2 \sigma^2}{1-\lambda}\right)$ &                Exact diffusion            \\
	\cline{2-2}	& Dense ($\lambda \rightarrow 0$) & $O\left(\mu\sigma^2  \right)  $    &  $O\left(\mu\sigma^2 +\frac{\mu^2 \sigma^2}{1-\lambda}\right)$ &                Diffusion   
	 \\ 
			
		\hline \hline
\multicolumn{5}{ |c| }{\cellcolor{gray!45} \bf Sufficiently small step-size $\mu$} \\
	  \hline	
	 \cellcolor{gray!15} \bf Scenario &   \cellcolor{gray!15} \bf  Network 
		&     \cellcolor{gray!15} \bf  Diffusion &     \cellcolor{gray!15} \bf  Exact Diffusion &     \cellcolor{gray!15} \bf  Better Algorithm  \\
\hline  
		$b^2 \neq 0, \sigma^2=0$    & Sparse  or dense & $O(\mu^2 b^2 \lambda^2 / (1- \lambda)^2)$       & 0 &                                   Exact diffusion$^\dagger$ \\ 
		\hline
		All other scenarios    & Sparse  or dense  & $O(\mu\sigma^2)$       & $O(\mu\sigma^2)$  &                Similar performance         \\
		\hline
	\end{tabular} 
	\newline \newline
	{{\color{white}s}\hspace{-103mm} $^\dagger$ Exact diffusion performs better unless $\lambda=0$}
	\label{table-ncomparison}
	\end{table*}

{Third, the difference between exact diffusion and diffusion will vanish as the step-size $\mu$ approaches $0$. This is because $O(\mu \sigma^2/K \nu)$ will  dominate the $O(\mu^2)$ terms when $\mu$ is sufficiently small, i.e., 
	%
	\eq{
		\limsup_{i\to \infty}\frac{1}{K}\sum_{k=1}^K\bE\|\w_{k,i} - w^\star\|^2_{\mathrm{ed}} &= O\Big({\mu\sigma^2}/{K \nu}\Big), \label{7sgh-1-0} \\
		\limsup_{i\to \infty}\frac{1}{K}\sum_{k=1}^K\bE\|\w_{k,i} - w^\star\|^2_{\mathrm{d}} &= O\Big({\mu\sigma^2}/{K \nu}\Big). \label{7sgh-2-0}
	}
	The ``sufficiently'' small $\mu$ can be roughly characterized as $\mu \le c_3 (1-\lambda)^{2+x}$, 
	where $x$ is any positive constant. 
	While relations \eqref{7sgh-1-0} and \eqref{7sgh-2-0} show diffusion and exact diffusion have the same upper bound on the steady-state performance, however, it is still an upper bound and not an exact expression. To more accurately characterize the steady-state performance of diffusion and exact diffusion when $\mu$ is sufficiently small, we shall establish the precise MSD expression defined as \cite{sayed2014adaptation}:
	\eq{
		\label{msd-definition-0}
		\mbox{MSD} = \mu \left(\lim_{\mu \to 0} \limsup_{i\to \infty} \frac{1}{\mu K}\sum_{k=1}^K \bE\|\tw_{k,i}\|^2\right).
	}
for exact diffusion	and find that it matches that of diffusion:
	\eq{\label{38sdgsd00}
		\mbox{MSD}_{\rm ed}	\hspace{-1mm}= \hspace{-1mm} \mbox{MSD}_{\rm d} \hspace{-1mm} = \hspace{-1mm} \frac{\mu}{2K} \Tr\left\{\left(\sum_{k=1}^K H_k\right)^{-1}\left(\sum_{k=1}^K S_k\right)\right\},
	}
	where $H_k = \grad^2 J_k(w^\star)$ and $S_k$ is the covariance matrix of gradient noise.
	Obviously, the MSD expression \eqref{msd-definition-0} is exact to  first order in $\mu$ and ignores all  higher-order terms. Equality \eqref{38sdgsd00} states that when  $\mu$ is sufficiently small, both diffusion and exact diffusion perform {\em exactly the same} during the steady-state stage. The main results derived in this paper are summarized in Table \ref{table-ncomparison} in which we omit the constants $\delta$, $\nu$ and $K$ for clarity.
}

\subsection{Related work}
{
	In addition to exact diffusion, there exist some other useful bias-correction methods such as  EXTRA\cite{shi2015extra,shi2015proximal}, DIGing or gradient-tracking methods \cite{di2016next,nedic2017achieving
	,qu2018harnessing,pu2018push,xin2018linear}, Aug-DGM\cite{xu2015augmented,nedic2016geometrically} and NIDS\cite{li2017decentralized}. All these methods converge linearly to the exact solution under the deterministic setting, but their performance (especially their advantage over diffusion or consensus) in the stochastic and adaptive settings remains unexplored and/or unclear. The recent work \cite{pu2018distributed} studies the gradient-tracking method (referred to as DIGing in \cite{nedic2017achieving}) to the stochastic  setting and shows that it can outperform the decentralized gradient descent (DGD) \cite{yuan2016convergence,nedic2009distributed} via numerical simulations. However, it does not analytically discuss {\em when} and {\em why} bias-correction methods can outperform consensus. Similarly,  the work \cite{xin2019distributed} studies the gradient-tracking method \cite{pu2018push,xin2018linear} under the stochastic setting and shows that it converges linearly around a neighborhood of the minimizer. No comparison with diffusion or consensus is presented in \cite{xin2019distributed}. Another useful work is \cite{tang2018d}, which establishes the convergence property of exact diffusion with decaying step-sizes in the stochastic and non-convex setting. It proves exact diffusion is less sensitive to the data variance across the network than diffusion and is therefore endowed with a better convergence rate when the data variance is large. Different from \cite{tang2018d}, our bound in \eqref{d-steady-state} shows that even small data variances (i.e., $b^2$) can be significantly amplified by a bad network connectivity -- see the example graph topologies discussed in Sec. \ref{sec-comparison-bad-network}. This observation implies that the  superiority of exact diffusion does not just rely on its robustness to data variance, but more importantly, on  the network connectivity as well. In addition, different from the works  \cite{pu2018distributed,tang2018d}, which claim or suggest that the gradient-tracking method\cite{pu2018distributed} or exact diffusion\cite{tang2018d} always converges better than traditional DGD or diffusion, our current work disproves this statement and clarifies analytically that there are important scenarios where exact diffusion performs similarly or even worse than diffusion. {Simulations also suggest that gradient tracking methods \cite{pu2018distributed,xin2019distributed} may also degrade the performance of traditional diffusion, which was not explored prior to this work.} Finally, we remark that work \cite{towfic2015stability} showed that diffusion outperforms traditional primal-dual methods in the stochastic setting for $b^2=0$ and quadratic problems only, and is hence more restricted than our result.  Our results recover this case (see Remark \ref{remark_towfic_result}) and show that exact diffusion, which is also a primal-dual  method, can outperform diffusion when $b^2\neq 0$.
}

\noindent \textbf{Notation.} Throughout the paper we use $\col\{x_1, \cdots, x_K\}$ and $\diag\{x_1,\cdots,x_K\}$ to denote a column vector  and a diagonal matrix formed from $x_1,\cdots, x_K$. The notation $\mathds{1}_K = \col\{1,\cdots, 1\}\in \RR^K$ and $I_K \in \RR^{K\times K}$ is an identity matrix. The Kronecker product is denoted by ``$\otimes$''. For two matrices $X$ and $Y$, the notation $X\ge Y$ denotes $X-Y$ is nonnegative.

\section{Exact Diffusion Strategy}
\subsection{Exact Diffusion Recursions}
{The exact diffusion strategy from \cite{yuan2017exact1,yuan2017exact2} was originally proposed to solve deterministic optimization problems. We adapt it to solve stochastic optimization problems by replacing the gradient of the local cost $J_k(w)$ by the stochastic gradient of the corresponding loss function. That is, we now use:}
\eq{
	\bpsi_{k,i} &= \w_{k,i-1} \hspace{-0.5mm}-\hspace{-0.5mm} \mu \grad Q(\w_{k,i-1}; \x_{k,i}), \hspace{4.8mm} \mbox{{\bf(Adapt)}} \label{adapt}\\
	\bphi_{k,i} &= \bpsi_{k,i} + \w_{k,i-1}  - \bpsi_{k,i-1}, \hspace{11.5mm} \mbox{{\bf (Correct)}} \label{correct}\\
	\w_{k,i} &= \sum_{\ell\in \cN_k} {\bar{a}_{\ell k}} \bphi_{\ell,i}. \hspace{2.93cm} \mbox{{\bf (Combine)}}  \label{combine}
}
{\color{black}For the initialization, we let $\w_{k,-1}=\bpsi_{k,-1}=0$.} Observe that the fusion step \eqref{combine} now employs the corrected iterates from \eqref{correct} rather than the intermediate iterates from \eqref{adapt}. Note that the weight $\bar{a}_{\ell k}$ is different from $a_{\ell k}$ used in the  diffusion recursion \eqref{diffusion-2}. If we let $A = [a_{\ell k}]\in \RR^{K\times K}$ and $\bar{A} = [\bar{a}_{\ell k}] \in \RR^{K\times K} $ denote the combination matrices used in diffusion and exact diffusion respectively, then the relation between them is $\bar{A} = (A + I_K)/2$. In the paper, we assume $A$ (and, hence, $\bar{A}$) to be symmetric and doubly stochastic.  

As explained in \cite{yuan2017exact1,yuan2017exact2}, exact diffusion is essentially a primal-dual method. We can describe its operation more succinctly by collecting the iterates and gradients from across the network into global vectors. Specifically, we introduce
\eq{\label{grad Q definition}
	\hspace{-2mm}	\swb_i \hspace{-1mm}=\hspace{-1mm}
	\ba{c}
	\hspace{-2mm}\w_{1,i} \hspace{-2mm}\\
	\hspace{-2mm}\vdots \hspace{-2mm}\\
	\hspace{-2mm}\w_{K,i}\hspace{-2mm}
	\ea\ 
	, \ \grad \cQ(\swb_{i-1};\sxb_i) \hspace{-1mm}=\hspace{-1mm} 
	\ba{cc}
	\hspace{-2mm} \grad Q(\w_{1,i-1};\x_{1,i}) \\
	\hspace{-2mm} \vdots \\
	\hspace{-2mm} \grad Q(\w_{K,i-1};\x_{K,i})
	\ea\hspace{-1mm} \hspace{-2mm}
}
$\cA = A \otimes I_M$ and $\tcA = (\cA + I_{KM})/2$.  Then recursions \eqref{adapt}--\eqref{combine} lead to the second-order recursion
\eq{\label{exact-diffusion-compact}
	\swb_i &= \tcA \Big( 2\swb_{i-1} - \swb_{i-2} - \mu \grad \cQ(\swb_{i-1};\sxb_i) \nnb
	&\hspace{1cm} + \mu \grad \cQ(\swb_{i-2};\sxb_{i-1}) \Big) 	.
}
The initialization is $\swb_{-1}=0$ and $\swb_0=\tcA (\swb_{-1} - \mu\grad \cQ(\swb_{-1};\sxb_i))$. We can rewrite the update \eqref{exact-diffusion-compact} in a primal-dual form as follows. First, since the combination matrix $\bar{A}$ is symmetric and doubly stochastic, it holds that $I-\bar{A}$ is positive semi-definite. By introducing the eigen-decomposition $I - \bar{A} = U \Sigma U\tran$ and defining $V = U \Sigma^{1/2} U\tran \in \RR^{K\times K}$, where $\Sigma$ is a non-negative diagonal matrix, we know that $V$ is also positive semi-definite and $V^2 = I-\bar{A}$. Furthermore, if we let $\cV = V\otimes I_M$ then $\cV^2 = I_{KM} - \tcA$. With these relations, it can be verified\footnote{To verify it, one can substitute the second recursion in \eqref{exact-diffusion-primal-dual} into the first recursion to remove $\syb_i$ and arrive at \eqref{exact-diffusion-compact}.} that recursion \eqref{exact-diffusion-compact} is equivalent to
\eq{\label{exact-diffusion-primal-dual}
	\begin{cases}
		\swb_i = \tcA \big(\swb_{i-1} - \mu \grad \cQ(\swb_{i-1};\sxb_i)\big) - \cV \syb_{i-1}, \\
		{\color{white}l} \syb_i = \syb_{i-1} + \cV \swb_{i},
	\end{cases}	
}
for $i \geq 0$ with $\sy_{-1}=0$ where $\sy_i \in \RR^{KM}$ plays the role of a dual variable. 
{The analysis in \cite{yuan2017exact1,yuan2017exact2} explains how the correction term in \eqref{correct} guarantees {\em exact} convergence to $w^{\star}$ by all agents in deterministic optimization problems where the true gradient $\grad J_k(w)$ is available. In the following sections, we examine the convergence of exact diffusion \eqref{adapt}--\eqref{combine} in the stochastic setting.
}

\section{Error Dynamics of Exact Diffusion}
To establish the error dynamics of exact diffusion, we first introduce some standard assumptions. These assumptions are common in the literature (e.g, \cite{pu2018distributed,sayed2014adaptation}).
\begin{assumption}[\sc Conditions on cost functions]
	\label{ass-lip} 
	Each $J_k(w)$ is $\nu$-strongly convex and twice differentiable, and its Hessian matrix satisfies 
	\eq{\label{xzh2300}
		\nu I_M \le \grad^2 J_k(w) \le \delta I_M, \quad \forall\ k. 
	}
	\qd
\end{assumption}
\noindent {We remark that the twice differentiability assumption is necessary to derive the MSD expression in Sec. \ref{sec-MSD}.} 

{\begin{assumption}[\sc Conditions on combination matrix]
		\label{ass-combination-matrix} 
		The network is undirected and strongly connected, and the combination matrix $A$ satisfies
		\eq{\label{A-cond}
			A = A\tran, \quad A\mathds{1}_K = \mathds{1}_K, \quad \mathds{1}_K\tran A = \mathds{1}_K\tran.
		}
		\qd
	\end{assumption}
	Assumption \ref{ass-combination-matrix} implies that $\bar{A} = (I + A)/2$ is also symmetric and doubly-stochastic. Since the network is strongly connected, it holds that 
	\eq{
		1 = \lambda_1(\bar{A}) > \lambda_2(\bar{A}) \ge \cdots \ge \lambda_K(\bar{A}) > 0.	
	}

	To establish the optimality condition for problem \eqref{general-prob}, we introduce the following notation:
	\eq{
		\sw &= \col\{w_1, \cdots, w_K\} \in \RR^{KM}, \label{23nds} \\
		\grad \cJ(\sw) &= \col\{\grad J_1(w_1),\cdots, \grad J_K(w_K)\},
	}
	where $w_k$ in \eqref{23nds} is the $k$-th block entry of vector $\sw$. 
	With the above notation, the following lemma from \cite{yuan2017exact2} states the optimality condition for problem \eqref{general-prob}.}


\begin{lemma}[\sc Optimality Condition]\label{lm-opt-cond-1} Under Assumption \ref{ass-lip}, 
	if some block vectors $(\sw^\star, \sy^\star)$ exist that satisfy:
	\eq{
		\mu \tcA \grad \cJ(\sw^\star) + \cV \sy^\star & = 0, \label{KKT-1-1} \\
		\cV \sw^\star &= 0. \label{KKT-2-2}
	}
	then it holds that the block entries in ${\sw}^{\star}$ satisfy: 
	\eq{\label{7280m}
		w_1^\star=w_2^\star=\cdots=w_N^\star=w^\star，
	}
	where $w^\star$ is the unique solution to problem \eqref{general-prob}. 
	\qd
\end{lemma}

\subsection{Error Dynamics}
We define the gradient noise at agent $k$ as 
\eq{
	\label{grad-noise}
	\s_{k,i}(\w_{k, i-1}) \define \grad Q(\w_{k,i-1}; \x_{k,i}) - \grad J_k(\w_{k,i-1})
}
and collect them into the network vector
\eq{
	\hspace{-2mm}	\s_i(\swb_{i-1}) &= \col\{\s_{1,i}(\w_{1, i-1}),\cdots, \s_{K,i}(\w_{K, i-1})\} \label{network-gn}\\
	\hspace{-2mm}	\grad \cJ(\swb_{i-1}) &= \col\{\grad J_1(\w_{1, i-1}),\cdots, \grad J_K(\w_{K, i-1})\}
}
It then follows that 
\eq{\label{grad J = grad Q + s}
	{\grad \cQ(\swb_{i-1}; \sxb_i) = \grad \cJ(\swb_{i-1}) + \s_i(\swb_{i-1}).}
}
Next, we introduce the error vectors
\eq{\label{error-vectors}
	\twb_i = \sw^\star - \swb_i, \quad \tyb_i = \sy^\star - \syb_i
}
where $(\sw^\star, \sy^\star)$ are optimal solutions satisfying \eqref{KKT-1-1}--\eqref{KKT-2-2}. By combining \eqref{exact-diffusion-primal-dual}, \eqref{KKT-1-1}, \eqref{KKT-2-2}, \eqref{grad J = grad Q + s} and \eqref{error-vectors}, we reach 
\eq{\label{exact-diffusion-error}
	\begin{cases}
		\twb_i = \tcA \big[\twb_{i-1} + \mu (\grad \cJ(\swb_{i-1}) - \grad \cJ(\swb^\star) )\big] \\
		\hspace{3cm} - \cV \tyb_{i-1} + \mu \tcA \s_i(\swb_{i-1}), \\
		{\color{white}l} \tyb_i = \tyb_{i-1} + \cV \twb_{i}.
	\end{cases}	
}
Since each $J_k(w)$ is twice-differentiable (see Assumption \ref{ass-lip}), we can appeal to the mean-value theorem from Lemma D.1 in \cite{sayed2014adaptation}, which allows us to express each difference in \eqref{exact-diffusion-error} in terms of Hessian matrices for any~ $k=1,2,\ldots,N$:
\eq{
	\grad J_k(\w_{k,i-1}) - \grad J_k(w^\star)  =  - \H_{k,i-1} \widetilde{\w}_{k,i-1}, \nn
}
where 
\eq{\label{H_k_i-1}
	\H_{k,i-1} \hspace{-1.5mm}\define \hspace{-1.5mm} \int_0^1 \grad^2 J_k\big(w^\star \hspace{-0.5mm}-\hspace{-0.5mm} r\widetilde{\w}_{k,i-1}\big)dr \in \RR^{M\times M}
}
We introduce the block diagonal matrix
\eq{\label{H_i-1}
	{\boldsymbol \cH}_{i-1} \hspace{-1.5mm}\define \hspace{-1.5mm} \mathrm{diag}\{\H_{1,i-1},\H_{2,i-1},\cdots,\H_{K,i-1}\}
}
so that
\eq{\label{mean-value}
	\grad \cJ(\swb_{i-1}) - \grad \cJ(\sw^\star) = - {\boldsymbol \cH}_{i-1} \twb_{i-1}.
}
Substituting \eqref{mean-value} into the first recursion in \eqref{exact-diffusion-error}, we reach
\eq{\label{exact-diffusion-error-2}
	\hspace{-2mm}	\begin{cases}
		\twb_i \hspace{-1mm}=\hspace{-1mm} \tcA (I_{KM} \hspace{-1mm}-\hspace{-1mm} \mu {\boldsymbol \cH}_{i-1}) \twb_{i-1} \hspace{-1mm}-\hspace{-1mm} \cV \tyb_{i-1} \hspace{-1mm}+\hspace{-1mm} \mu \tcA \s_i(\swb_{i-1}),\hspace{-10mm} \\ 
		{\color{white}l} \tyb_i \hspace{-1mm}=\hspace{-1mm} \tyb_{i-1} + \cV \twb_{i}.
	\end{cases}	\hspace{-5mm}
}
{Next, if we substitute the first recursion in \eqref{exact-diffusion-error-2} into the second one, and recall that $\cV^2 = I_{KM} - \tcA$, we reach the following error dynamics.}

\begin{lemma}[\sc Error Dynamics] 
	Under Assumption \ref{ass-lip}, the error dynamics for the exact diffusion recursions \eqref{adapt}--\eqref{combine} is as follows
	\eq{\label{original-error-dynamics}
		\ba{c}
		\twb_i\\
		\tyb_i
		\ea
		\hspace{-1mm}=&\hspace{-0.2mm} 
		\underbrace{\ba{cc}
			\tcA & -\cV \\
			\cV \tcA & \tcA
			\ea}_{\define \cB} \Big( I_{2KM} \hspace{-0.5mm} 
		- \mu 
		\underbrace{\ba{cc}
			{\boldsymbol \cH}_{i-1} & 0 \\
			0 & 0
			\ea}_{\define {\boldsymbol\cT_{i-1}}} \hspace{-0.5mm} \Big) 	\hspace{-1mm}
		\ba{c}
		\twb_{i-1}\\
		\tyb_{i-1}
		\ea \nnb
		&\ + \mu \underbrace{\ba{c}
			\tcA \\
			\cV \tcA
			\ea}_{\define \cB_\ell} \s_i(\swb_{i-1}),
	}
	and ${\boldsymbol \cH}_{i}$ is defined in \eqref{H_i-1}.  \qd
\end{lemma}

\subsection{Transformed Error Dynamics}
The direct convergence analysis of recursion \eqref{original-error-dynamics} is challenging. To facilitate the analysis, we identify a convenient change of basis and transform \eqref{original-error-dynamics} into another equivalent form that is easier to handle. To this end, we introduce a fundamental decomposition from \cite{yuan2017exact2} here.

\begin{lemma}[\sc Fundamental Decomposition]\label{lemma-fundamental-decomposition}
	Under Assumptions \ref{ass-lip} and \ref{ass-combination-matrix}, the matrix $\cB$ defined in \eqref{original-error-dynamics} can be decomposed as
	\eq{\label{cB-decom}
		\hspace{-2mm}		\cB \hspace{-1mm}=\hspace{-1mm} 
		\underbrace{
			\ba{ccc}
			\hspace{-1mm}\cR_1 & \cR_2 & c\cX_R\hspace{-1mm}
			\ea}_{\cX}
		\underbrace{\ba{ccc}
			\hspace{-1mm}I_M & 0 & 0\hspace{-1mm}\\
			\hspace{-1mm}0 & I_M & 0\hspace{-1mm}\\
			\hspace{-1mm}0 & 0 & \cD_1\hspace{-1mm}
			\ea}_{\cD}
		\underbrace{\ba{c}
			\cL_1\tran \\
			\cL_2\tran \\
			\frac{1}{c}\cX_L
			\ea}_{\cX^{-1}}
	}
	where $c$ can be any positive constant, and $\cD\in \RR^{2KM\times 2KM}$ is a diagonal matrix. Moreover, we have
	\eq{
		\cR_1 &= \ba{c}
		\hspace{-1mm}\cI\hspace{-1mm} \\
		\hspace{-1mm}0\hspace{-1mm}
		\ea \in \RR^{2KM\times M},\hspace{3mm}
		\cR_2 \hspace{-1mm}= 
		\ba{c}
		\hspace{-1mm}0\hspace{-1mm} \\
		\hspace{-1mm}\cI\hspace{-1mm}
		\ea \in \RR^{2KM \times M}, \\
		\cL_1 &= \ba{c}
		\hspace{-1.5mm}\frac{1}{K}\cI\hspace{-1.5mm}\\
		\hspace{-1.5mm}0\hspace{-1.5mm}
		\ea \hspace{-1mm}\in \RR^{2KM\times M}, \ 
		\cL_2 = \hspace{-1mm}
		\ba{c}
		\hspace{-1.5mm}0\hspace{-1.5mm} \\
		\hspace{-1.5mm}\frac{1}{K}\cI\hspace{-1.5mm}
		\ea \hspace{-1mm} \in \RR^{2KM\times M}, \label{L}\\
		\cX_R &\in \RR^{2KM \times 2(K-1)M}, \hspace{6mm}
		\cX_L \in \RR^{2(K-1)M \times 2KM}.
	}
	where $\cI = \mathds{1}_K \otimes I_M \in \RR^{KM\times M}$. Also, the matrix $\cD_1$ is a diagonal matrix with complex entries. The magnitudes of the diagonal entries in $\cD_1$ are all strictly less than $1$. \qd
\end{lemma}

By multiplying $\cX^{-1}$ to both sides of the error dynamics \eqref{original-error-dynamics} and simplifying we arrive at the following result. 

\begin{lemma}[\sc Transformed Error Dynamics] \label{lm-transform-error-dyanmic}
	Under Assumption \ref{ass-lip} and \ref{ass-combination-matrix}, the transformed error dynamics for exact diffusion recursions \eqref{adapt}--\eqref{combine} is as follows
	\eq{
		\label{transformed-error-dynamics}
		\ba{c}
		\bar{\szb}_i \\
		\check{\szb}_i
		\ea	
		\hspace{-1mm}&=\hspace{-1mm} 
		\ba{cc}
		\hspace{-1mm}I_M \hspace{-1mm}-\hspace{-1mm} \frac{\mu}{K}\sum_{k=1}^K \H_{k,i-1} & -\frac{c\mu}{K}\cI\tran {\boldsymbol \cH}_{i-1} \cX_{R,u}\hspace{-1mm}\\
		\hspace{-1mm}-\frac{\mu}{c} \cD_1\cX_L {\boldsymbol \cT}_{i-1} \cR_1 & \cD_1 - \mu \cD_1\cX_L {\boldsymbol \cT}_{i-1} \cX_R\hspace{-1mm}
		\ea \nnb
		&\quad \times 
		\ba{c}
		\bar{\szb}_{i-1} \\
		\check{\szb}_{i-1}
		\ea	
		+ \mu 
		\ba{c}
		\frac{1}{K}\cI\tran\\
		\frac{1}{c}\cD_1\cX_L \cB_\ell
		\ea \s_i(\swb_{i-1}).
	}
	where $\cX_{R,u}\in \RR^{KM\times 2(K-1)M}$ is the upper part of matrix $\cX_R = [\cX_{R,u}; \cX_{R,d}]$. 
	The relation between the original and transformed error vectors are 
	\eq{\label{w-y-vs-bz-cz}
		\ba{c}
		\hspace{-1mm}\twb_i\hspace{-1mm}\\
		\hspace{-1mm}\tyb_i\hspace{-1mm}
		\ea
		\hspace{-1mm}	= {\hspace{-1mm}
			\ba{ccc}
			\hspace{-1mm}\cR_1 \hspace{-1mm}&\hspace{-1mm} c\cX_R\hspace{-1mm}
			\ea}\hspace{-1mm}
		\ba{c}
		\hspace{-1mm}\bar{\szb}_i\hspace{-1mm} \\
		\hspace{-1mm}\check{\szb}_i\hspace{-1mm}
		\ea.\hspace{-1mm}
	}
	\qd
\end{lemma}

\section{Mean-square Convergence}\label{sec-mean-square-stability}
Using the transformed error dynamics derived in \eqref{transformed-error-dynamics}, we can now analyze the mean-square convergence of exact diffusion \eqref{adapt}--\eqref{combine} in the stochastic and adaptive setting. To begin with, we introduce the filtration 
\eq{
	\filt_{i-1} = \mbox{filtration}\{ \w_{k,-1}, \w_{k,0}, \cdots, \w_{k,i-1}, \mbox{ all } k \}.
}
The following assumption is standard on the gradient noise process (see \cite{sayed2014adaptation,pu2018distributed}) and is satisfied in many situations of interest such as  linear and logistic regression problems.
\begin{assumption}[\sc Conditions on gradient noise]
	\label{ass-grad-noise} 
	It is assumed that the first and second-order conditional moments of the individual gradient noises for any $k$ and $i$ satisfy
	\eq{
		\bE[\s_{k,i}(\w_{k, i-1})|\filt_{i-1}]	& = 0, \label{1st-moment}\\
		\bE[\|\s_{k,i}(\w_{k, i-1})\|^2|\filt_{i-1}]& \le \beta^2_k \|\tw_{k,i-1}\|^2 \hspace{-1mm}+\hspace{-1mm} \sigma_k^2  \label{2nd-moment}
	}
	for some constants $\beta_k$ and $\sigma_k$. Moreover, we assume the $\s_{k,i}(\w_{k,i-1})$ are independent of each other for any $k,i$ given $\filt_{i-1}$.
	\qd
\end{assumption}

With Assumption \ref{ass-grad-noise}, it can be verified that
\eq{
	\bE[\s_{i}(\swb_{i-1})|\filt_{i-1}]	& = 0, \quad \forall\ i, \label{network-ns-1st}  \\ 
	\bE\left[\Big\|\frac{1}{K}\hspace{-0.5mm}\sum_{k=1}^K\s_{k,i}(\w_{k,i-1})\Big\|^2\Big|\filt_{i-1}\right] & \hspace{-0.5mm} \le \hspace{-0.5mm} \frac{\beta^2}{K} \|\twb_{i-1}\|^2 \hspace{-1mm} + \hspace{-1mm} \frac{\sigma^2}{K} \label{network-ns-2nd} 
}
where $\beta^2 \define \max_k\{\beta^2_k\}/K$ and $\sigma^2 \define \sum_{k=1}^K \sigma_k^2/K$. 

\begin{theorem}[\sc Mean-Square Convergence]\label{ed-stability}
	Under Assumptions \ref{ass-lip}--\ref{ass-grad-noise}, if the step-size $\mu$ satisfies 
	\eq{\label{ed-stability-range}
		\hspace{-2mm} \mu \le \frac{(1-\lambda)\nu}{(32\hspace{-0.5mm}+\hspace{-0.5mm}16c_1c_2\hspace{-0.5mm}+\hspace{-0.5mm}8\sqrt{c_1c_2})(\delta^2\hspace{-0.5mm}+\hspace{-0.5mm}\beta_{\rm max}^2)} \hspace{-1mm}=\hspace{-1mm} O\left(\hspace{-0.5mm}\frac{(1-\lambda)\nu}{\delta^2 \hspace{-0.8mm}+\hspace{-0.8mm} \beta_{\rm max}^2}\hspace{-0.5mm}\right)\hspace{-1mm}
	}
	where $\lambda = \max\{|\lambda_2(A)|, |\lambda_K(A)|\}$, $\beta^2_{\rm max}=\max_k\{\beta^2_k\}$, and $c_1$, $c_2$ are constants defined in \eqref{constants-c1-c2}, then the $\w_{k,i}$ generated by exact diffusion recursion \eqref{exact-diffusion-primal-dual} converges exponentially fast to a neighborhood around $w^\star$. The convergence rate is $\rho = 1-O(\mu\nu)$, and the size of the neighborhood can be characterized as follows:
	\eq{\label{ed-neighborhood}
		\hspace{-1mm}\limsup_{i\to \infty}\frac{1}{K}\sum_{k=1}^K\bE\|\widetilde{\w}_{k,i}\|^2 = O\left(\frac{\mu\sigma^2}{K \nu} \hspace{-1mm}+\hspace{-1mm} \frac{\delta^2}{ \nu^2} \cdot \frac{\mu^2\sigma^2}{1-{\lambda}}\right)
	}
\end{theorem}
\noindent \textbf{Proof.} See Appendix \ref{appendix_ed_stability_proof}. \qd

Theorem \ref{ed-stability} indicates that when $\mu$ is smaller than a specified upper bound, the exact diffusion over adaptive networks is stable. The theorem also provides a bound on the size of the steady-state mean-square error. To compare exact diffusion with diffusion, we examine the mean-square convergence property of diffusion as well.
\begin{lemma}[\sc Mean-square stability of Diffusion]\label{diffusion-stability}
Under Assumptions \ref{ass-lip}--\ref{ass-grad-noise}, if $\mu$ satisfies 
	\eq{\label{mu_range_diffusion-0}
		\hspace{-2.5mm}	\mu \le \frac{(1-\lambda)\nu}{(12 \hspace{-0.5mm}+\hspace{-0.5mm} 4e_1e_2 \hspace{-0.5mm}+\hspace{-0.5mm} \sqrt{6e_1e_2})(\delta^2 \hspace{-0.5mm}+ \hspace{-0.5mm} \beta_{\rm max}^2)}  \hspace{-0.5mm}= \hspace{-0.5mm} O\left( \frac{(1-\lambda)\nu}{\delta^2 \hspace{-0.8mm}+\hspace{-0.8mm} \beta_{\rm max}^2} \right) \hspace{-1.5mm}
	}
	where $\lambda = \max\{|\lambda_2(A)|, |\lambda_K(A)|\}$, $\beta^2_{\rm max}=\max_k\{\beta^2_k\}$, $e_1$ and $e_2$ are constants that are independent of $\lambda$, $\delta$, $\nu$ and $\beta$, then $\w_{k,i}$ generated by the diffusion recursions \eqref{diffusion-1}--\eqref{diffusion-2} converge exponentially fast to a neighborhood around $w^\star$. The convergence rate is $1-O(\mu\nu)$, and the size of the neighborhood can be characterized as follows
	\eq{\label{diffusion-neighborhood}
		&\hspace{-3mm}\limsup_{i\to \infty}\frac{1}{K}\sum_{k=1}^K \|\widetilde{\w}_{k,i}\|^2 \nnb
		&= O \left( \frac{\mu\sigma^2}{K \nu} + \frac{\delta^2}{\nu^2}\cdot\frac{\mu^2 \lambda^2 \sigma^2}{1-\lambda} + \frac{\delta^2}{\nu^2}\cdot \frac{\mu^2 \lambda^2 b^2}{(1-\lambda)^2} \right),
	}
	where $b^2 = (1/K)\sum_{k=1}^K \|\grad J_k(w^\star)\|^2$ is a bias term.
\end{lemma}
\noindent \textbf{Proof.} See Appendix \ref{appendix_diff_stability}. \qd

Comparing \eqref{ed-neighborhood} and \eqref{diffusion-neighborhood}, it is observed that the expressions for both algorithms consist of two major terms -- one $O(\mu)$ term and one $O(\mu^2)$ term. However, diffusion suffers from an additional bias term $O(\mu^2 \lambda^2 b^2/(1-\lambda)^2)$. 

{\color{black}
\begin{remark}[\sc Deterministic case]
When  $\sigma^2 = 0$, both diffusion and exact diffusion reduce to the deterministic scenario in which the real gradient $\grad J_k(w)$ is available. In this scenario, it is observed from \eqref{ed-neighborhood} and \eqref{diffusion-neighborhood} that the error $\tw_{k,i}$ in exact diffusion converges to $0$ while that in diffusion converges to $O(\mu^2 b^2)$, which is consistent with the results presented in \cite{chen2013distributed,yuan2016convergence,yuan2017exact1,yuan2017exact2}. \qd
\end{remark}

\begin{remark} [\sc Zero bias] \label{remark_towfic_result}
	When $b^2 = 0$, it holds that each local minimizer $w_k^\star$ coincides with the global minimizer $w^\star$, i.e., $w_k^\star = w^\star$ for any $k$. In this scenario, it is observed from \eqref{diffusion-neighborhood} that diffusion has the steady-state error bound 
	\eq{\label{diffusion-neighborhood-1}
	\limsup_{i\to \infty}\frac{1}{K}\sum_{k=1}^K \|\widetilde{\w}_{k,i}\|^2_{\rm d} = O \left( \frac{\mu\sigma^2}{K \nu} + \frac{\delta^2}{\nu^2}\cdot\frac{\mu^2 \lambda^2 \sigma^2}{1-\lambda} \right)
	}
	which is smaller than the error bound \eqref{ed-neighborhood} for exact diffusion especially when $\lambda$ approaches $0$. This result is consistent with \cite{towfic2015stability}, which finds diffusion outperforms primal-dual distributed adaptive methods when $w_k^\star = w^\star$ in terms of steady-state performance.  \qd
\end{remark}

\begin{remark} [\sc Large bias]
	When  $b^2$ is sufficiently large  so that the bias term (i.e., the third term) in \eqref{diffusion-neighborhood} dominates the entire error bound, it is observed from \eqref{ed-neighborhood} and \eqref{diffusion-neighborhood} that exact diffusion performs better than diffusion since it removes the bias term completely. This result is consistent with \cite{tang2018d}, which claims exact diffusion is endowed with faster convergence rate when the data variance across the network is large. \qd 
\end{remark}

In the following subsections, we will focus on the scenario where $\sigma^2 > 0$ and the bias $b^2$ is a small positive constant. In this scenario, we will study how the step-size $\mu$ and topology $\lambda$ influence the diffusion and exact diffusion algorithms. 

\subsection{Well-connected Network}\label{sec-comparison-good-network}
When the network is well-connected, it holds that $\lambda$ approaches $0$. For example, the fully-connected network has $\lambda=0$. In this scenario, the $O(\mu^2)$ terms inside diffusion's error bound will vanish and \eqref{diffusion-neighborhood} becomes
\eq{\label{diffusion-neighborhood-2}
\limsup_{i\to \infty}\frac{1}{K}\sum_{k=1}^K \|\widetilde{\w}_{k,i}\|^2_{\rm d}	= O \left( \frac{\mu\sigma^2}{K \nu} \right).
}
In comparison, the error bound \eqref{ed-neighborhood} for exact diffusion is
\eq{\label{ed-neighborhood-1}
	\hspace{-1mm}\limsup_{i\to \infty}\frac{1}{K}\sum_{k=1}^K\bE\|\widetilde{\w}_{k,i}\|^2_{\rm ed} = O\left(\frac{\mu\sigma^2}{K \nu} \hspace{-1mm}+\hspace{-1mm} \frac{\mu^2\delta^2\sigma^2}{ \nu^2} \right)
}
as $\lambda \rightarrow 0$. When $\mu$ is moderately small such that the term $O(\mu^2\delta^2\sigma^2/\nu^2)$ is non-negligible, we conclude that diffusion works better than exact diffusion. To roughly characterize the ``moderately'' small step-size, we assume $O(\mu^2\delta^2\sigma^2/\nu^2)$ is non-negligible if $\mu^2\delta^2\sigma^2/\nu^2 \ge \mu\sigma^2/(K\nu)$,
from which we get $\mu \ge \nu/(K\delta^2)$. Combining it with \eqref{ed-stability-range} we conclude that if $\mu$ satisfies (note that $\lambda\rightarrow 0$)
\eq{\label{23bsd8s8}
\frac{\nu}{K \delta^2} \le \mu \le \frac{d_1 \nu}{\delta^2 + \beta_{\rm max}^2}
}
where $d_1 = 1/(32+16c_1c_2+8\sqrt{c_1c_2})$, it holds that $O(\mu^2\delta^2\sigma^2/\nu^2)$ is non-trivial and diffusion has better steady-state performance than exact diffusion. To make the interval in \eqref{23bsd8s8} valid, it is enough to let $K$ be sufficiently large.

However, if the step-size $\mu$ is chosen sufficiently small, then the second term in \eqref{ed-neighborhood-1} is also negligible and hence both diffusion and exact diffusion will perform similarly. An example for ``sufficiently'' small step-size is when $\mu = \nu/(K^2\delta^2)$. By substituting $\mu = \nu/(K^2\delta^2)$ into \eqref{ed-neighborhood-1}, we reach $\limsup_{i\to \infty}\frac{1}{K}\sum_{k=1}^K\bE\|\widetilde{\w}_{k,i}\|^2_{\rm ed}=O(\frac{\sigma^2}{K^3 \delta^2} \hspace{-1mm}+\hspace{-1mm} \frac{\sigma^2}{K^4 \delta^2})=O(\frac{\sigma^2}{K^3 \delta^2})=O(\frac{\mu\sigma^2}{K\nu})$ in which the $O(\mu^2)$ term is negligible.

\subsection{Sparsely-connected Network}\label{sec-comparison-bad-network}
When the network is sparsely-connected, it holds that $\lambda$ approaches $1$. In this scenario, even a trivial bias constant $b^2$ can be significantly amplified by the coefficient $1/(1-\lambda)^2$. 
When $\lambda$ approaches $1$, the first two terms in \eqref{diffusion-neighborhood} will be the same as those in \eqref{ed-neighborhood}. As a result, when $\mu$ is moderately small and $\lambda$ is close to $1$ such that the bias term $O({\mu^2 \delta^2 \lambda^2 b^2}/{(1-\lambda)^2\nu^2})$ is non-negligible, we conclude that exact diffusion works better than diffusion. Furthermore, the advantage of exact diffusion will be more evident if the bias gets more significant as $\lambda \rightarrow 1$. In the following example, we list several network topologies in which the bias $O(\mu^2 b^2/(1-\lambda)^2)$ dominates \eqref{d-steady-state} easily.

\noindent \textbf{Example (Linear, Cyclic, and Grid networks).} A linear or cyclic network with $K$ agents is a network where each agent connects with its previous and next neighbors. On the other hand, a grid network with $K$ agents is a network in which each node connects with its neighbors from left, right, top, and bottom. The grid and cycle networks are illustrated in Fig.\ref{fig:topologies}. For these networks, it is shown in \cite{mao2018walkman,seaman2017optimal} that 
\eq{
	1 - \lambda &= O(1/K^2) \quad \text{ (linear or cyclic network)} \\
	1-\lambda &= O(1/K) \quad \text{\hspace{1.6mm} (grid network)}
}
and therefore, the bias term $O(\mu^2b^2/(1-\lambda)^2)$ in diffusion over linear (or cyclic) graph and grid graph becomes $O(\mu^2 b^2 K^4)$ and $O(\mu^2 b^2 K^2)$ respectively, which increases rapidly with the size of the network. As a result, exact diffusion, by correcting the bias term, is evidently superior to diffusion over these network topologies.  \qd
\begin{figure}[h!]
	\centering
	\includegraphics[scale=0.3]{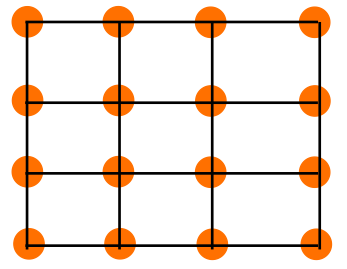}
	\quad\quad\quad
	\includegraphics[scale=0.3]{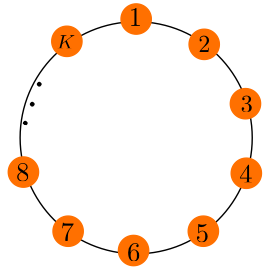}
	\vspace{-3mm}
	\caption{Illustration of the grid topology and cyclic topology.}
	\label{fig:topologies}
\end{figure}

To roughly characterize the ``moderately'' small step-size, we assume $O({\mu^2 \delta^2 \lambda^2 b^2}/{(1-\lambda)^2\nu^2})$ is non-trivial if 
\eq{
	\frac{\delta^2}{\nu^2}\cdot \frac{\mu^2 b^2}{(1-\lambda)^2} \ge \frac{\mu \sigma^2}{K \nu}	
}
from which we get $\mu \ge {(1-\lambda)^2\sigma^2\nu}/{K \delta^2 b^2}.$ Combining it with  \eqref{mu_range_diffusion-0}, we conclude that if $\mu$ satisfies 
\eq{\label{23bsd9-0}
	\frac{(1-\lambda)^2\sigma^2\nu}{K \delta^2 b^2} \le \mu \le \frac{d_2(1-\lambda)\nu}{\delta^2 + \beta_{\rm max}^2},
}
where $d_2 = 12 + 4e_1 e_2 + \sqrt{6e_1 e_2}$ is a constant, then the bias term in \eqref{diffusion-neighborhood} is significant and exact diffusion is expected to have better performance than diffusion in steady-state. To make the interval in \eqref{23bsd9-0} valid, it is enough to let $\lambda$ be sufficiently close to $1$ and $K$ be sufficiently large such that
\eq{\label{ed-better-condition-0}
	\frac{(1\hspace{-0.8mm}-\hspace{-0.8mm}\lambda)^2\sigma^2\nu}{K\delta^2 b^2} \hspace{-0.8mm}<\hspace{-0.8mm} \frac{d_2(1-\lambda)\nu}{\delta^2 + \beta^2} \Longleftrightarrow \frac{b^2}{1-\lambda} > \frac{(\delta^2+\beta^2)}{d_2 K \delta^2} \sigma^2.	
}

On the other hand, if we adjust $\mu$ to be sufficiently small, the $O(\mu)$ term in both expressions \eqref{ed-neighborhood} and \eqref{diffusion-neighborhood} will eventually dominate for any fixed $b^2$ and $\lambda$. In such scenario, it holds that
	\eq{
		\limsup_{i\to \infty}\frac{1}{K}\sum_{k=1}^K\bE\|\widetilde{\w}_{k,i}\|^2_{\mathrm{ed}} & = O\Big(\frac{\mu\sigma^2}{K \nu}\Big), \label{7sgh-1}\\
		\limsup_{i\to \infty}\frac{1}{K}\sum_{k=1}^K\bE\|\widetilde{\w}_{k,i}\|^2_{\mathrm{d}} &= O\Big(\frac{\mu\sigma^2}{K\nu}\Big). \label{7sgh-2}
	}
	It is observed that both diffusion and exact diffusion will have the same mean-square error order, which implies that diffusion and exact diffusion will perform similarly in this scenario. Such ``sufficiently'' small step-size can be roughly characterized by the range
	\eq{\label{sufficiently-small-mu}
		\mu \le d_3 (1-\lambda)^{2+x}\quad \mbox{where} \quad x>0.
	}
	for some $d_3>0$. The comparison between exact diffusion and diffusion is listed in Table \ref{table-ncomparison}.}

\section{Mean-square Deviation Expression}
\label{sec-MSD}
In the last section, we showed that when  $\mu$ is sufficiently small, the steady-state mean-square deviation of both diffusion and exact diffusion will be dominated by a term on the order of $O(\mu \sigma^2 /\nu)$, as illustrated by \eqref{7sgh-1}--\eqref{7sgh-2}. However, the hidden constants inside the big-$O$ notation are still unclear. In this section, we show that, when $\mu$ is approaching $0$, i.e., $\mu \to 0$, diffusion and exact diffusion will have exactly the same MSD expression in steady state. To this end, we recall the definition of mean-square deviation (MSD) from \cite{sayed2014adaptation} as follows:
\eq{
\label{msd-definition}
\mbox{MSD} = \mu \Big(\lim_{\mu \to 0} \limsup_{i\to \infty} \frac{1}{\mu K} \sum_{k=1}^K \bE\|\tw_{k,i}\|^2 \Big).
}
Note that the MSD defined above is precise to the first-order in the step-size. All higher order terms are ignored.

\subsection{Approximate Error Dynamics}
It is generally difficult to derive the MSD performance of exact diffusion with the original transformed error dynamics developed in Lemma \ref{lm-transform-error-dyanmic}. We therefore propose an approximate error dynamics and employ it to assess the MSD performance. To this end, we define
\eq{
H_k \hspace{-1mm} = \hspace{-1mm} \grad^2 J_k(w^\star),\ \cH \hspace{-1mm}=\hspace{-1mm} \diag\{H_1,\cdots, H_K\},\ \cT \hspace{-1mm} = \hspace{-1mm}
\ba{cc}
\hspace{-1mm}\cH & 0 \\
\hspace{-1mm}0 & 0
\ea \hspace{-1mm}.
}
Obviously, it holds that $\H_{k,i}\to H$, ${\boldsymbol \cH}_i \to \cH$ and ${\boldsymbol \cT}_i \to \cT$ if $\swb_i \to \sw^\star$. Next, we consider the approximate error dynamic as follows.
\eq{
	\label{approximate-error-dynamics}
	\ba{c}
	\bar{\szb}_i' \\
	\check{\szb}_i'
	\ea	
	\hspace{-1mm}&=\hspace{-1mm} 
	\ba{cc}
	\hspace{-1mm}I_M \hspace{-1mm}-\hspace{-1mm} \frac{\mu}{K}\sum_{k=1}^K H_k & -\frac{c\mu}{K}\cI\tran \cH \cX_{R,u}\hspace{-1mm}\\
	\hspace{-1mm}-\frac{\mu}{c} \cD_1 \cX_L \cT \cR_1 & \cD_1 - \mu \cD_1 \cX_L \cT \cX_R\hspace{-1mm}
	\ea 
	\ba{c}
	\bar{\szb}'_{i-1} \\
	\check{\szb}'_{i-1}
	\ea	\nnb
	& \quad + \mu 
	\ba{c}
	\frac{1}{K}\cI\tran\\
	\frac{1}{c}\cD_1\cX_L \cB_\ell
	\ea \s_i(\swb_{i-1}).
}
Note that we replace $\H_{k,i-1}$, ${\boldsymbol \cH}_{i-1}$ and ${\boldsymbol \cT}_{i-1}$ in \eqref{transformed-error-dynamics} with $H_k$, $\cH$ and $\cT$ in \eqref{approximate-error-dynamics}. We can show that the iterates $\bar{\szb}_i'$ and $\check{\szb}_i'$ generated through the approximate error dynamic \eqref{approximate-error-dynamics} are close to $\bar{\szb}_i$ and $\check{\szb}_i$ generated from the original recursion \eqref{transformed-error-dynamics} -- see Lemma \ref{lm-approx-error} below. This implies that we can employ recursion \eqref{approximate-error-dynamics} rather than \eqref{transformed-error-dynamics} to establish the MSD performance. To this end, we first introduce a few more assumptions on cost functions and the gradient noise. These assumptions are adapted from \cite{sayed2014adaptation}.

\begin{assumption}[\sc Smoothness condition in the limit]
	\label{ass-Hessian} 
	For each cost function $J_k(w)$, it is assumed that
	\eq{\label{Hessian-local-cost}
		\|\grad^2 J_k(w^\star + \Delta w) - \grad^2 J_k(w^\star)\| \le \kappa \|\Delta w\|
	}
	for small perturbations $\|\Delta w\| \le \epsilon$, where $\kappa > 0$ is a constant.
	\rightline	\qd
\end{assumption}

\begin{assumption}[\sc Forth-Order Moment]
	\label{ass-grad-noise-more} 
	It is assumed for each $k$ and $i$ that
	\eq{
		\bE[\|\s_{k,i}(\w_{k, i-1})\|^4|\filt_{i-1}] \le \beta^4_{4,k} \|\tw_{k,i-1}\|^4 \hspace{-1mm}+\hspace{-1mm} \sigma_{4,k}^4.
	}
	where $\beta_{4,k}$ and $\sigma_{4,k}$ are some positive constants.
	\qd
\end{assumption}

By following the proof of Theorem 10.2 from \cite{sayed2014adaptation}, we can prove in the following lemma that difference between the original iterates \eqref{transformed-error-dynamics} and the transformed iterates \eqref{approximate-error-dynamics} is small.
\begin{lemma}[\sc Approximation Error]\label{lm-approx-error} Under Assumptions \ref{ass-lip}--\ref{ass-grad-noise-more}, it holds for sufficiently small step-sizes that
\eq{\label{z-minus-z'}
\limsup_{i\to \infty} \bE\left\| 
\ba{c}
\bar{\szb}_i \\
\check{\szb}_i
\ea
- 
\ba{c}
\bar{\szb}'_i \\
\check{\szb}'_i
\ea
 \right\|^2  = O(\mu^2)
}
\rightline \qd
\end{lemma}

\subsection{Deriving the MSD expression}
Recall from \eqref{w-y-vs-bz-cz} that 
\eq{
\twb_i = 
\ba{cc}
\cI & c \cX_{R, u}
\ea
\ba{c}
\bar{\szb}_i \\
\check{\szb}_i
\ea.
}
This together with $\cI\tran \cX_{R,u} = 0$\footnote{Since $\cX^{-1} \cX = I$ with $\cX$ and $\cX^{-1}$ defined in \eqref{cB-decom}, we have $c \cL_1\tran \cX_R = \frac{c}{K}\cI\tran \cX_{R,u}=0.$ } implies that 
\eq{\label{bsdkw836dg}
\|\twb_i\|^2 &= 
\ba{c}
\hspace{-1mm}\bar{\szb}_i \hspace{-1mm}\\
\hspace{-1mm}\check{\szb}_i\hspace{-1mm}
\ea\tran
\underbrace{\ba{cc}
K I_{KM} & 0 \\
0 & c^2 \cX_{R,u}\tran \cX_{R,u}
\ea}_{\define \Gamma}
\ba{c}
\hspace{-1mm}\bar{\szb}_i \hspace{-1mm}\\
\hspace{-1mm}\check{\szb}_i\hspace{-1mm}
\ea 
}
For simplicity, in the following we let 
\eq{\label{z and z'}
\szb_i = 
\ba{c}
\bar{\szb}_i\\
\check{\szb}_i
\ea,\quad
\szb_i' = 
\ba{c}
\bar{\szb}'_i\\
\check{\szb}'_i
\ea.
}
and it holds that $\bE\|\twb_i\|^2=\bE\|\szb_i\|_{\Gamma}^2$.
The following lemma shows that $\bE\|\szb_i\|^2_{\Gamma}$ is close to $\bE\|\szb'_i\|^2_{\Gamma}$.
\begin{lemma}[\sc Approximation Scaled Error] \label{lm-approximation-error}
	Under Assumptions \ref{ass-lip}--\ref{ass-grad-noise-more}, it holds for sufficiently small step-sizes that
	\eq{\label{2378sdbsd}
		\limsup_{i\to \infty}\ \bE\|\szb_i\|^2_\Gamma - \bE\|\szb_i'\|^2_\Gamma  = O(\mu^{3/2})
	}
	\rightline \qd
\end{lemma}
\noindent \textbf{Proof.} It holds that
\eq{
\bE\|\szb_i'\|^2_\Gamma &= \bE\|\szb_i' - \szb_i + \szb_i\|^2_\Gamma \nnb
&\le \bE\|\szb_i' - \szb_i\|^2_\Gamma  + \bE\|\szb_i\|^2_\Gamma + 2 \bE[(\szb_i' - \szb_i)\tran\Gamma\szb_i] \nnb
&\le \bE\|\szb_i' \hspace{-1mm}-\hspace{-1mm} \szb_i\|^2_\Gamma  \hspace{-1mm}+\hspace{-1mm} \bE\|\szb_i\|^2_\Gamma \hspace{-1mm}+\hspace{-1mm} 2 \sqrt{\bE\|\szb_i' - \szb_i\|^2_\Gamma \bE\|\szb_i\|^2_\Gamma}, \nonumber
}
which implies that
\eq{
&\ \bE\|\szb_i'\|^2_\Gamma - \bE\|\szb_i\|^2_\Gamma \nnb
\le&\ \bE\|\szb_i' \hspace{-1mm}-\hspace{-1mm} \szb_i\|^2_\Gamma + 2 \sqrt{\bE\|\szb_i' - \szb_i\|^2_\Gamma \bE\|\szb_i\|^2_\Gamma} \nnb
\le&\ \lambda_{\max}(\Gamma) \bE\|\szb_i' \hspace{-1mm}-\hspace{-1mm} \szb_i\|^2 + 2\lambda_{\max}(\Gamma) \sqrt{\bE\|\szb_i' \hspace{-1mm}-\hspace{-1mm} \szb_i\|^2 \bE\|\szb_i\|^2} \nn
}
\begin{figure*}[t]
	\centering
	\includegraphics[scale=0.33]{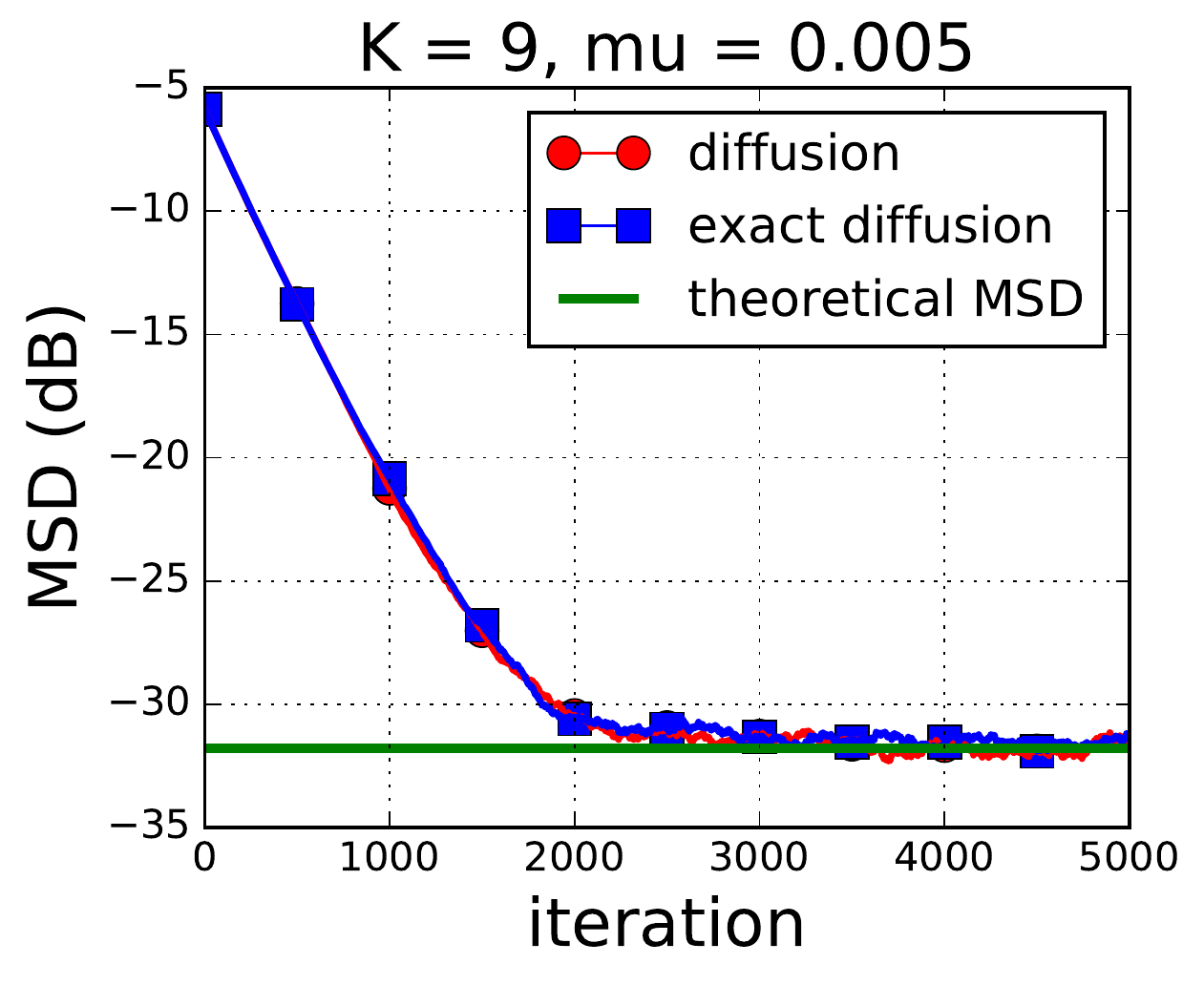}
	\includegraphics[scale=0.33]{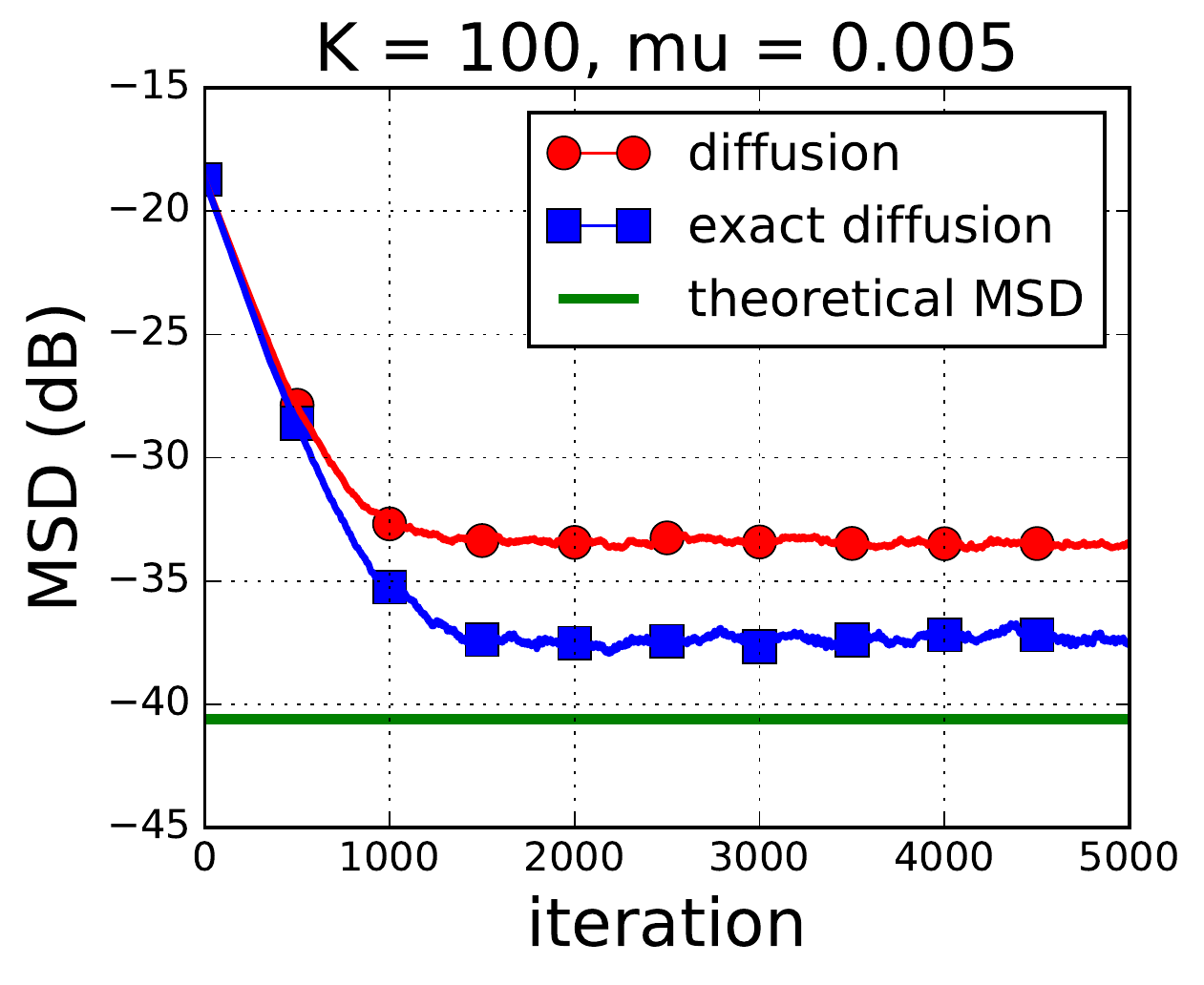}
	\includegraphics[scale=0.33]{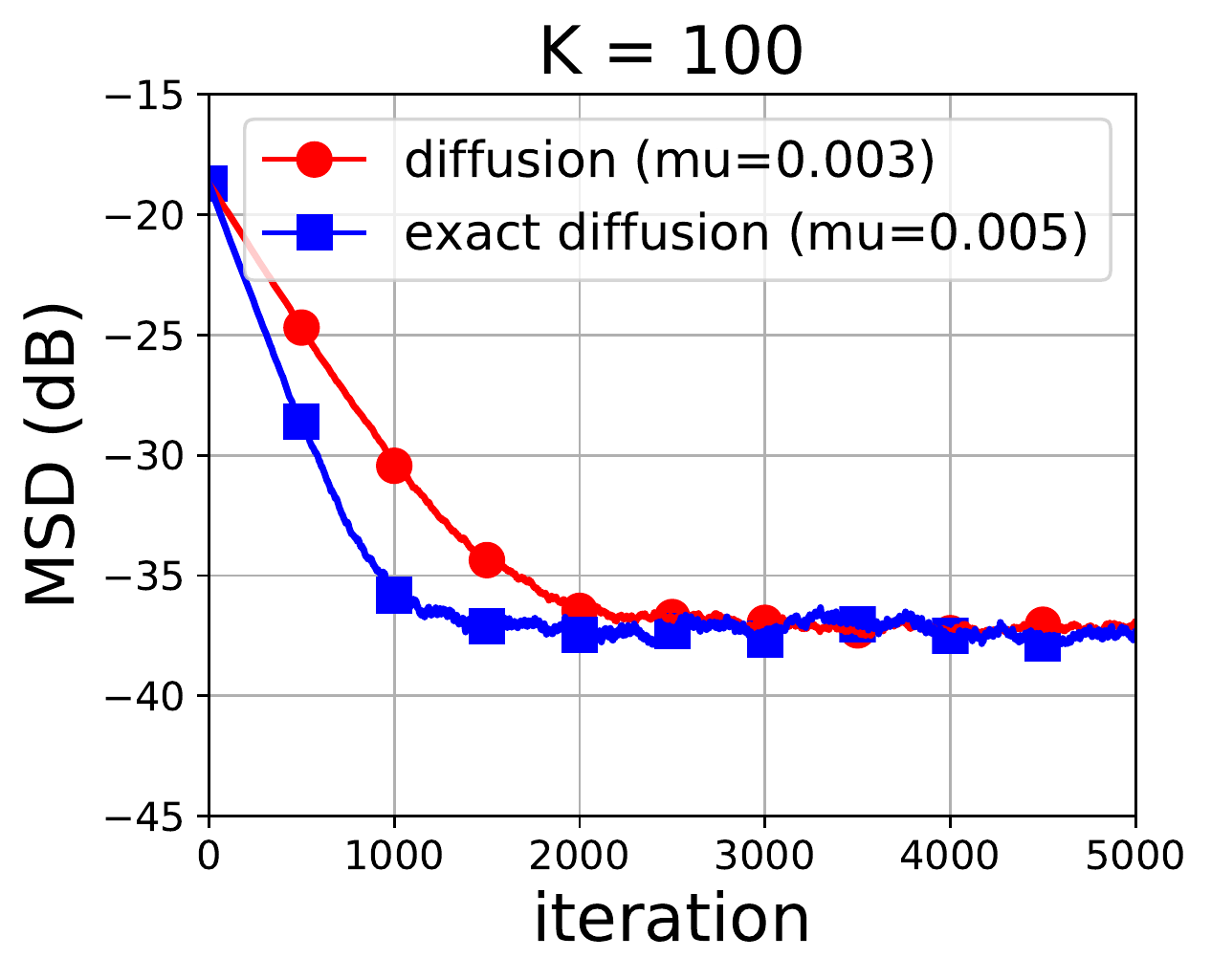}
	\includegraphics[scale=0.33]{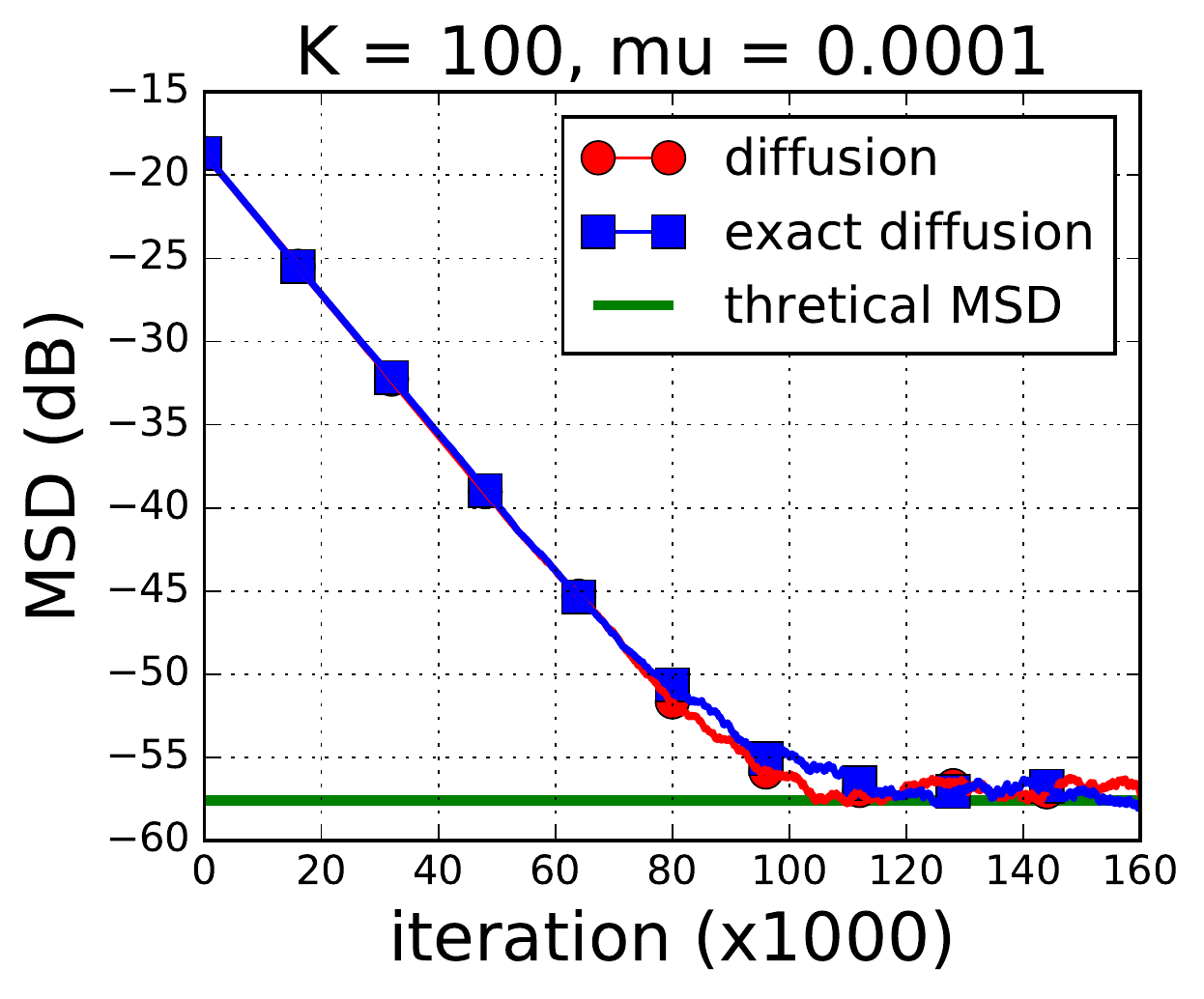}
	\vspace{-3mm}
	\caption{Diffusion v.s. exact diffusion over grid networks for problem \eqref{mse-network}.}
	\label{fig:grid-change-size-ls}
\end{figure*}

\begin{figure*}[t]
	\centering
	\includegraphics[scale=0.33]{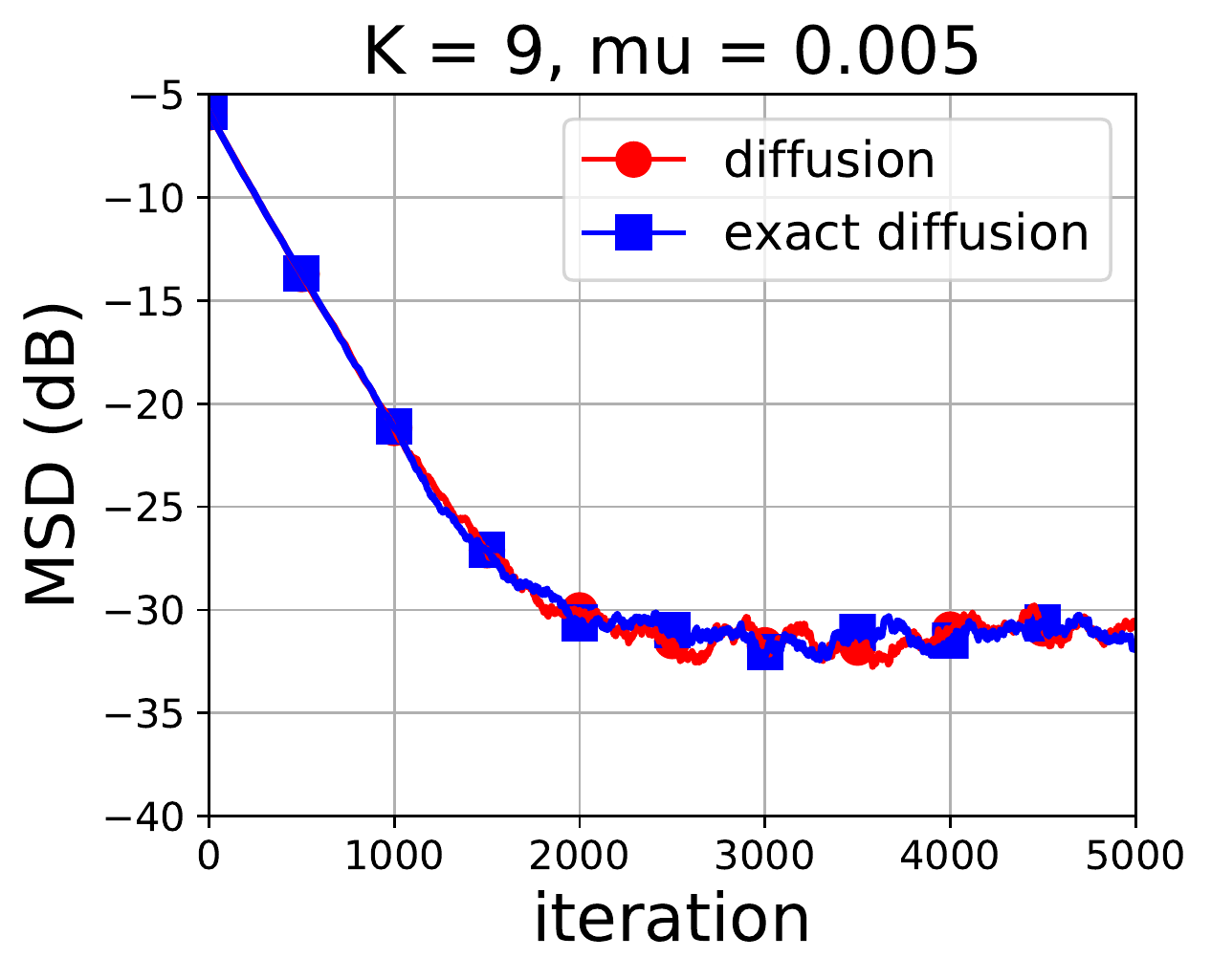}
	\includegraphics[scale=0.33]{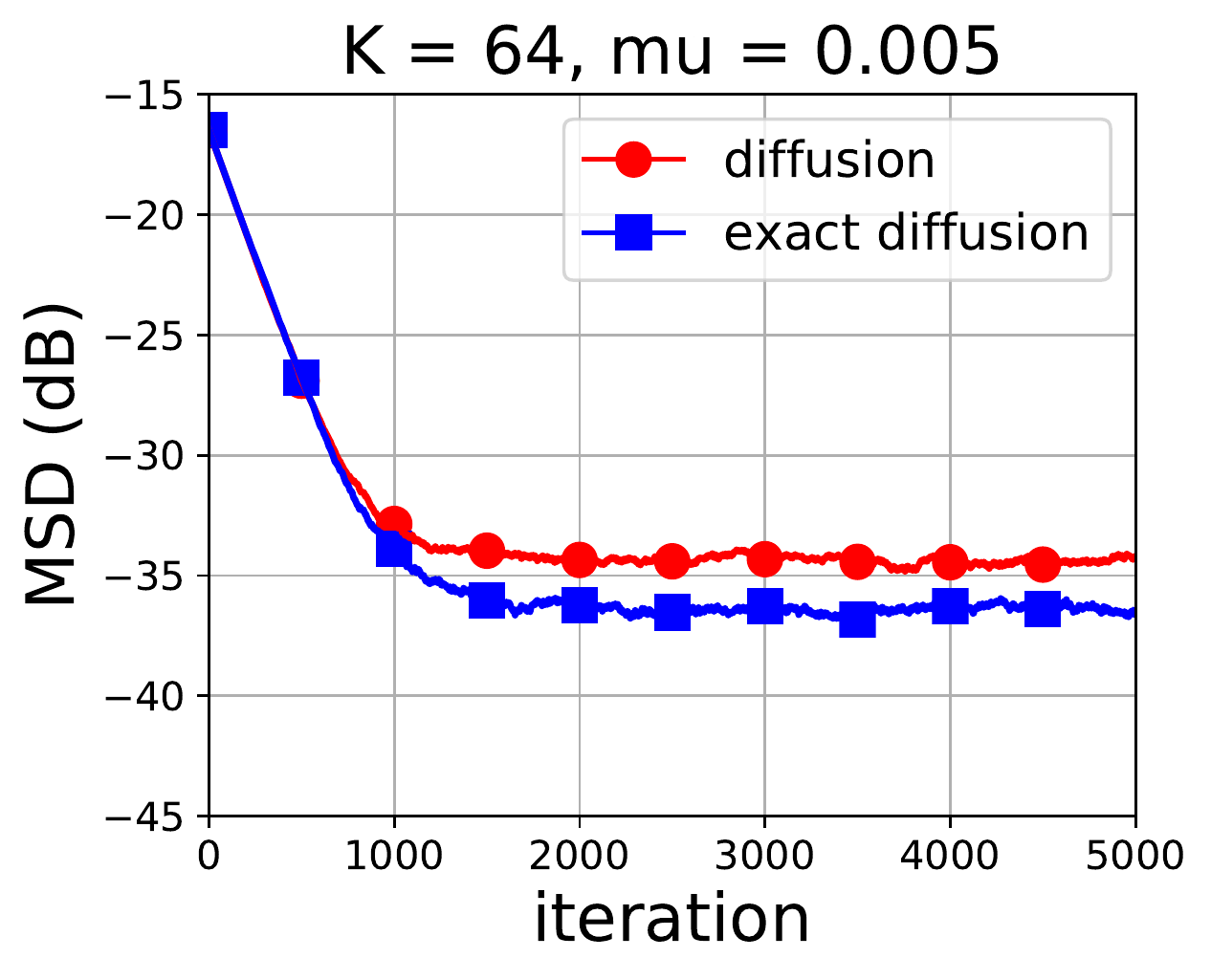}
	\includegraphics[scale=0.33]{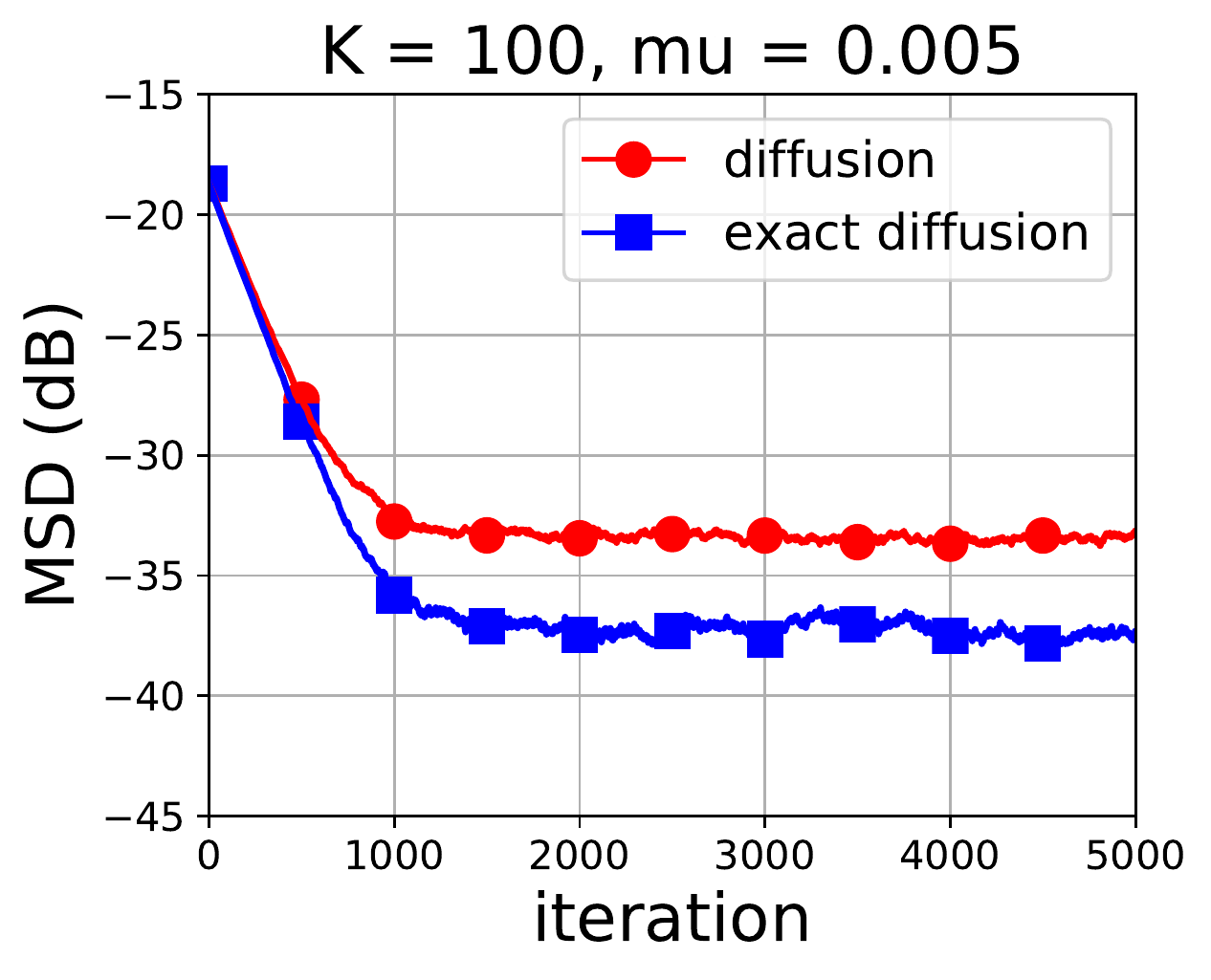}
	\includegraphics[scale=0.33]{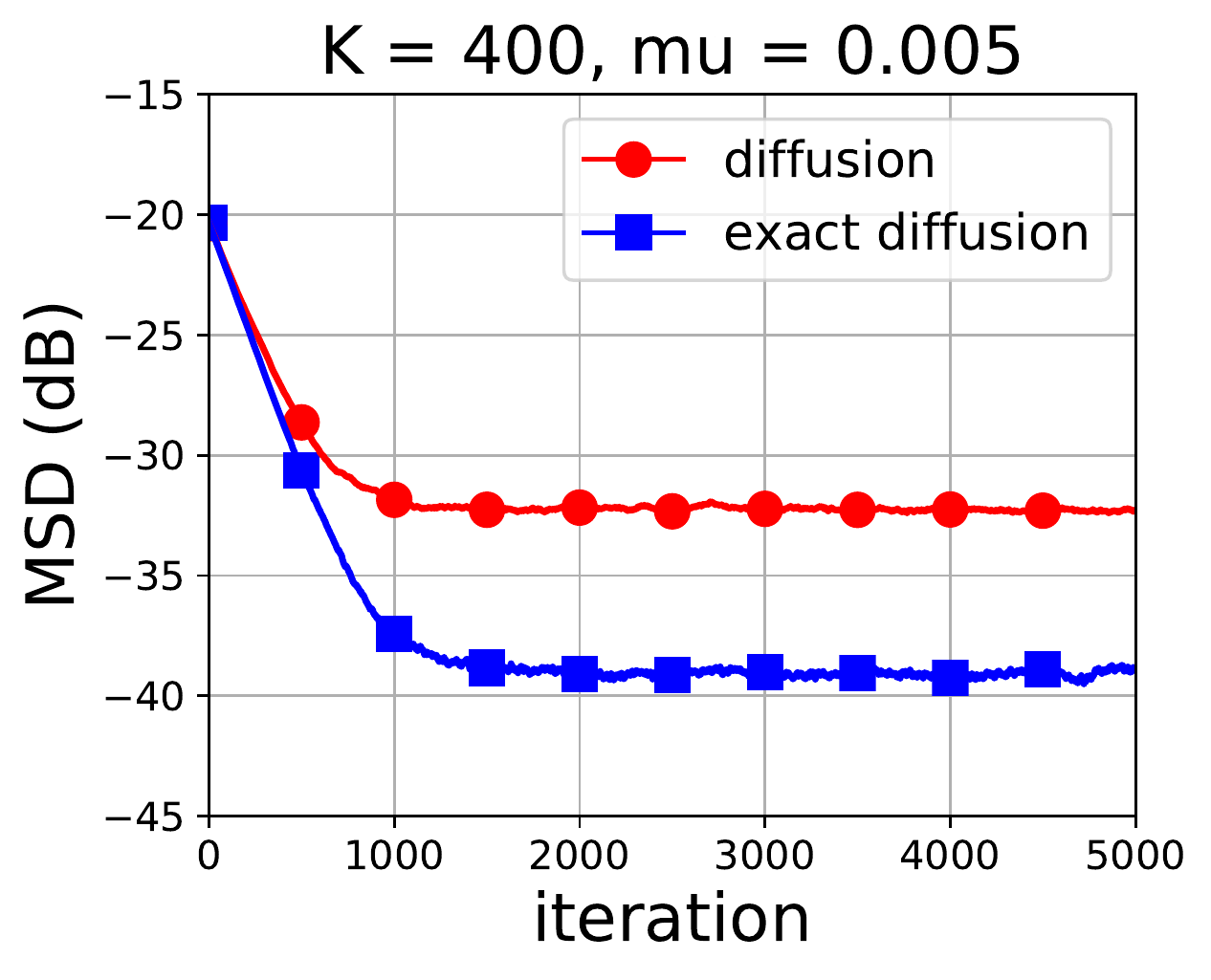}
	\vspace{-3mm}
	\caption{The superiority of exact diffusion is more evident as the grid network becomes larger when solving problem \eqref{mse-network}.}
	\label{fig:grid-change-size-ls-2}
\end{figure*}
where $\lambda_{\max}(\Gamma)$ is the largest eigenvalue of $\Gamma$. From \eqref{z and z'} we know it holds for sufficiently small $\mu$ that 
\eq{
\limsup_{i\to\infty}\bE\|\szb_i\|^2 &= \limsup_{i\to\infty}\bE\|\bar{\szb}_i\|^2 + \limsup_{i\to\infty}\bE\|\check{\szb}_i\|^2 \nnb
& \overset{\eqref{z-steady-state}}{=} O(\mu) + O(\mu^2) = O(\mu).
}
Also, from \eqref{z-minus-z'} we have 
\eq{
\limsup_{i\to\infty}\bE\|\szb_i' - \szb_i\|^2 = O(\mu^2).
}
Since $\Gamma$ is independent of $\mu$, it therefore holds that
\eq{
\limsup_{i\to\infty} \left(\bE\|\szb_i'\|^2_\Gamma - \bE\|\szb_i\|^2_\Gamma \right) = O(\mu^{3/2}).
}
\qd

Now we establish the MSD expression for exact diffusion. Since  $\bE\|\twb_i\|^2=\bE\|\szb_i\|_{\Gamma}^2$ is close to $\bE\|\szb_i^\prime\|_{\Gamma}^2$ as proved in Lemma \ref{lm-approximation-error}, we will first derive the MSD expression for $\bE\|\szb_i^\prime\|_{\Gamma}^2$ and use it to facilitate the derivation of the MSD for exact diffusion, i.e., $\bE\|\twb_i\|^2$. To proceed, we assume that, in the limit, the following covariance matrix evaluated at the global solution $w^\star$ exists
\eq{
S_k \define \lim_{i\to \infty} \bE[\s_{k,i}(w^\star) \s_{k,i}(w^\star)\tran]. \label{S_k_noise_limit}
}
The following theorem establishes the MSD expression of the approximate error dynamics.

\begin{theorem}[\sc MSD expression] \label{Thm2}
	Under Assumptions \ref{ass-lip}--\ref{ass-grad-noise-more}, it holds for exact diffusion that
	\eq{\label{MSD-expression-exact-diffusion}
	{\rm MSD}_{\rm ed} = 	\frac{\mu}{2K} \Tr\left\{\left(\sum_{k=1}^K H_k\right)^{-1}\left(\sum_{k=1}^K S_k\right)\right\}.
	}
\end{theorem}
\noindent \textbf{Proof.} See Appendix \ref{appendix_MSD_proof}. \qd 

Recall the MSD expression for standard diffusion is \cite[Equation (11.140)]{sayed2014adaptation}: 
\eq{\label{MSD-expression-diffusion}
	{\rm MSD}_{\rm d} = 	\frac{\mu}{2K} \Tr\left\{\left(\sum_{k=1}^K H_k\right)^{-1}\left(\sum_{k=1}^K S_k\right)\right\}.
}
It is observed that the MSD expression for diffusion \eqref{MSD-expression-diffusion} is equal to that of exact diffusion \eqref{MSD-expression-exact-diffusion}. This implies that diffusion and exact diffusion will perform exactly the same in steady state for sufficiently small step-sizes.

\section{Numerical simulation}

%

{\color{black}
\subsection{Mean-square-error Network}
In this subsection we consider the scenario in which $K$ agents observe streaming data $\{\d_k(i),\u_{k,i}\}$ that satisfy the regression model
\eq{\label{dki}
\d_k(i) = \u_{k,i}\tran w_k^\star + \v_k(i)	
}
where $w_k^\star$ is the local optimal solution at agent $k$, and the noise process, $\v_k(i)$, is independent of the regression data, $\u_{k,i}$. The cost over the mean-square-error (MSE) network is defined by
\eq{\label{mse-network}
\min_{w\in \RR^M}\quad \sum_{k=1}^K \bE\big(\d_k(i) - \u_{k,i}\tran w\big)^2.
}
To generate $\{\d_k(i),\u_{k,i}\}$, we first generate the local optimal solution following a standard Gaussian distribution, i.e., $w_k^\star \sim \cN(0, I_M)$. Next we generate $\u_{k,i}\sim \cN(0, \Lambda_k)$ where $\Lambda_k$ is a positive diagonal matrix and $\v_k(i)\sim \cN(0, 0.1 I_M)$. With $w_k^\star,\ \u_{k,i}$ and $\v_k(i)$, we generate $\d_k(i)$ according to \eqref{dki}. Also, we can verify that the global solution to \eqref{mse-network} is given by 
\eq{
w^\star = \left(\sum_{k=1}^K \Lambda_k\right)^{-1}\sum_{k=1}^{K}\Lambda_k w_k^\star.
}
In all figures below, the $y$-axis indicates the MSD performance $\sum_{k=1}^{K}\bE\|\w_{k,i}-w^\star\|^2/K$.

\begin{figure}[b!]
	\centering
	\includegraphics[scale=0.33]{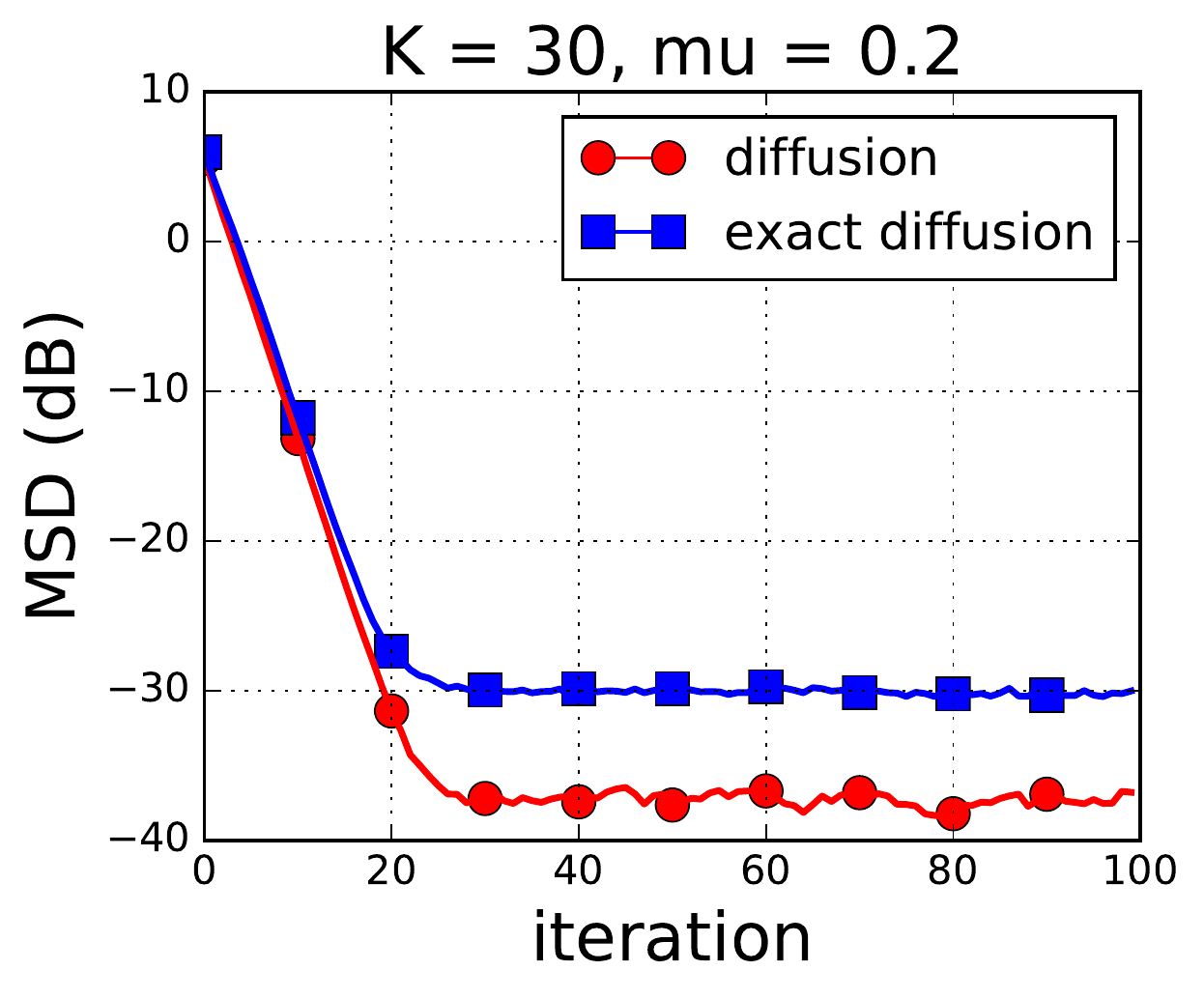}
	\includegraphics[scale=0.33]{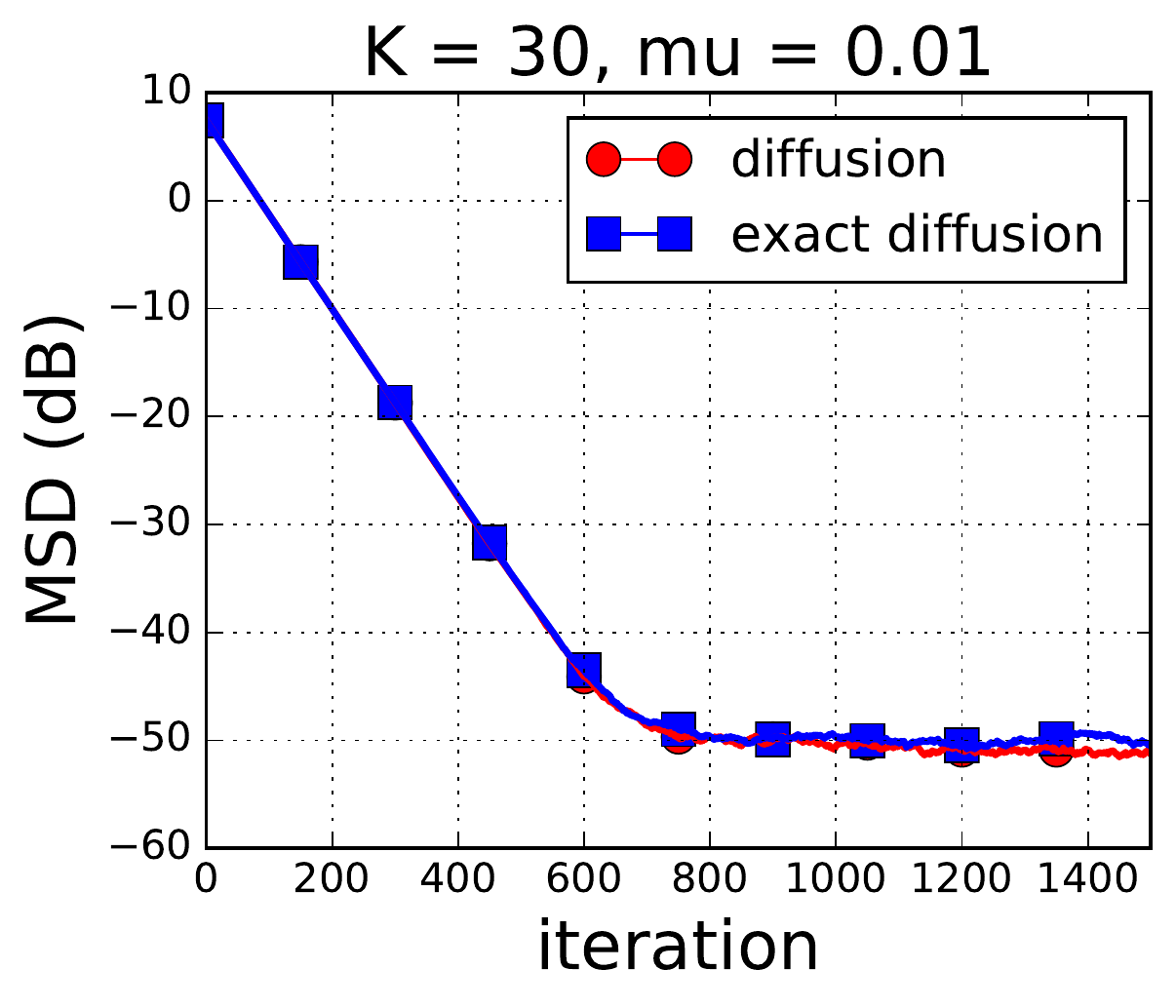}
	\vspace{-3mm}
	\caption{Diffusion v.s. exact diffusion over a fully connected network for problem \eqref{mse-network}.}
	\label{fig:fullyconnected-ls}
\end{figure}

\begin{figure*}[t!]
	\centering
	\includegraphics[scale=0.33]{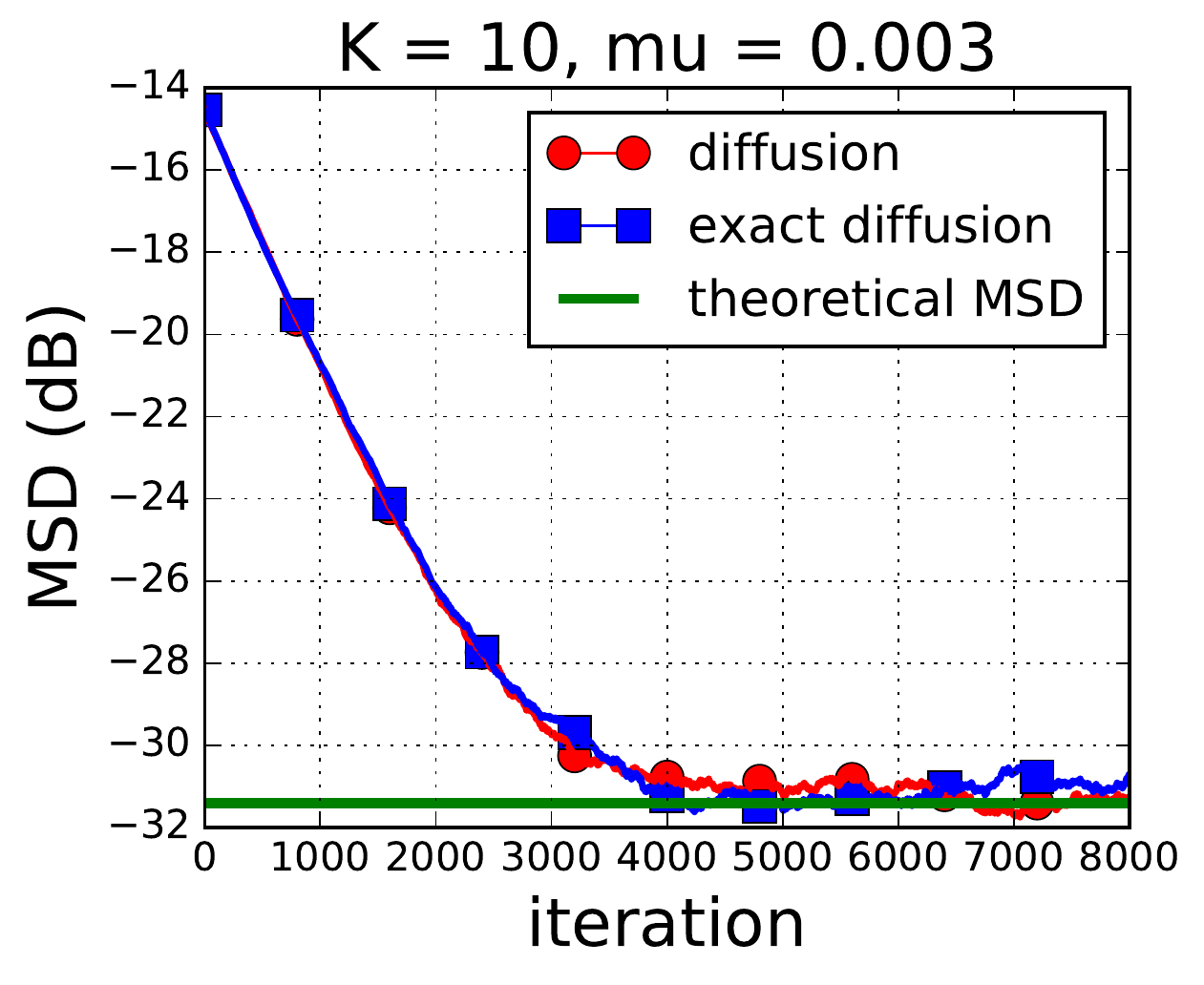}
	\includegraphics[scale=0.33]{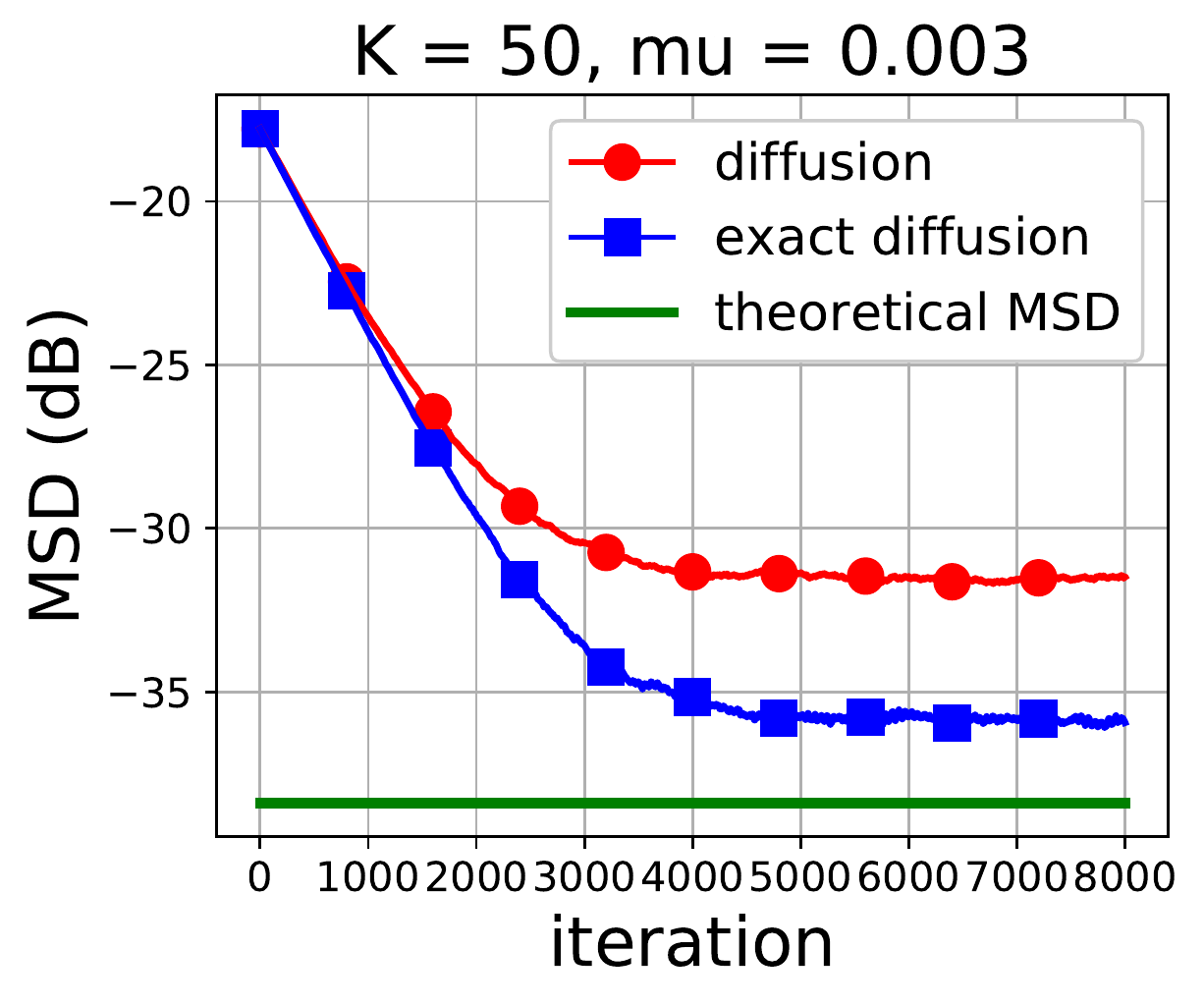}
	\includegraphics[scale=0.33]{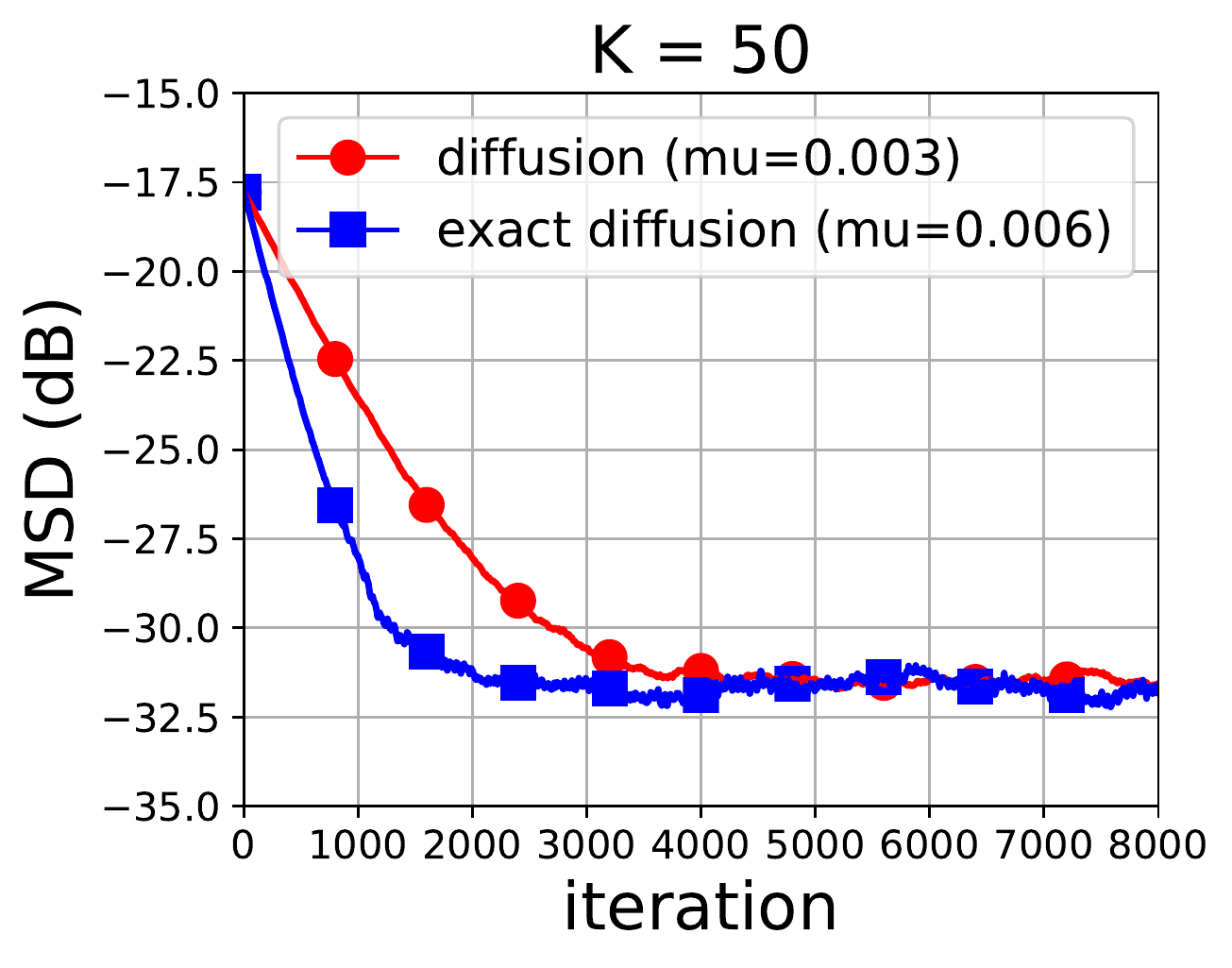}
	\includegraphics[scale=0.33]{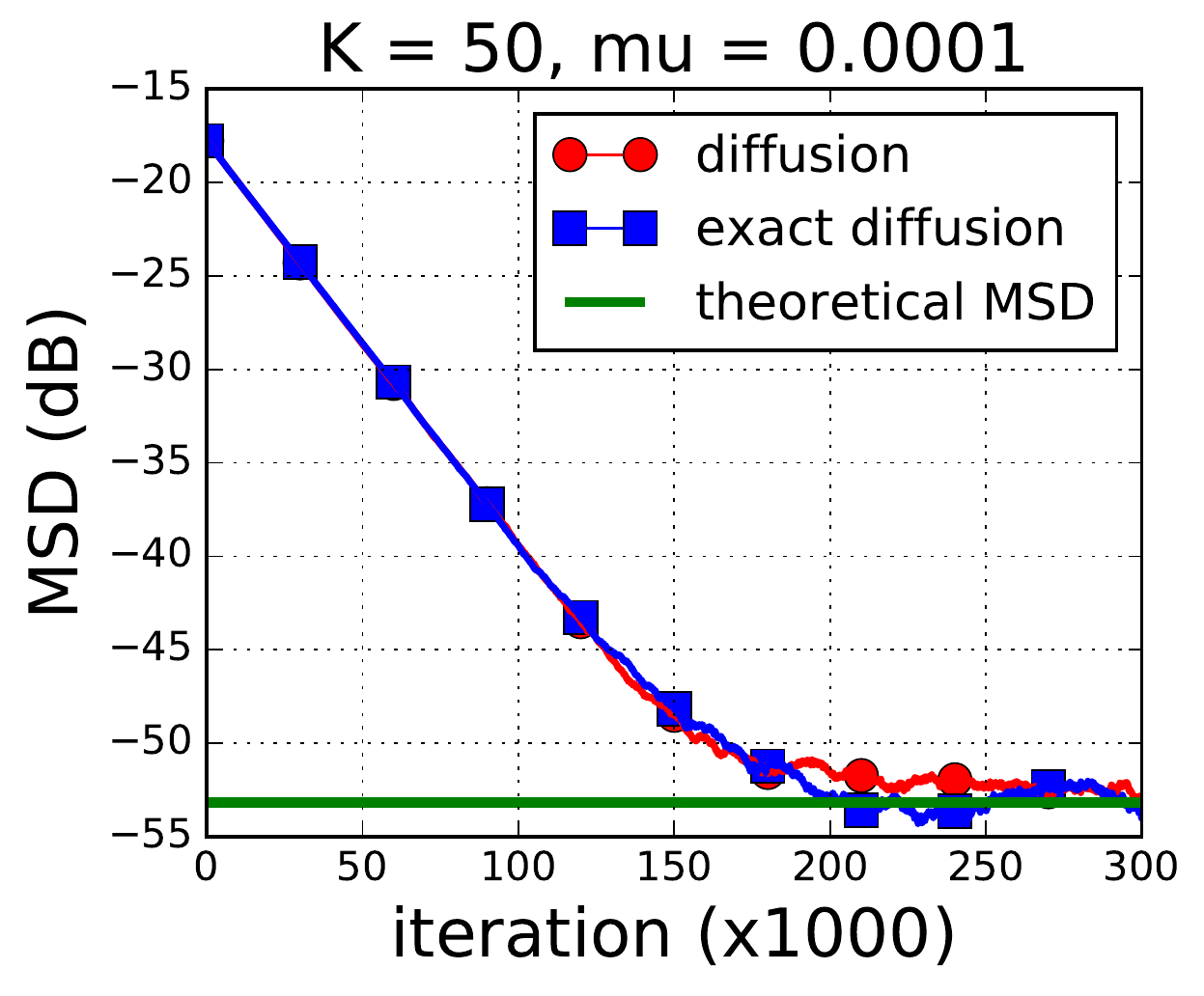}
	\vspace{-3mm}
	\caption{Diffusion v.s. exact diffusion over cyclic networks for problem \ref{dist_log_regression}.}
	\label{fig:cyclic-change-size}
\end{figure*}

\begin{figure*}[t!]
	\centering
	\includegraphics[scale=0.33]{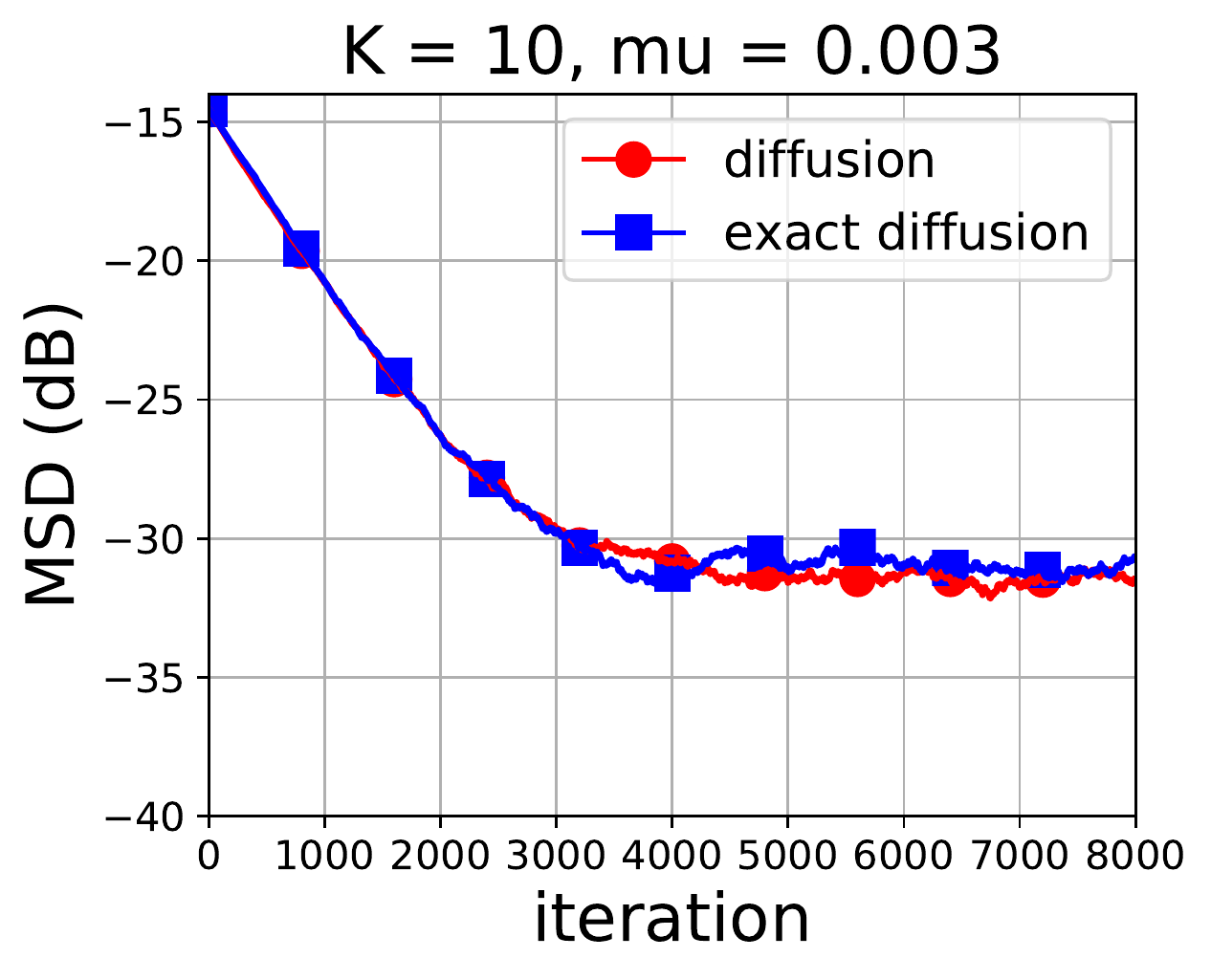}
	\includegraphics[scale=0.33]{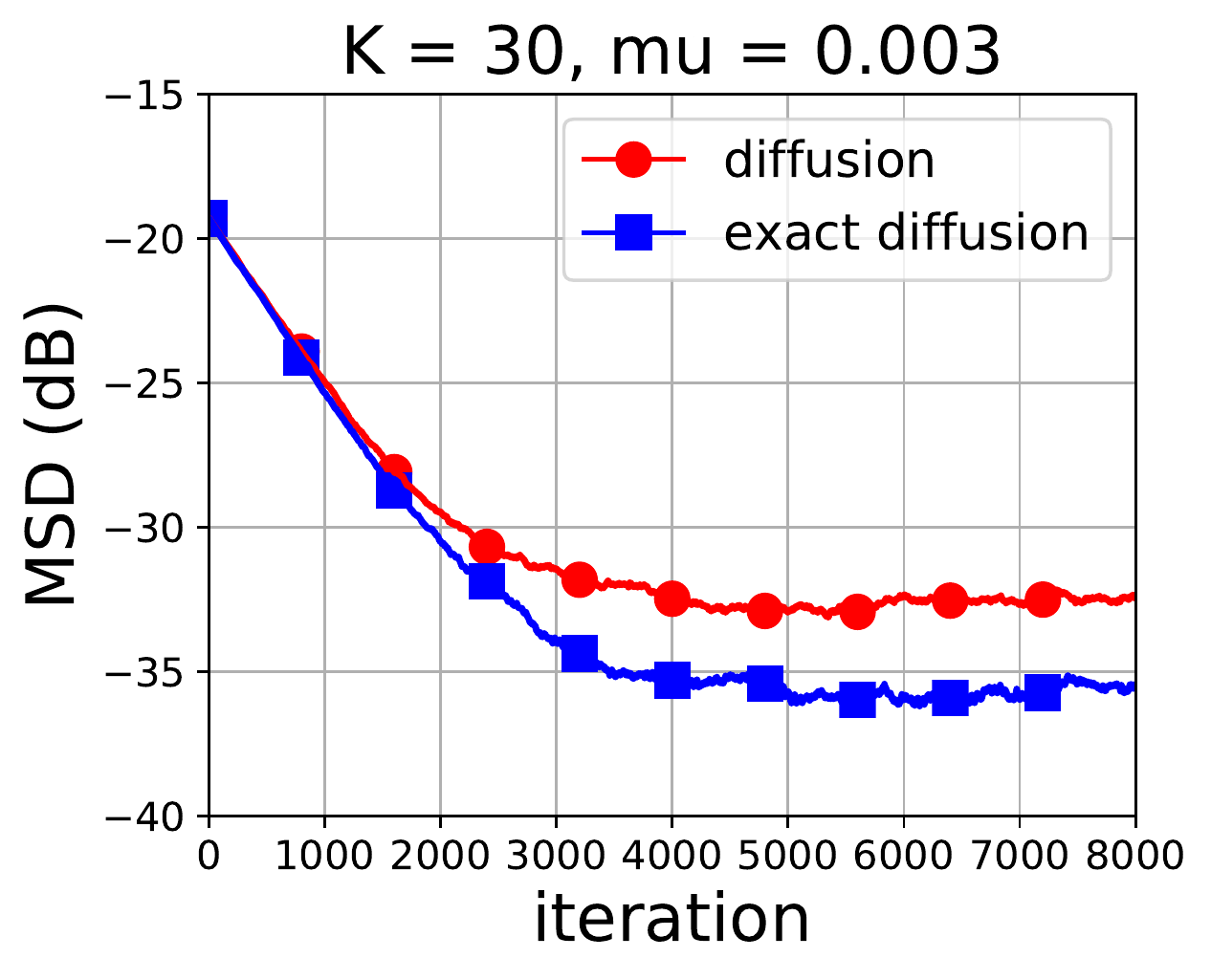}
	\includegraphics[scale=0.33]{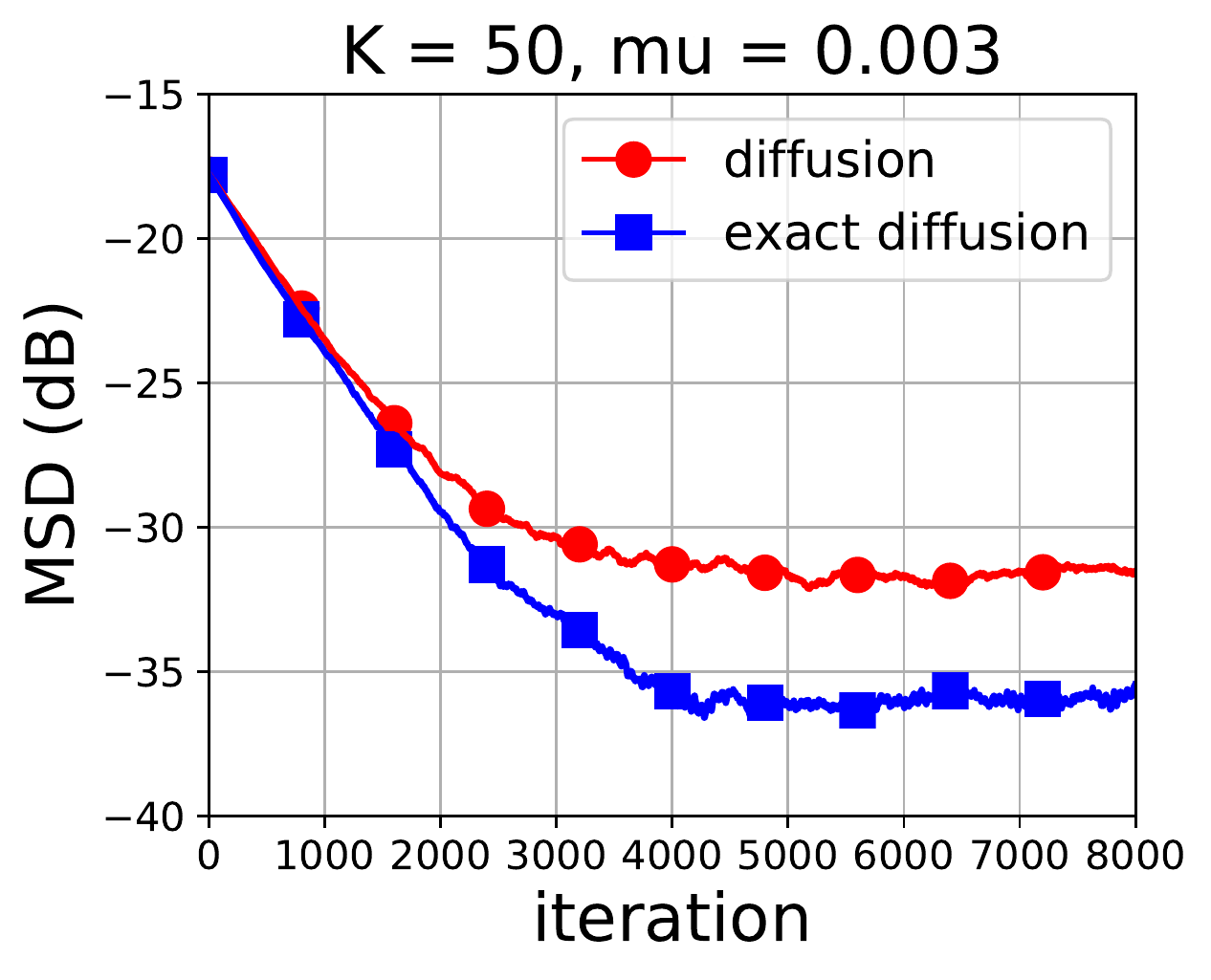}
	\includegraphics[scale=0.33]{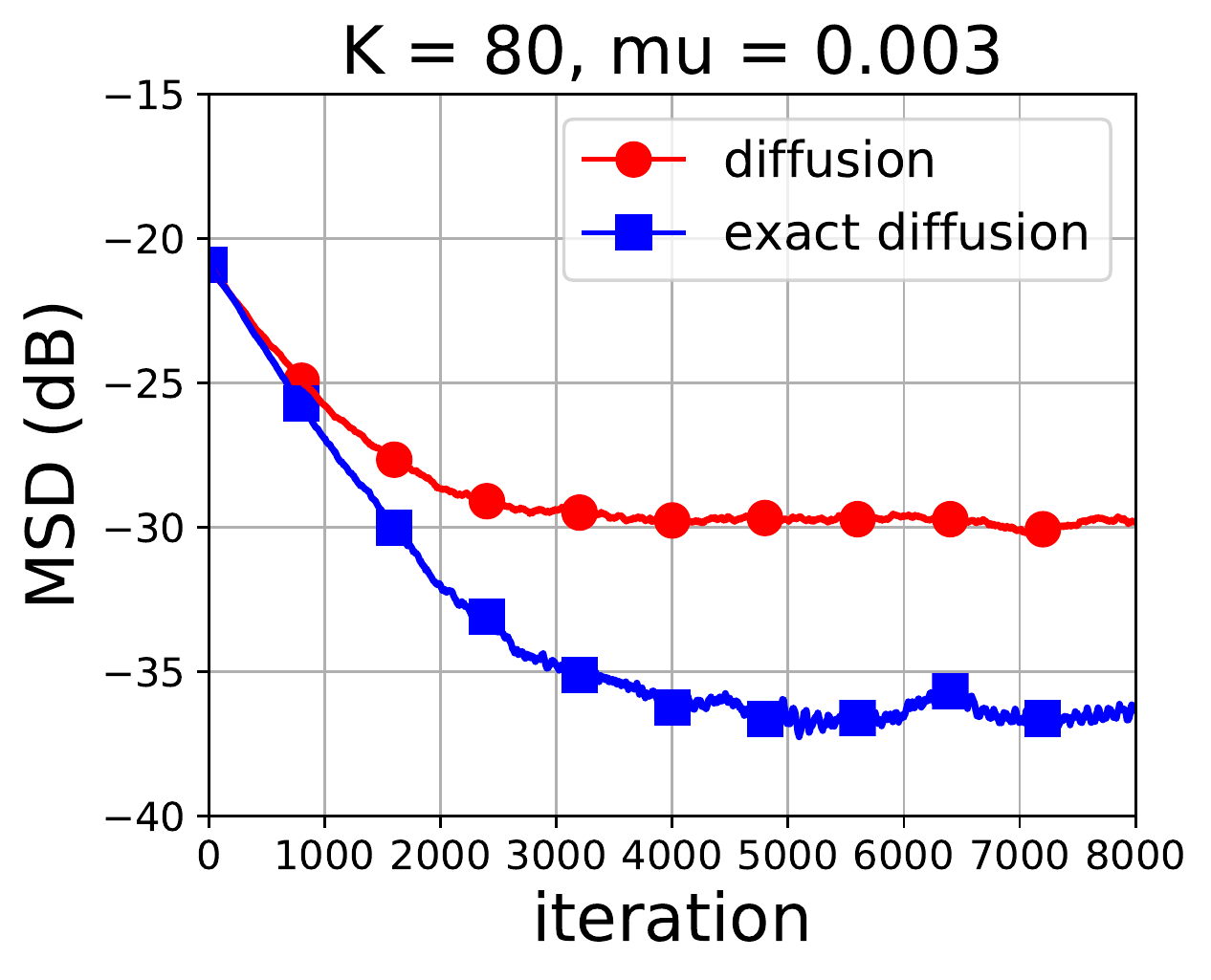}
	\vspace{-3mm}
	\caption{The superiority of exact diffusion gets more evident as the cyclic networks gets larger when solving problem \ref{dist_log_regression}.}
	\label{fig:cyclic-change-size-2}
\end{figure*}

We first compare the performance of exact diffusion and diffusion over a grid topology --- see the first plot in Fig.\ref{fig:topologies}. We first let $K=9$ and $\mu=0.005$ and compare exact diffusion and diffusion. With these two parameters, it is shown in the first plot in Fig.\ref{fig:grid-change-size-ls} that both methods perform almost the same, and the steady-state MSD performance of both methods coincide with the derived MSD expression \eqref{MSD-expression-exact-diffusion}. In the second plot in Fig.\ref{fig:grid-change-size-ls}, we maintain $\mu=0.005$ but increase the network size to $100$ nodes. As we explained in Sec.\ref{sec-comparison-bad-network}, a grid topology with larger network size has $\lambda$ closer to $1$, which amplifies the inherent bias $O(\mu^2b^2/(1-\lambda)^2)$ suffered by diffusion. It is observed that exact diffusion has a clear advantage over diffusion during the steady-state stage. Note that in the second plot both diffusion and exact diffusion do not coincide with the derived theoretical MSD expression. This is because the theoretical MSD expression in \eqref{MSD-expression-exact-diffusion} is only precise to first-order in $\mu$. When $\lambda$ approaches $1$ as the grid network gets larger, the second-order term of $\mu$ is amplified by $1/(1-\lambda)$ and becomes  non-negligible. In the third plot, we maintain $K=100$ and $\mu_{\rm ed} = 0.005$ for exact diffusion while decreasing the step-size of diffusion to  ($\mu_{\rm d}=0.003$) so that it has the same steady-state MSD performance as diffusion. It is observed that in this scenario exact diffusion converges faster than diffusion to reach the same steady-state performance, which implies that exact diffusion has faster adaptive and tracking abilities than diffusion over large grid graphs. In the fourth plot of Fig.\ref{fig:grid-change-size-ls}, we adjust $\mu=0.0001$ {for both methods} while keeping $K=100$. Since $\mu$ gets much smaller, the inherent bias in diffusion \eqref{diffusion-neighborhood} becomes trivial and both methods perform similarly again, and they coincide with the derived MSD expression. 

To further show how superior the exact diffusion can be compared to diffusion over the grid network, we depict the performance of diffusion and exact diffusion for different network sizes in Fig.\ref{fig:grid-change-size-ls-2}. It is observed that the superiority of exact diffusion {becomes} more evident as the grid network gets larger, and exact diffusion performs much better than diffusion when $K=400$.

{In the third experiment, we compare diffusion with exact diffusion over a fully connected network with $K=30$. Since $\lambda = 0$ for this scenario, it is expected diffusion has better steady-state performance than exact diffusion when $\mu$ is moderately small, see the discussion in Sec. \ref{sec-comparison-good-network}. Also, the superiority of diffusion should vanish as the step-size becomes sufficiently small. The comparison results shown in Fig.\ref{fig:fullyconnected-ls} are consistent with our discussion in \ref{sec-comparison-good-network}.

}

\begin{figure}[b!]
	\centering
	\includegraphics[scale=0.33]{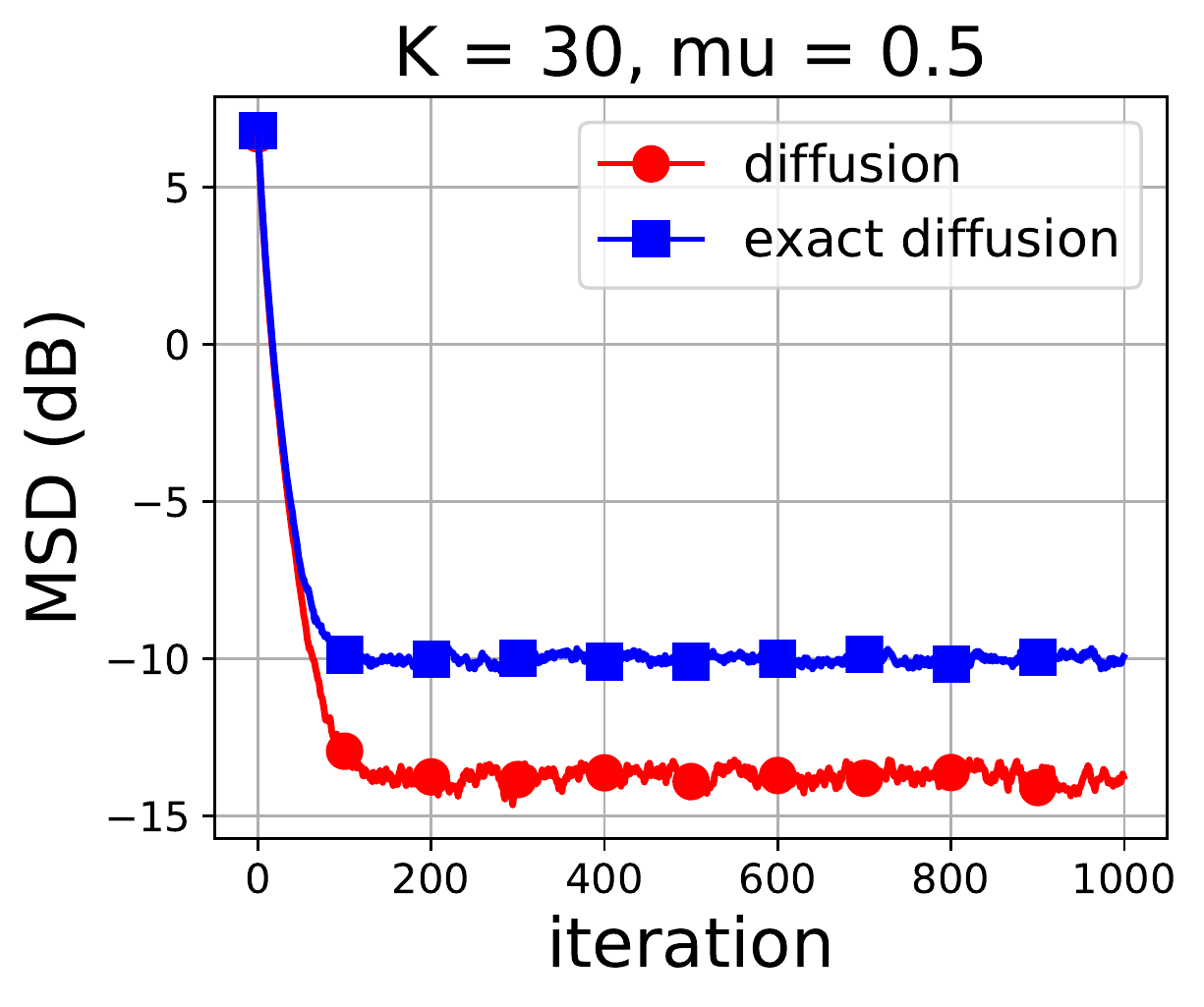}
	\includegraphics[scale=0.33]{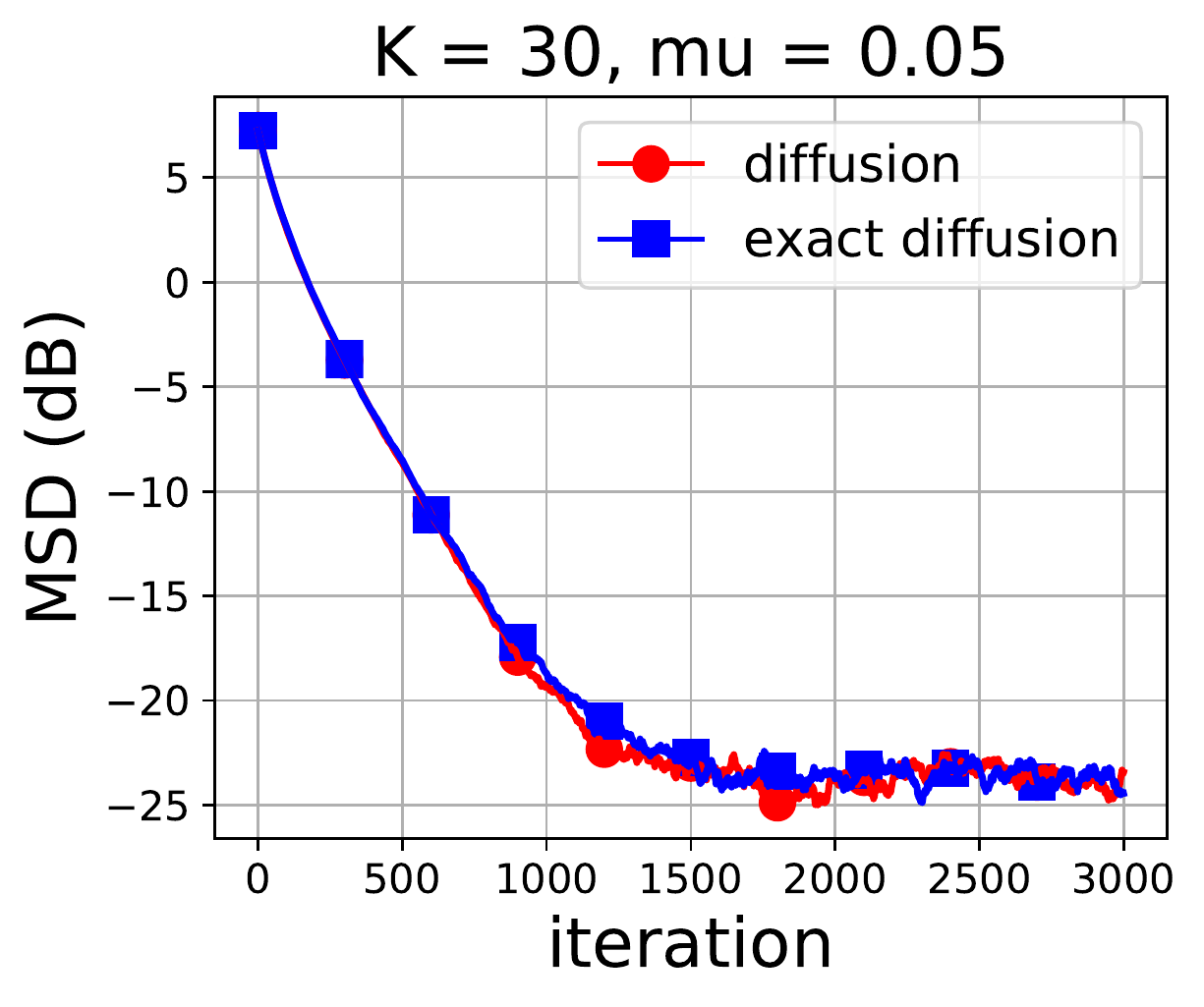}
	\vspace{-3mm}
	\caption{Diffusion v.s. exact diffusion over a fully connected network for problem \eqref{dist_log_regression}.}
	\label{fig:fullyconnected-lr}
\end{figure}


\subsection{Distributed Logistic Regression}
In this subsection we compare the performance of exact diffusion and diffusion when solving a decentralized logistic regression problem of the form:
\eq{\label{dist_log_regression}
	\min_{w \in \RR^M}\ \sum_{k=1}^K \bE\left\{\ln\left(1 + e^{-\bgamma_k \h_k\tran w}\right)\right\} + \frac{\rho}{2}\|w\|^2,
} 
where $(\h_k, \bgamma_k)$ represent the streaming data received by agent $k$. Variable $\h_k \in \RR^M$ is the feature vector and $\bgamma_k \in \{-1, +1\}$ is the label scalar. In all experiments, we set $M=20$ and $\rho = 0.001$. To make the $J_k(w)$'s have different minimizers, we first generate $K$ different local minimizers $\{w^\star_k\}$. All $w_k^\star$ are normalized so that $\|w_k^\star\|^2 = 1$. At agent $k$, we generate each feature vector $\h_{k,i} \sim \cN(0, I_{20})$. To generate the corresponding label $\bgamma_k(i)$, we generate a random variable $z_{k,i}\in \cU(0,1)$. If $z_{k,i}\le 1/(1+\exp(-\h_{k,i}\tran w_k^\star))$, we set $\bgamma_k(i) = 1$; otherwise $\bgamma_k(i) = -1$.

\begin{figure*}[t!]
	\centering
	\includegraphics[scale=0.33]{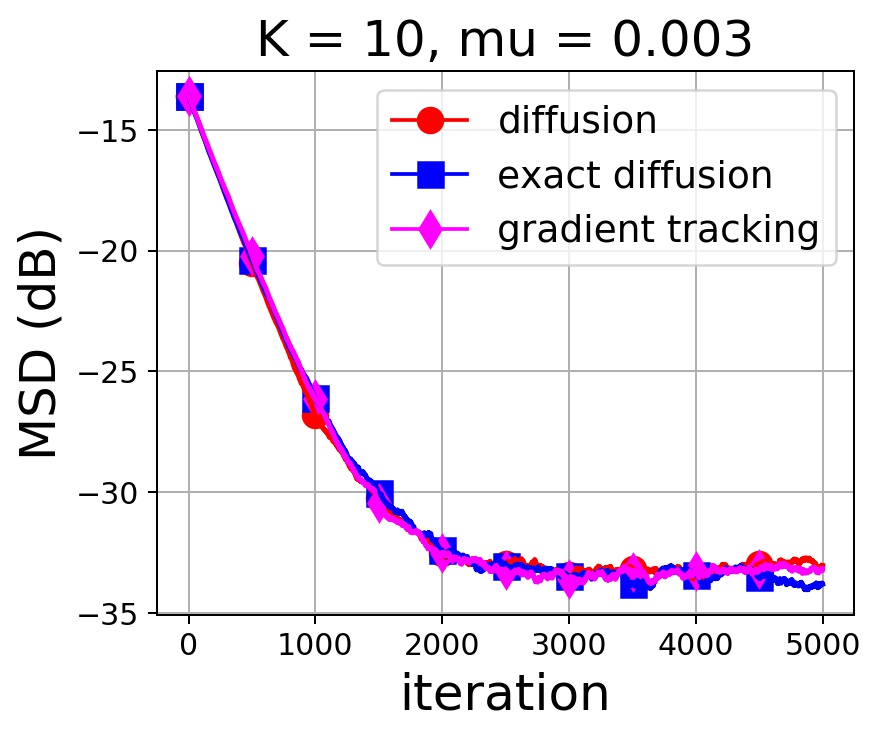}
	\includegraphics[scale=0.33]{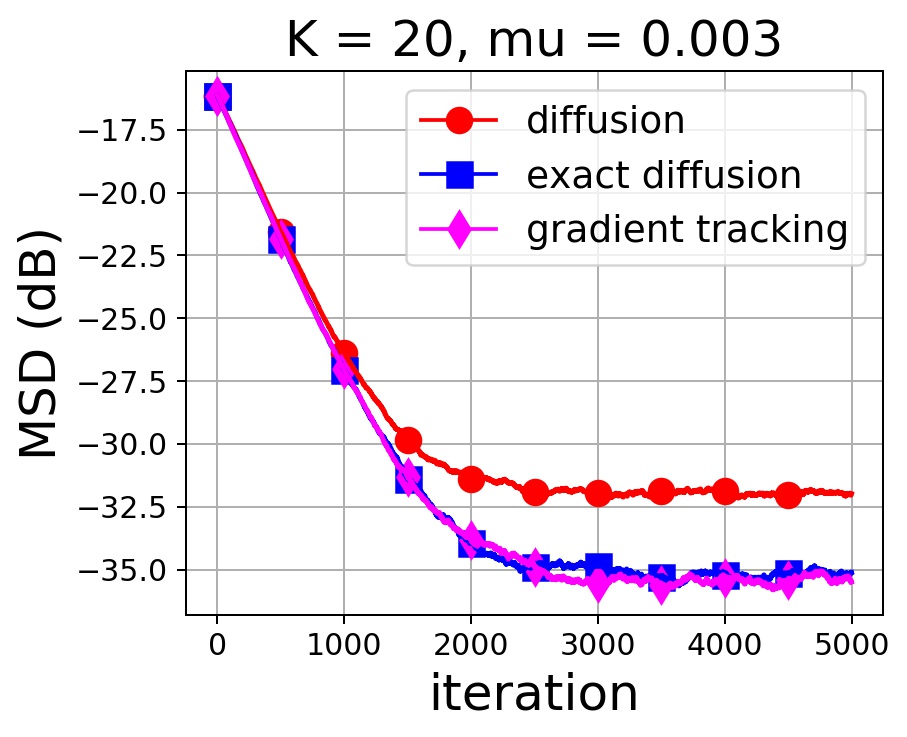}
	\includegraphics[scale=0.33]{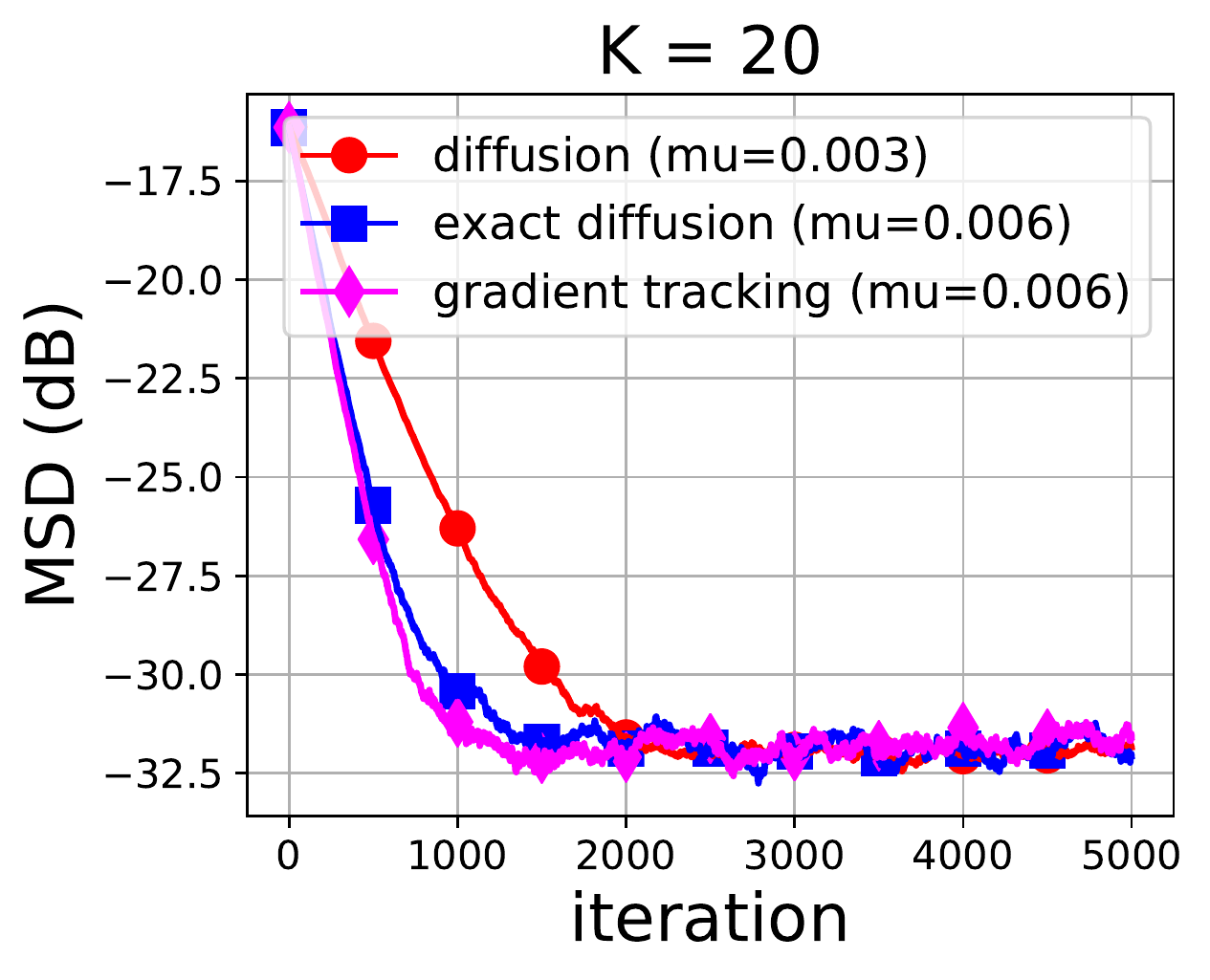}
	\includegraphics[scale=0.33]{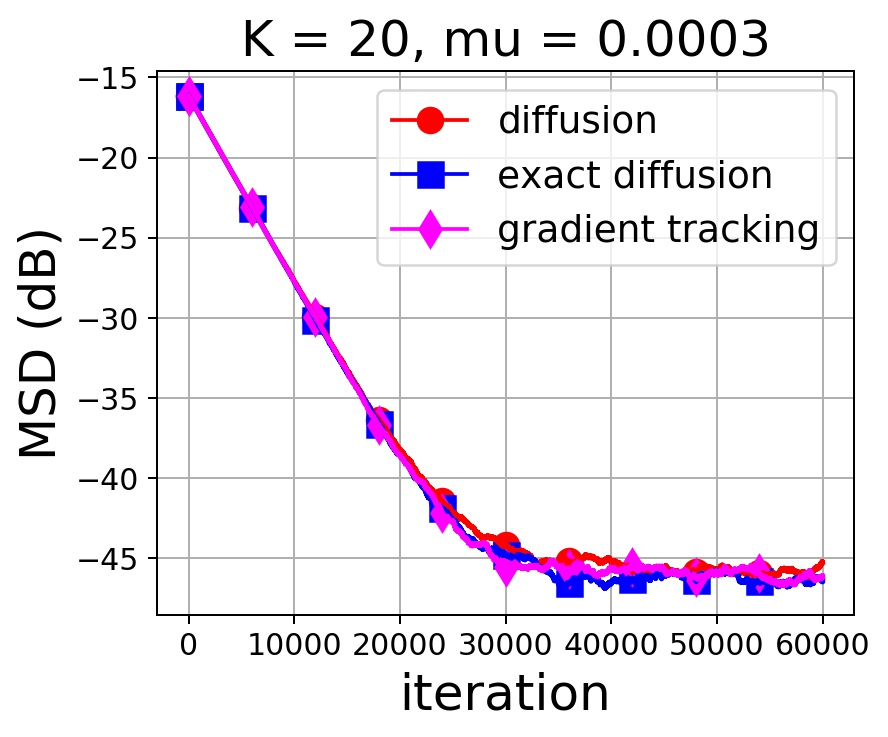}
	\vspace{-3mm}
	\caption{Comparison between diffusion\cite{sayed2014adaptation}, exact diffusion (proposed), and gradient tracking\cite{pu2018distributed} over cyclic networks for problem \eqref{mse-network}.}
	\label{fig:cyclic-change-size-ls-comparisons}
\end{figure*}

We first compare these two methods over a cyclic network, see the simulation in Figs. \ref{fig:cyclic-change-size} and \ref{fig:cyclic-change-size-2}. Similar to Sec. VI.A, the simulation results shown in Figs. \ref{fig:cyclic-change-size} and \ref{fig:cyclic-change-size-2} are also consistent with our discussions in Sec.\ref{sec-comparison-bad-network}. In the third plot in Fig.\ref{fig:cyclic-change-size}, we set $\mu_{\rm d}=0.003$ and $\mu_{\rm ed}=0.006$ so that both diffusion and exact diffusion have the same MSD performance. {Next, we compare diffusion with exact diffusion over a fully connected network in Fig.\ref{fig:fullyconnected-lr}. It is observed that the results are consistent with the discussion in Sec.\ref{sec-comparison-good-network}. }

\subsection{Comparison with Gradient Tracking Methods}
In this subsection we compare exact diffusion with the distributed stochastic gradient tracking method\cite{pu2018distributed,xin2019distributed}. {While \cite{pu2018distributed} shows stochastic gradient tracking has better steady-state MSD performance than {decentralized gradient descent (DGD)} via numerical simulations, it does not study {\em when} and {\em why} gradient tracking can be better DGD. 
In fact, since gradient tracking can also be used to correct the bias suffered by diffusion, we can expect the gradient tracking method to have roughly a similar behavior to exact diffusion. In other words, gradient tracking will have better MSD performance than diffusion when the network is sparsely-connected and worse MSD performance when the network is well-connected. Moreover, the difference between diffusion and gradient tracking will diminish for small step-sizes.} In this subsection, we verify this conclusion using simulations. We first consider the MSE-network \eqref{mse-network} over a cyclic network (which is a sparsely-connected network). The results in Fig.\ref{fig:cyclic-change-size-ls-comparisons} show stochastic gradient tracking behaves as we expected, and it has almost the same performance as exact diffusion in all scenarios. Note though that the gradient tracking method \cite{pu2018distributed} requires twice the amount of communication that is required by exact diffusion, which implies exact diffusion is more communication efficient. In the third plot in Fig.\ref{fig:cyclic-change-size-ls-comparisons}, we set $\mu_{\rm d}=0.003$ and $\mu_{\rm ed} = \mu_{\rm track} = 0.006$ to endow the algorithms with the same steady-state MSD performance. 

{We next compare diffusion, exact diffusion, and gradient tracking method over a fully-connected network (which is a well-connected network). It is observed in Fig.\ref{fig:fullyconnected-ls-comparison} that diffusion has the best MSD performance compared to exact diffusion and gradient tracking, which confirms our conclusion. While reference \cite{pu2018distributed} suggests that gradient tracking is superior to consensus, we observe from the analytical results in the current manuscript and from the simulations in Fig.\ref{fig:fullyconnected-ls-comparison} that there are situations when gradient tracking cannot outperform the traditional diffusion; their performance measures match each other and sometimes gradient tracking can be worse. }
\begin{figure}[h!]
	\centering
	\includegraphics[scale=0.33]{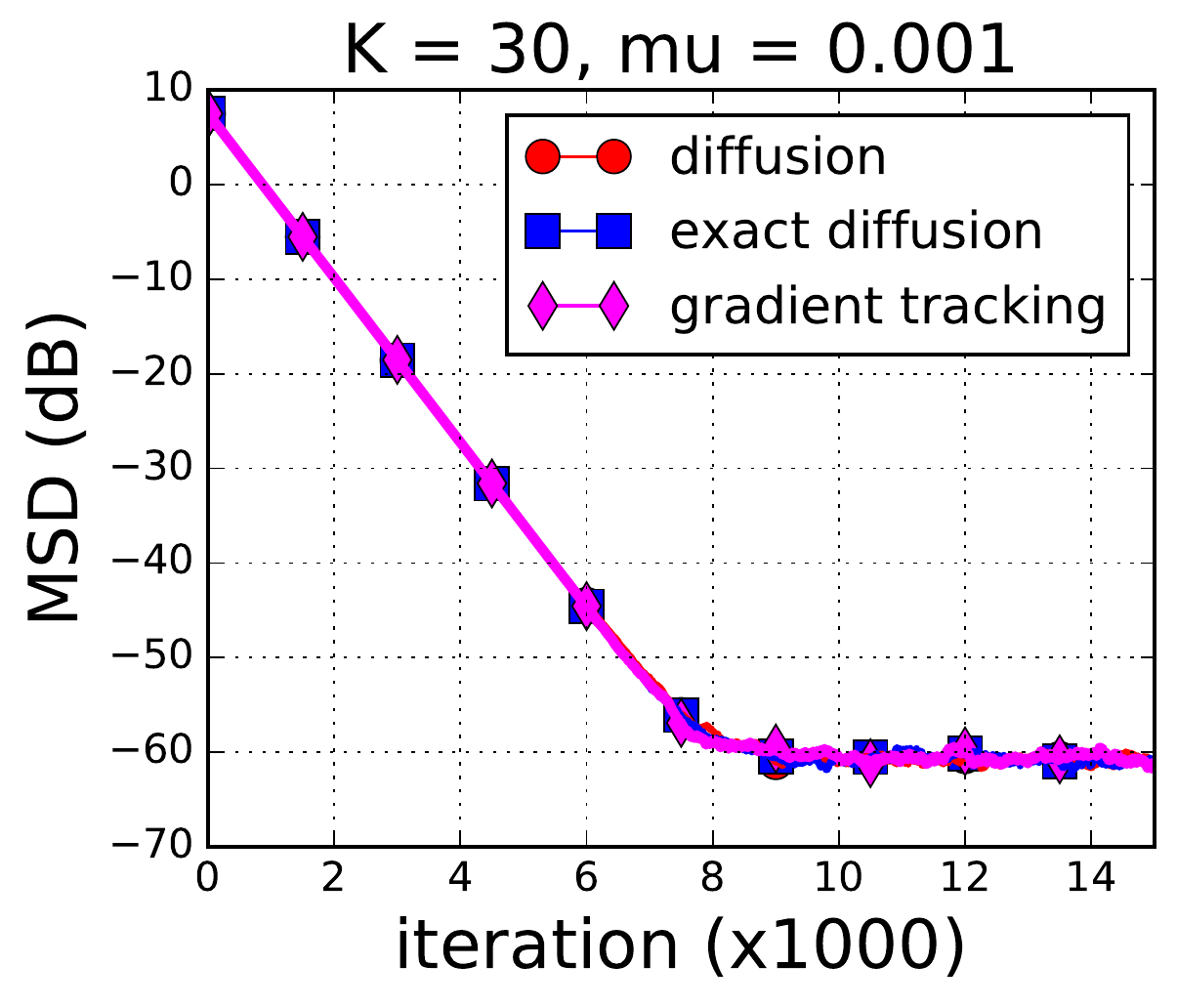}
	\includegraphics[scale=0.33]{ls_alg_comp_fullyConnected_30nodes_mu0p001_07072019-eps-converted-to.pdf}
	\vspace{-3mm}
	\caption{Comparison between diffusion\cite{sayed2014adaptation}, exact diffusion (proposed), and gradient tracking\cite{pu2018distributed, xin2019distributed} over a fully connected network when solving problem \eqref{mse-network}.}
	\label{fig:fullyconnected-ls-comparison}
\end{figure}

\appendices
\section{Proof of Theorem \ref{ed-stability}}
\label{appendix_ed_stability_proof}
From the first line in the transformed error dynamics \eqref{transformed-error-dynamics}, we know that
\eq{\label{nsh8888}
	\bar{\szb}_i & = \Big( I_M \hspace{-1mm}-\hspace{-1mm} \frac{\mu}{K}\sum_{k=1}^K \H_{k,i-1} \Big)	\bar{\szb}_{i-1} - \frac{c\mu}{K}\cI\tran {\boldsymbol \cH}_{i-1} \cX_{R,u} \check{\szb}_{i-1} \nnb
	&\quad \quad + \frac{\mu}{K}\cI\tran \s_i(\swb_{i-1}).
}
By squaring and taking conditional expectation of both sides of the recursion and recalling \eqref{1st-moment}, we get
\eq{\label{23ns8s}
	&\ \bE[\|\bar{\szb}_i\|^2|\filt_{i-1}]	= \nnb
	&\hspace{5mm} \Big\|\hspace{-0.5mm}\Big( I \hspace{-1mm}-\hspace{-1mm} \frac{\mu}{K}\sum_{k=1}^K \hspace{-1mm}\H_{k,i-1} \Big)	\bar{\szb}_{i-1} \hspace{-1mm}-\hspace{-1mm} \frac{c\mu}{K}\cI\tran {\boldsymbol \cH}_{i-1} \cX_{R,u} \check{\szb}_{i-1}\Big\|^2 \nnb
	&\quad + \mu^2 \bE\Big[\Big\| \frac{1}{K}\sum_{k=1}^K\s_{k,i}(\w_{k,i}) \Big\|^2 \Big| \filt_{i-1} \Big].
}
Next note that
\eq{\label{23ns99}
	&\ \Big\|\hspace{-0.5mm}\Big( I \hspace{-1mm}-\hspace{-1mm} \frac{\mu}{K}\sum_{k=1}^K \hspace{-1mm}\H_{k,i-1} \Big)	\bar{\szb}_{i-1} \hspace{-1mm}-\hspace{-1mm} \frac{c\mu}{K}\cI\tran {\boldsymbol \cH}_{i-1} \cX_{R,u} \check{\szb}_{i-1}\Big\|^2 \nnb
	\overset{(a)}{\le} &\ \frac{1}{1-t} \Big\| I \hspace{-1mm}-\hspace{-1mm} \frac{\mu}{K}\sum_{k=1}^K \hspace{-1mm}\H_{k,i-1} \Big\|^2 \|\bar{\szb}_{i-1}\|^2 \nnb
	&\quad + \frac{c^2\mu^2}{K^2 t} \|\cI\|^2 \|{\boldsymbol \cH}_{i-1}\|^2 \|\cX_{R,u}\|^2 \|\check{\szb}_{i-1}\|^2 \nnb
	\overset{(b)}{\le}&\ \frac{(1-\mu \nu)^2}{1-t} \|\bar{\szb}_{i-1}\|^2 + \frac{c^2\mu^2\delta^2 \|\cX_{R,u}\|^2}{Kt} \|\check{\szb}_{i-1}\|^2 \nnb
	\overset{(c)}{=}&\ (1-\mu\nu) \|\bar{\szb}_{i-1}\|^2 + \frac{\mu c^2 \delta^2 \|\cX_{R,u}\|^2}{K \nu} \|\check{\szb}_{i-1}\|^2,
}
where (a) holds for $t\in(0,1)$ because of Jensen's inequality, and (b) holds since $\nu^2 \le \|{\boldsymbol \cH}_{i-1}\|^2 \le \delta^2$, $\|\cI\|^2 = K$, and $\| I - \frac{\mu}{K}\sum_{k=1}^K \hspace{-1mm}\H_{k,i-1} \|^2 \le (1-\mu\nu)^2$
when $\mu \le 1/\delta$. Moreover, equality (c) holds if we choose $t = \mu \nu$. In addition, recall from \eqref{network-ns-2nd} that
\eq{\label{23ns99-2}
	\bE\Big[\Big\|\frac{1}{K}\hspace{-0.5mm}\sum_{k=1}^K\s_{k,i}(\w_{k,i-1})\Big\|^2\Big|\filt_{i-1}\Big] \hspace{-0.5mm} \le \hspace{-0.5mm} \frac{\beta^2}{K} \|\twb_{i-1}\|^2 \hspace{-1mm} + \hspace{-1mm} \frac{\sigma^2}{K}
}
Moreover, we can bound $\|\twb_{i-1}\|^2$ as 
\eq{\label{23nsnn}
	\|\twb_{i-1}\|^2 &\overset{\eqref{w-y-vs-bz-cz}}{=} \| \cI \bar{\szb}_{i-1} + c \cX_{R,u} \check{\szb}_{i-1} \|^2 \nnb
	&\le 2 \|\cI \bar{\szb}_{i-1}\|^2 + 2c^2\| \cX_{R,u} \check{\szb}_{i-1} \|^2 \nnb
	&\le 2K \|\bar{\szb}_{i-1}\|^2 + 2 c^2 \| \cX_{R,u}\|^2 \|\check{\szb}_{i-1} \|^2.
}
Substituting \eqref{23ns99}, \eqref{23ns99-2} and \eqref{23nsnn} into \eqref{23ns8s}, we reach
\eq{\label{1st-inquality}
	&\bE[\|\bar{\szb}_i\|^2|\filt_{i-1}] \nnb
	&\le  (1-\mu\nu + 2\mu^2\beta^2) \|\bar{\szb}_{i-1}\|^2 \nnb
	& \quad + \Big(\frac{\mu c^2 \delta^2}{K \nu} + \frac{2\mu^2 c^2 \beta^2}{K}\Big)\|\cX_{R,u}\|^2 \|\check{\szb}_{i-1}\|^2  + \frac{\mu^2 \sigma^2}{K} \nnb
	&\le  (1-\mu\nu + 2\mu^2\beta^2) \|\bar{\szb}_{i-1}\|^2 \nnb
	& \quad + \Big(\frac{\mu c^2 \delta^2}{K \nu} + \frac{2\mu^2 c^2 \beta^2}{K}\Big)\|\cX_{R}\|^2 \|\check{\szb}_{i-1}\|^2  + \frac{ \mu^2 \sigma^2}{K},
}
where the last inequality holds since
\eq{\label{xb238dsysd}
	\|\cX_{R,u}\|^2 &= 	
	\|\ba{cc}
	I_{KM} & 0
	\ea
	\cX_R\|^2 \nnb
	&\le \|
	\ba{cc}
	I_{KM} & 0
	\ea\|^2 \|\cX_R\|^2 = \|\cX_R\|^2
}
By taking expectation over the filtration, we get
\eq{\label{1st-inquality-exp}
	\bE\|\bar{\szb}_i\|^2 &\le  (1-\mu\nu + 2\mu^2\beta^2) \bE\|\bar{\szb}_{i-1}\|^2 \nnb
	&\hspace{-8mm} + \Big(\frac{\mu c^2 \delta^2}{K \nu} + \frac{2\mu^2 c^2 \beta^2}{K}\Big)\|\cX_{R}\|^2 \bE\|\check{\szb}_{i-1}\|^2  +  \frac{ \mu^2 \sigma^2}{K}.
}
On the other hand, from the second line in \eqref{transformed-error-dynamics} we have
\eq{\label{b73809}
	\check{\szb}_i =&\ \cD_1 \check{\szb}_{i-1} - \frac{\mu}{c} \cD_1\cX_L{\boldsymbol \cT_{i-1}} (\cR_1 \bar{\szb}_{i-1} + c \cX_R \check{\szb}_{i-1}) \nnb
	&\quad + \frac{\mu}{c}\cD_1 \cX_L \cB_\ell \s_i(\swb_{i-1}).
}
By squaring and taking conditional expectation of both sides of the above recursion and recalling \eqref{1st-moment}, we get
\eq{\label{xbnw87}
	&\ \bE[\|\check{\szb}_i\|^2|\filt_{i-1}] \nnb
	=&\ \|\cD_1 	\check{\szb}_{i-1} - \frac{\mu}{c} \cD_1 \cX_L{\boldsymbol \cT_{i-1}} (\cR_1 \bar{\szb}_{i-1} + c \cX_R \check{\szb}_{i-1})\|^2 \nnb
	&\quad + \frac{\mu^2\|\cD_1\|^2}{c^2} \bE[\|\cX_L \cB_\ell \s_i(\swb_{i-1})\|^2|\filt_{i-1}].
}
Note that
\eq{\label{na97dh}
	&\	\|\cD_1 	\check{\szb}_{i-1} - (\mu/c) \cX_L{\boldsymbol \cT_{i-1}} (\cR_1 \bar{\szb}_{i-1} + c\cX_R \check{\szb}_{i-1})\|^2 \nnb
	\le &\ \frac{1}{t}\|\cD_1 	\check{\szb}_{i-1}\|^2 + \frac{\mu^2\|\cD_1\|^2}{c^2(1-t)}\left\|\cX_L{\boldsymbol \cT_{i-1}} (\cR_1 \bar{\szb}_{i-1} + c \cX_R \check{\szb}_{i-1})\right\|^2 \nnb
	\le &\ \frac{1}{t}\|\cD_1\|^2 	\|\check{\szb}_{i-1}\|^2 + \frac{2\mu^2\|\cD_1\|^2 }{c^2(1-t)} \|\cX_L\|^2\|{\boldsymbol \cT_{i-1}}\|^2\|\cR_1\|^2 \|\bar{\szb}_{i-1}\|^2 \nnb
	&\quad + \frac{2\mu^2\|\cD_1\|^2 }{1-t}\|\cX_L\|^2\|{\boldsymbol \cT_{i-1}}\|^2\|\cX_R\|^2 \| \check{\szb}_{i-1}\|^2,
}
where $t\in (0,1)$. To simplify the above inequality, we denote 
\eq{
	\lambda_2 &\define \lambda_2({A}),\quad 	{\lambda}^\prime \define \lambda_2(\bar{A}), \label{lambda2} \\
	\lambda &\define \max\{|\lambda_2(A)|, |\lambda_K(A)|\}. \label{lambda}
}
Since $\bar{A} = (A+I_K)/2$ and $A$ is doubly-stochastic, we have
\eq{\label{lambda_prime}
	{\lambda}' = {(1 + \lambda_2)}/{2} \in (0, 1).	
}
From Lemma 4 in \cite{yuan2017exact2} we know that 
\eq{
	\|\cD_1\| = \sqrt{\lambda'} \in (0, 1).	
}
Also, from the definition of ${\boldsymbol \cT_i}$ in \eqref{original-error-dynamics}, we have
\eq{\label{xnwe88}
	\|\boldsymbol \cT_i\|^2 =
	\left\| 
	\ba{cc}
	{\boldsymbol \cH_{i}} & 0 \\
	0 & 0
	\ea
	\right\|^2 \le \delta^2.
}
By substituting \eqref{xnwe88} into \eqref{na97dh}, setting $t=\sqrt{\lambda'}$ and recalling {$\|\cR_1\|^2 = \|\cI\|^2 = K$}, we get
\eq{\label{na97dh-2}
	&\	\|\cD_1 	\check{\szb}_{i-1} - \frac{\mu}{c} \cX_L{\boldsymbol \cT_{i-1}} (\cR_1 \bar{\szb}_{i-1} + c\cX_R \check{\szb}_{i-1})\|^2 \nnb
	\le &\ \hspace{-1mm} \Big( \hspace{-1mm} \sqrt{\lambda'} \hspace{-1mm}+\hspace{-1mm} \frac{2\mu^2\delta^2\lambda^\prime\|\cX_L\|^2\|\cX_R\|^2}{1-\sqrt{\lambda'}} \hspace{-0.6mm} \Big)	\|\check{\szb}_{i-1}\|^2 \nnb
	&\quad \hspace{-1mm}+\hspace{-1mm} \frac{2K\mu^2\delta^2\|\cX_L\|^2\lambda^\prime}{c^2(1-\sqrt{\lambda'})} \|\hspace{-0.5mm}\bar{\szb}_{i-1}\hspace{-0.5mm}\|^2
}
In addition, it also holds that
\eq{\label{zn28s7dfg}
	&\ \bE[\|\cX_L \cB_\ell \s_i(\swb_{i-1})\|^2|\filt_{i-1}]	\nnb
	\le &\ \|\cX_L\|^2\|\cB_\ell\|^2\bE[\|\s_i(\swb_{i-1})\|^2|\filt_{i-1}] \nnb
	\overset{(a)}{\le} &\ K\|\cX_L\|^2 \beta^2 \|\twb_{i-1}\|^2 + K\|\cX_L\|^2  \sigma^2 \nnb
	\overset{\eqref{23nsnn}}{\le} &\ 2K^2 \|\cX_L\|^2 \beta^2 \|\bar{\szb}_{i-1}\|^2 + 2Kc^2\|\cX_{R,u}\|^2 \|\cX_L\|^2 \beta^2 \|\check{\szb}_{i-1}\|^2 \nnb
	&\quad + K\|\cX_L\|^2 \sigma^2 \nnb
	{\overset{(b)}{\le}} &\ {\frac{2K^2 \|\cX_L\|^2 \beta^2 \|\bar{\szb}_{i-1}\|^2}{1 - \sqrt{\lambda'}} + \frac{2c^2K\|\cX_{R}\|^2 \|\cX_L\|^2 \beta^2 \|\check{\szb}_{i-1}\|^2}{1 - \sqrt{\lambda'}}} \nnb
	&\quad { + K\|\cX_L\|^2 \sigma^2}
}
where (a) holds because of inequality \eqref{network-ns-2nd} and the fact
\eq{
	\|\cB_\ell\|^2 = \left\| \cB \ba{c} I_{KM} \nnb
	0
	\ea \right\|^2 \le \|\cB\|^2 = 1 
}
in which the last equality holds because of Lemma \ref{lemma-fundamental-decomposition}. {The inequality (b) holds since $1-\sqrt{\lambda'} \in (0,1)$ and inequality \eqref{xb238dsysd}.	By substituting \eqref{na97dh-2} and \eqref{zn28s7dfg} into \eqref{xbnw87}, we have
	\eq{\label{2nd-inquality}
		&\ \bE[\|\check{\szb}_i\|^2|\filt_{i-1}] \nnb
		\le&\ \Big( \sqrt{\lambda'} + \frac{2\lambda^\prime \mu^2(\delta^2 + K \beta^2)\|\cX_L\|^2\|\cX_R\|^2}{1-\sqrt{\lambda'}} \Big) \|\check{\szb}_{i-1}\|^2 \nnb
		&\quad  + \frac{2\lambda^\prime K\mu^2(\delta^2+ K \beta^2)\|\cX_L\|^2}{(1-\sqrt{\lambda'})c^2}\|\bar{\z}_{i-1}\|^2 \nnb
		&\quad + \frac{\mu^2 \lambda^\prime K \|\cX_L\|^2 \sigma^2}{c^2}
	}
	By taking expectation over the filtration, we get
	\eq{\label{2nd-inquality-exp}
		&\ \bE\|\check{\szb}_i\|^2 \nnb
		\le&\ \Big( \sqrt{\lambda'} + \frac{2\lambda^\prime \mu^2(\delta^2 + K \beta^2)\|\cX_L\|^2\|\cX_R\|^2}{1-\sqrt{\lambda'}} \Big) \bE \|\check{\szb}_{i-1}\|^2 \nnb
		&\ + \frac{2K\lambda^\prime \mu^2(\delta^2+ K \beta^2)\|\cX_L\|^2}{(1-\sqrt{\lambda'})c^2} \bE\|\bar{\szb}_{i-1}\|^2 \hspace{-1mm} \nnb
		&\ + \hspace{-1mm} \frac{\lambda^\prime\mu^2K}{c^2} \|\cX_L\|^2 \sigma^2
	}
	
	To simplify notation, we introduce the constants
	\eq{\label{constants-c1-c2}
		c_1 = \|\cX_L\|^2,\quad c_2 = \|\cX_R\|^2.
	}
	Combining \eqref{1st-inquality-exp} and \eqref{2nd-inquality-exp}, we have
	\eq{
		\ba{c}
		\hspace{-1mm}\bE\|\bar{\szb}_i\|^2\hspace{-1mm}\\
		\hspace{-1mm}\bE\|\check{\szb}_i\|^2\hspace{-1mm}
		\ea \hspace{-0.8mm}\le\hspace{-0.8mm}&
		\ba{cc}
		\hspace{-1mm}1\hspace{-0.8mm}-\hspace{-0.8mm}\mu\nu\hspace{-0.8mm}+\hspace{-0.8mm}2\mu^2\beta^2\hspace{-1mm} &\hspace{-1mm} \Big(\frac{\mu c^2 \delta^2}{K \nu} \hspace{-0.8mm}+\hspace{-0.8mm} \frac{2\mu^2 c^2 \beta^2}{K}\Big)c_2 \hspace{-0.8mm}\\
		\hspace{-1mm}\frac{2K\lambda^\prime \mu^2(\delta^2+K\beta^2)c_1}{(1-\sqrt{\lambda'})c^2} \hspace{-1mm}&\hspace{-1mm} \sqrt{\lambda'} \hspace{-0.8mm}+\hspace{-0.8mm} \frac{2\mu^2\lambda^\prime (\delta^2 + K\beta^2)c_1 c_2}{1-\sqrt{\lambda'}}\hspace{-0.8mm}
		\ea \nnb
		&\quad \times 
		\ba{c}
		\hspace{-1mm}\bE\|\bar{\szb}_{i-1}\|^2\hspace{-1mm}\\
		\hspace{-1mm}\bE\|\check{\szb}_{i-1}\|^2\hspace{-1mm}
		\ea  +
		\ba{c}
		\frac{1}{K}\mu^2 \sigma^2 \\
		\frac{K\lambda^\prime c_1}{c^2} \mu^2 \sigma^2
		\ea.
	}
	{Note that $c$ is a parameter that can be set to any positive value.} If we let $c^2 = Kc_1$, then the above inequality becomes
	\eq{\label{xbw7dhd9}
		\ba{c}
		\hspace{-1mm}\bE\|\bar{\szb}_i\|^2\hspace{-1mm}\\
		\hspace{-1mm}\bE\|\check{\szb}_i\|^2\hspace{-1mm}
		\ea \le&\  
		\ba{cc}
		1-\mu\nu+2\mu^2\beta^2 & \Big(\frac{\mu \delta^2}{\nu} + 2\mu^2 \beta^2\Big)c_1 c_2 \\
		\frac{2\lambda^\prime\mu^2(\delta^2+K\beta^2)}{1-\sqrt{\lambda'}} & \sqrt{\lambda'} + \frac{2\lambda^\prime \mu^2(\delta^2 + K\beta^2)c_1 c_2}{1-\sqrt{\lambda'}}
		\ea \nnb
		&\quad \times 
		\ba{c}
		\hspace{-1mm}\bE\|\bar{\szb}_{i-1}\|^2\hspace{-1mm}\\
		\hspace{-1mm}\bE\|\check{\szb}_{i-1}\|^2\hspace{-1mm}
		\ea  +
		\ba{c}
		\frac{1}{K}\mu^2 \sigma^2 \\
		\lambda^\prime \mu^2 \sigma^2
		\ea.
	}
	If we choose  $\mu$ sufficiently small such that
	\eq{
		1-\mu\nu+2\mu^2\beta^2 &\le 1 - \frac{1}{2}\mu\nu, \label{cond-1}\\
		\Big(\frac{\mu \delta^2}{\nu} + 2\mu^2 \beta^2\Big)c_1 c_2 &\le \frac{2\mu\delta^2c_1c_2}{\nu}, \\
		\frac{2\lambda^\prime \mu^2(\delta^2+K\beta^2)}{1-\sqrt{\lambda'}} &\le \frac{1}{4}\lambda^\prime\mu\nu,\\
		\sqrt{\lambda'} + \frac{2\lambda^\prime \mu^2(\delta^2 + K\beta^2)c_1 c_2}{1-\sqrt{\lambda'}} &\le \frac{1+\sqrt{\lambda'}}{2}, \label{cond-4}
	}
	then inequality \eqref{xbw7dhd9} becomes 
	\eq{\label{xbw7dhd9-2}
		\ba{c}
		\hspace{-1mm}\bE\|\bar{\szb}_i\|^2\hspace{-1mm}\\
		\hspace{-1mm}\bE\|\check{\szb}_i\|^2\hspace{-1mm}
		\ea \le&\  
		\underbrace{\ba{cc}
			1-\frac{1}{2}\mu\nu & \frac{2\mu\delta^2c_1c_2}{\nu} \\
			\frac{1}{4}\lambda^\prime\mu\nu & \frac{1+\sqrt{\lambda'}}{2}
			\ea}_{\define \cC}
		\ba{c}
		\hspace{-1mm}\bE\|\bar{\szb}_{i-1}\|^2\hspace{-1mm}\\
		\hspace{-1mm}\bE\|\check{\szb}_{i-1}\|^2\hspace{-1mm}
		\ea \nnb
		&\quad +
		\ba{c}
		\frac{1}{K}\mu^2 \sigma^2 \\
		\lambda^\prime \mu^2 \sigma^2
		\ea.
	}
	To satisfy \eqref{cond-1}--\eqref{cond-4}, it is enough to let  $\mu$ satisfy
	\eq{\label{mu_range}
		\mu \le \frac{(1-\sqrt{\lambda'})\nu}{(8\hspace{-0.5mm}+\hspace{-0.5mm}4c_1c_2\hspace{-0.5mm}+\hspace{-0.5mm}\sqrt{4c_1c_2})(\delta^2\hspace{-0.5mm}+\hspace{-0.5mm}K\beta^2)},	
	}
	{Also, note that $1-\sqrt{\lambda'} = (1-\lambda')/(1+\sqrt{\lambda'})$. Since 
		$0 < \lambda' < 1$, we have ${(1-\lambda')}/{2} < 1-\sqrt{\lambda'} < 1-\lambda'$. Moreover, since $\lambda' = (1+\lambda_2)/2$ (see \eqref{lambda_prime}), we have
		\eq{\label{sqrt_lmd_2-0}
			\frac{1-\lambda_2}{4} < 1-\sqrt{\lambda'} < \frac{1-\lambda_2}{2}.
		}
		From \eqref{lambda} we have $|\lambda_2| \le \lambda$, which further implies $-\lambda \le \lambda_2 \le \lambda$. This together with \eqref{sqrt_lmd_2-0} leads to 
		\eq{\label{sqrt_lmd_2}
			\frac{1-\lambda}{4} < 1-\sqrt{\lambda'} < \frac{1+\lambda}{2}.
		}
		With relation \eqref{sqrt_lmd_2}, we know that if  $\mu$ satisfies 
		\eq{\label{mu_range_2}
			\mu \le \frac{(1-\lambda)\nu}{(32\hspace{-0.5mm}+\hspace{-0.5mm}16c_1c_2\hspace{-0.5mm}+\hspace{-0.5mm}8\sqrt{c_1c_2})(\delta^2\hspace{-0.5mm}+\hspace{-0.5mm}K\beta^2)},
		}
		then $\mu$ must also satisfy \eqref{mu_range}.} Recall that $\beta^2 = \frac{\max_k\{\beta^2_k\}}{K}$, we have $K\beta^2 = \beta^2_{\rm max}=\max_k\{\beta^2_k\}$.
	
	Next we examine the spectral radius of the matrix $\cC$. Note that $\lambda^\prime \in (0,1)$, it is easy to verify that
	\eq{
		\rho(\cC)\le \|\cC\|_1 	=&\ \max\left\{ 1\hspace{-0.8mm}-\hspace{-0.8mm}\frac{\mu\nu}{2} \hspace{-0.8mm}+\hspace{-0.8mm} \frac{\lambda^\prime \mu\nu}{4},\  \frac{1+\sqrt{\lambda'}}{2} \hspace{-0.8mm}+\hspace{-0.8mm} \frac{2\mu\delta^2c_1 c_2}{\nu} \right\} \nnb
		\le&\ \max\left\{ 1\hspace{-0.8mm}-\hspace{-0.8mm}\frac{\mu\nu}{4},\  \frac{1+\sqrt{\lambda'}}{2} \hspace{-0.8mm}+\hspace{-0.8mm} \frac{2\mu\delta^2c_1 c_2}{\nu} \right\} \nnb
		\overset{\eqref{mu_range}}{\le}&\ 1 - \frac{1}{4}\mu\nu < 1, \label{spectral_radius_C_ed}
	}
	and therefore $\cC$ is a stable matrix, and $\rho(C) = 1 - O(\mu\nu)$ is the convergence rate of $\bE\|\twb_i\|^2$. Next we examine:
	\eq{
		&\hspace{-8mm} (I-\cC)^{-1} \nnb
		=&\ 
		\ba{cc}
		\frac{\mu\nu}{2} & - \frac{2\mu\delta^2c_1c_2}{\nu} \nnb
		-\frac{\lambda^\prime \mu\nu}{4} & \frac{1-\sqrt{\lambda'}}{2}
		\ea^{-1} \nnb
		=&\ {\frac{4}{(1-\sqrt{\lambda'})\mu\nu - 2\lambda^\prime \mu^2\delta^2c_1 c_2}} \ba{cc}
		\frac{1-\sqrt{\lambda'}}{2} & \frac{2\mu\delta^2c_1c_2}{\nu} \\
		\frac{\mu\nu \lambda^\prime}{4} & \frac{\mu\nu}{2}
		\ea\nnb
		\overset{(a)}{\le}&\ \frac{8}{\mu\nu(1-\sqrt{\lambda'})}
		\ba{cc}
		\frac{1-\sqrt{\lambda'}}{2} & \frac{2\mu\delta^2c_1c_2}{\nu} \\
		\frac{\mu\nu\lambda^\prime}{4} & \frac{\mu\nu}{2}
		\ea \nnb
		=&\ 
		\ba{cc}
		\frac{4}{\mu\nu} & \frac{16\delta^2c_1c_2}{\nu^2(1-\sqrt{\lambda'})} \\
		\frac{2\lambda^\prime}{1-\sqrt{\lambda'}} & \frac{4}{1-\sqrt{\lambda'}}
		\ea, \label{I-C-inverse}
	} 
	where inequality (a) holds since 
	\eq{
		(1-\sqrt{\lambda'})\mu\nu - 2\lambda^\prime \mu^2\delta^2c_1 c_2 \ge \frac{(1-\sqrt{\lambda'})\mu\nu}{2}
	}
	when $\mu$ satisfies \eqref{mu_range}. By iterating \eqref{xbw7dhd9-2}, we conclude that 
	\eq{
		\limsup_{i\to\infty} 
		\ba{c}
		\hspace{-1mm}\bE\|\bar{\szb}_i\|^2\hspace{-1mm}\\
		\hspace{-1mm}\bE\|\check{\szb}_i\|^2\hspace{-1mm}
		\ea 
		\le&\	(I-\cC)^{-1}
		\ba{c}
		\frac{1}{K}\mu^2 \sigma^2 \\
		\lambda^\prime \mu^2 \sigma^2
		\ea \nnb
		\overset{\eqref{I-C-inverse}}{=}&\ \ba{c}
		\frac{4\mu\sigma^2}{K\nu} + \frac{16\lambda^\prime \delta^2c_1c_2\mu^2\sigma^2}{\nu^2(1-\sqrt{\lambda'})} \\
		\frac{2\lambda^\prime \mu^2\sigma^2}{K(1-\sqrt{\lambda'})} +  \frac{4\lambda^\prime \mu^2\sigma^2}{1-\sqrt{\lambda'}}
		\ea.\label{z-steady-state}
	}
	As a result, we obtain
	\eq{\label{23nsd27}
		& \limsup_{i\to\infty}\bE\|\twb_i\|^2 \nnb
		\overset{\eqref{23nsnn}}{\le}&\ \limsup_{i\to\infty} \big(2K\bE\|\bar{\szb}_i\|^2 + 2K c_1 c_2 \bE\|\check{\szb}_i\|^2\big) \nnb
		\overset{\eqref{z-steady-state}}{\le}&\ \frac{8\mu\sigma^2}{\nu} + \frac{(32K\delta^2 + 4\nu^2 + 8K\nu^2)\lambda^\prime c_1c_2\mu^2\sigma^2}{\nu^2(1-\sqrt{\lambda'})} \nnb
		\le&\ \frac{8\mu\sigma^2}{\nu} + \frac{44K\delta^2\lambda^\prime c_1c_2\mu^2\sigma^2}{\nu^2(1-\sqrt{\lambda'})} \nnb
		\overset{\eqref{sqrt_lmd_2}}{\le}&\ \frac{8\mu\sigma^2}{\nu} \hspace{-1mm}+\hspace{-1mm} \frac{176K\lambda^\prime c_1c_2\delta^2\mu^2\sigma^2}{\nu^2(1-{\lambda})} \nnb
		\overset{(a)}{\le}&\ \frac{8\mu\sigma^2}{\nu} \hspace{-1mm}+\hspace{-1mm} \frac{88K(1+\lambda) c_1c_2\delta^2\mu^2\sigma^2}{\nu^2(1-{\lambda})} \nnb
		\overset{(b)}{=}&\ \hspace{-0.5mm} O\left(\frac{\mu\sigma^2}{\nu} \hspace{-1mm}+ \hspace{-1mm} \frac{K\delta^2}{\nu^2}\cdot\frac{\mu^2\sigma^2}{1-\lambda}\right)
	}
	where (a) holds because $\lambda^\prime = (1+\lambda_2(A))/2 \le (1+\lambda)/2$ and (b) holds because $\lambda<1$. Result \eqref{23nsd27} leads to \eqref{ed-neighborhood} by dividing $K$ to both sides of \eqref{23nsd27}.
	
	\section{Proof of Lemma \ref{diffusion-stability}} \label{appendix_diff_stability}
	This section establishes the mean-square convergence of diffusion. With definition \eqref{grad Q definition}, we can rewrite diffusion recursions \eqref{diffusion-1}--\eqref{diffusion-2} as 
	\eq{
		\swb_i = \cA \big(\swb_{i-1} - \mu \grad  \cQ(\swb_{i-1};\sxb_i)\big).
	}
	With relation \eqref{grad J = grad Q + s}, the above recursion becomes
	\eq{
		\swb_i = \cA \big(\swb_{i-1} - \mu \grad  \cJ(\swb_{i-1}) - \mu \s_i(\swb_{i-1})\big),
	}
	which also leads to 
	\eq{\label{diffusion-3}
		\twb_i &= \cA \big(\twb_{i-1} + \mu \grad \cJ(\swb_{i-1}) + \mu \s_i(\swb_{i-1})\big)	\nnb
		&= \cA \big(\twb_{i-1} + \mu \grad \cJ(\swb_{i-1}) - \mu \grad \cJ(\sw^\star)\big)\nnb
		&\quad + \mu\cA \grad \cJ(\sw^\star) + \mu \cA\s_i(\swb_{i-1}) \nnb
		&\overset{\eqref{mean-value}}{=} \cA \bigg((I - \mu  {\boldsymbol \cH}_{i-1})\twb_{i-1} + \mu h + \mu \s_i(\swb_{i-1})\bigg),
	}
	where $\twb_i = \sw^\star - \swb_i$ and $h \hspace{-1mm} \define \hspace{-1mm} \grad \cJ(\sw^\star)$. Note that $\cA = A\otimes I_M$ is symmetric and doubly stochastic, it holds that 
	\eq{
		\cA = 
		\underbrace{
			\ba{ccc}
			\hspace{-1mm}\cI & c\cX_R\hspace{-1mm}
			\ea}_{\cX}
		\ba{cc}
			\hspace{-1mm}I_M & 0\hspace{-1mm}\\
			\hspace{-1mm}0 & \Lambda \hspace{-1mm}
			\ea
		\underbrace{\ba{c}
			\frac{1}{K}\cI\tran \\
			\frac{1}{c}\cX_L
			\ea}_{\cX^{-1}},
	}
	 where $\cI = \mathds{1}_K\otimes I_M$ and $\lambda \define \|\Lambda\| = \max\{|\lambda_2(A)|, |\lambda_K(A)|\} < 1$. Note that $\cX_R$ and $\cX_L$ are different matrices from the ones defined in \eqref{cB-decom}.
	Now we define
	\eq{\label{suweid9}
		\ba{c}
		\bar{\swb}_i\\
		\check{\swb}_i
		\ea	
		\define \cX^{-1} \twb_i
	}
	and multiply $\cX^{-1}$ to both sides of \eqref{diffusion-3}, it holds that
	\eq{\label{xn238sd}
		\ba{c}
		\bar{\swb}_i\\
		\check{\swb}_i
		\ea	&= 
		\ba{cc}
		\hspace{-1mm}I_M - \frac{\mu}{K}\sum_{k=1}^K\H_{k,i-1} & - \frac{c \mu}{K}\cI\tran{\boldsymbol \cH}_{i-1}\cX_R\hspace{-1mm}\\
		\hspace{-1mm}-\frac{\mu}{c} \Lambda \cX_L {\boldsymbol \cH}_{i-1}\cI & \Lambda - \mu\Lambda \cX_L  {\boldsymbol \cH}_{i-1} \cX_R\hspace{-1mm}
		\ea	\nnb
		&\quad \times 
		\ba{c}
		\bar{\swb}_{i-1}\\
		\check{\swb}_{i-1}
		\ea \hspace{-1mm}+ \hspace{-1mm}
		\ba{c}
		\frac{\mu}{K}\cI\tran h\\
		\frac{\mu}{c}\Lambda \cX_L  h
		\ea
		\hspace{-1mm}+\hspace{-1mm}
		\ba{c}
		\hspace{-1mm}\frac{\mu}{K}\cI\tran \hspace{-1mm}\\
		\hspace{-1mm}\frac{\mu}{c}\Lambda \cX_L  \hspace{-1mm}
		\ea
		\s_i(\swb_{i-1}).
	}
	For notational simplicity, we further define
	\eq{
		\check{h} &\define \frac{1}{c} \Lambda \cX_L  h,\\ \bar{\s}_i &\define \frac{1}{K}\cI\tran \s_i(\twb_{i-1}), \\
		\check{\s}_i &\define \frac{1}{c} \Lambda \cX_L  \s_i(\twb_{i-1}),
	}
		Recalling that $h=\grad \cJ(\sw^\star)$ and, thus, $
		\cI\tran h = \sum_{k=1}^K \grad J_k(w^\star) = 0$. 	
	Therefore, recursion \eqref{xn238sd} becomes
	\eq{\label{xn238sd-2}
		\ba{c}
		\hspace{-1mm} \bar{\swb}_i \\
		\hspace{-1mm} \check{\swb}_i
		\ea
		&= 
		\ba{cc}
		\hspace{-1mm}I_M - \frac{\mu}{K}\sum_{k=1}^K\H_{k,i-1} & - \frac{c \mu}{K}\cI\tran{\boldsymbol \cH}_{i-1}\cX_R\hspace{-1mm}\\
		\hspace{-1mm}-\frac{\mu}{c} \Lambda \cX_L {\boldsymbol \cH}_{i-1}\cI & \Lambda - \mu\Lambda \cX_L  {\boldsymbol \cH}_{i-1} \cX_R\hspace{-1mm}
		\ea	\nnb
		&\quad \times 
		\ba{c}
		\bar{\swb}_{i-1}\\
		\check{\swb}_{i-1}
		\ea +
		\ba{c}
		0\\
		\mu \check{h}
		\ea
		+
		\ba{c}
		\mu \bar{\s}_i\\
		\mu \check{\s}_i
		\ea.
	}
	In the first line of the above transformed recursion, we have
	\eq{
		\bar{\swb}_i =& \big(I_M - \frac{\mu}{K}\sum_{k=1}^K\H_{k,i-1}\big) \bar{\swb}_{i-1} \nnb
		&\quad - \frac{c \mu}{K}\cI\tran{\boldsymbol \cH}_{i-1}\cX_R \check{\swb}_{i-1} + \mu \bar{\s}_i.
	}
	By following arguments in \eqref{nsh8888}--\eqref{1st-inquality-exp}, we reach
	\eq{\label{1st-inquality-exp-2}
		\bE\|\bar{\swb}_i\|^2 & \le (1-\mu \nu + 2\mu^2 \beta^2)	\bE\|\bar{\swb}_{i-1}\|^2 \nnb
		&\hspace{-10mm} + \big(\frac{c^2\delta^2\mu}{K\nu} + \frac{2c^2\beta^2\mu^2}{K}\big)\|\cX_R\|^2 \bE\|\check{\swb}_{i-1}\|^2 + \frac{\mu^2 \sigma^2}{K}.
	}
	In the second line of \eqref{xn238sd-2}, we have
	\eq{
		\check{\swb}_i =& (\Lambda - \mu \Lambda\cX_L  {\boldsymbol \cH}_{i-1} \cX_R)\check{\swb}_{i-1} \nnb
		& -\frac{\mu}{c} \Lambda \cX_L {\boldsymbol \cH}_{i-1}\cI\bar{\swb}_{i-1} + \mu \check{h} + \mu \check{\s}_i.
	}
	By following arguments similar to the ones in \eqref{b73809}--\eqref{2nd-inquality-exp}, we have
	\eq{\label{2nd-inquality-exp-2}
		&\ \bE\|\check{\swb}_i\|^2 \nnb
		\le&\ \Big( \lambda + \frac{3 \mu^2  \lambda^2(\delta^2 + K \beta^2)\|\cX_L\|^2\|\cX_R\|^2}{1-\lambda} \Big) \bE \|\check{\swb}_{i-1}\|^2 \nnb
		&\quad + \frac{3K\mu^2  \lambda^2 (\delta^2 + K \beta^2)\|\cX_L\|^2}{(1-\lambda)c^2} \bE\|\bar{\swb}_{i-1}\|^2 \nnb &\quad + \frac{3\mu^2  \lambda^2 \|\cX_L\|^2\|h\|^2}{(1-\lambda)c^2} + \frac{K\mu^2  \lambda^2}{c^2} \|\cX_L\|^2 \sigma^2
	}
	 To simplify notation, we introduce the constants
	\eq{
		e_1 = \|\cX_L\|^2,\quad e_2 = \|\cX_R\|^2, \quad b^2 = \|h\|^2/K.
	}
	Meanwhile, we also set $c^2 = e_1 K$ in \eqref{1st-inquality-exp-2} and \eqref{2nd-inquality-exp-2}. With these notations and operations, we combine \eqref{1st-inquality-exp-2} and \eqref{2nd-inquality-exp-2} to get
	\eq{\label{2hsn999}
		{\ba{c}
			\hspace{-1mm}\bE\|\bar{\swb}_i\|^2\hspace{-1mm}\\
			\hspace{-1mm}\bE\|\check{\swb}_i\|^2\hspace{-1mm}
			\ea} \le&\  
		{\ba{cc}
			1-\mu\nu+2\mu^2\beta^2 & \Big(\frac{\mu \delta^2}{\nu} + 2\mu^2 \beta^2\Big)e_1 e_2 \\
			\frac{3\mu^2  \lambda^2(\delta^2+K\beta^2)}{1-\lambda} & \lambda + \frac{3\mu^2  \lambda^2(\delta^2 + K\beta^2)e_1 e_2}{1-\lambda}
			\ea} \nnb
		& \times 
		\ba{c}
		\hspace{-1mm}\bE\|\bar{\swb}_{i-1}\|^2\hspace{-1mm}\\
		\hspace{-1mm}\bE\|\check{\swb}_{i-1}\|^2\hspace{-1mm}
		\ea  +
		\ba{c}
		\frac{1}{K}\mu^2   \sigma^2 \\
		{\mu^2  \lambda^2 \sigma^2}   + \frac{3\mu^2  \lambda^2  b^2}{1-\lambda}
		\ea.
	}
	If we choose sufficiently small  $\mu$ such that
	\eq{
		1-\mu\nu+2\mu^2\beta^2 & \le 1 - \frac{1}{2}\mu\nu, \label{238sb0-1}\\
		\Big(\frac{\mu \delta^2}{\nu} + 2\mu^2 \beta^2\Big)e_1 e_2 & \le \frac{2\mu\delta^2 e_1 e_2}{\nu}, \\
		\frac{3\mu^2  \lambda^2 (\delta^2+K\beta^2)}{1-\lambda}  & \le \frac{1}{4}  \mu  \lambda^2 \nu, \\
		\lambda + \frac{3\mu^2  \lambda^2 (\delta^2 + K \beta^2)e_1 e_2}{1-\lambda} & \le   {1+ \lambda \over 2},\label{238sb0-4}
	}
	then inequality \eqref{2hsn999} becomes 
	\eq{\label{2hsn999-3}
		{\ba{c}
			\hspace{-1mm}\bE\|\bar{\swb}_i\|^2\hspace{-1mm}\\
			\hspace{-1mm}\bE\|\check{\swb}_i\|^2\hspace{-1mm}
			\ea} \le&\  
		\underbrace{\ba{cc}
			1-\frac{\mu\nu}{2} & \frac{2\mu\delta^2 e_1 e_2}{\nu} \\
			\frac{\mu \lambda^2 \nu}{4} & {1+ \lambda \over 2}
			\ea}_{\define \cC}
		\ba{c}
		\hspace{-1mm}\bE\|\bar{\swb}_{i-1}\|^2\hspace{-1mm}\\
		\hspace{-1mm}\bE\|\check{\swb}_{i-1}\|^2\hspace{-1mm}
		\ea  \nnb
		&\quad +
		\ba{c}
		\frac{1}{K}\mu^2 \sigma^2 \\
		\mu^2  \lambda^2 \sigma^2   + \frac{3\mu^2  \lambda^2  b^2}{1-\lambda}
		\ea.
	}
To make inequalities \eqref{238sb0-1}--\eqref{238sb0-4} hold, it is enough to set 
	\eq{\label{mu_range_diffusion}
		\mu \le \frac{(1-\lambda)\nu}{(12 \hspace{-0.5mm}+\hspace{-0.5mm} 4e_1e_2 \hspace{-0.5mm}+\hspace{-0.5mm}  \lambda \sqrt{6 e_1e_2})(\delta^2 + K\beta^2)} = O\left( \frac{(1-\lambda)\nu}{\delta^2 + K\beta^2} \right).
	}
	Note that $K\beta^2 = \beta^2_{\rm max}$.	 Similar to \eqref{spectral_radius_C_ed}, it can be easily verified that when $\mu$ satisfies \eqref{mu_range_diffusion}, we have that $\rho(\cC) < 1$.	Moreover, we also have
	\eq{
		(I-\cC)^{-1}=&  
		\ba{cc}
		\frac{\mu\nu}{2} & - \frac{2\mu\delta^2 e_1 e_2}{\nu}\nnb
		-\frac{\mu \lambda^2 \nu}{4} & {1- \lambda \over 2}
		\ea^{-1} \nnb
		=&\ \frac{1}{\frac{\mu\nu(1-\lambda)}{4} - \frac{\mu^2\delta^2  \lambda^2 e_1 e_2}{2}}
		\ba{cc}
		{1- \lambda \over 2} & \frac{2\mu\delta^2 e_1 e_2}{\nu} \\
		\frac{\mu \lambda^2 \nu}{4} & \frac{\mu\nu}{2}
		\ea \nnb
		\overset{(a)}{\le}&\ \frac{8}{\mu\nu(1-\lambda)}
		\ba{cc}
		\frac{1-\lambda}{2} & \frac{2\mu\delta^2 e_1 e_2}{\nu} \\
		\frac{\mu \lambda^2 \nu}{4} & \frac{\mu\nu}{2}
		\ea \nnb
		=&\ 
		\ba{cc}
		\frac{4}{\mu\nu} & \frac{16\delta^2 e_1 e_2}{\nu^2(1-\lambda)} \\
		\frac{2 \lambda^2}{1-\lambda} & \frac{4}{1-\lambda}
		\ea.
	}
	where step (a) denotes entry-wise inequality, which holds because
	\eq{
		\frac{\mu\nu(1-\lambda)}{4} - \frac{\mu^2\delta^2 \lambda^2 e_1 e_2}{2} \ge \frac{\mu\nu(1-\lambda)}{8}
	}
	when $\mu$ satisfies \eqref{mu_range_diffusion}. By iterating \eqref{2hsn999-3}, we get
	\eq{
		&\hspace{-1cm} \limsup_{i\to\infty} 
		\ba{c}
		\hspace{-1mm}\bE\|\bar{\swb}_i\|^2\hspace{-1mm}\\
		\hspace{-1mm}\bE\|\check{\swb}_i\|^2\hspace{-1mm}
		\ea \nnb
		=&\ (I-\cC)^{-1} \ba{c}
		\frac{1}{K}\mu^2 \sigma^2 \\
		\mu^2 \sigma^2   + \frac{3\mu^2  b^2}{1-\lambda}
		\ea \nnb
		=&\	\ba{cc}
		\frac{4}{\mu\nu} & \frac{16\delta^2 e_1 e_2}{\nu^2(1-\lambda)} \\
		\frac{2 \lambda^2}{1-\lambda} & \frac{4}{1-\lambda}
		\ea \hspace{-1mm}
		\ba{c}
		\frac{1}{K}\mu^2 \sigma^2 \\
		\mu^2 \lambda^2 \sigma^2   + \frac{3\mu^2 \lambda^2  b^2}{1-\lambda}
		\ea \nnb
		=&\ \ba{c}
		\frac{4\mu\sigma^2}{K\nu} + \frac{16 \delta^2 e_1 e_2 \mu^2 \lambda^2 \sigma^2}{\nu^2(1-\lambda)} + \frac{48\delta^2 e_1 e_2 \mu^2 \lambda^2 b^2}{\nu^2 (1-\lambda)^2} \\
		\frac{2\mu^2 \lambda^2 \sigma^2}{K(1-\lambda)} + \frac{4\mu^2 \lambda^2 \sigma^2}{1-\lambda} + \frac{12\mu^2 \lambda^2 b^2}{(1-\lambda)^2}
		\ea
	}
	Therefore, 
	\eq{\label{2378sdyds}
		&\hspace{-8mm} \limsup_{i\to\infty}\bE\|\twb_i\|^2 \nnb
		\overset{\eqref{suweid9}}{\le}&\ \limsup_{i\to\infty} \big(2K\bE\|\bar{\swb}_i\|^2 + 2K e_1 e_2 \bE\|\check{\swb}_i\|^2\big) \nnb
		=&\ \frac{8\mu\sigma^2}{\nu} + \frac{(4\nu^2 + 32K\delta^2 + 8K\nu^2)e_1 e_2 \mu^2 \lambda^2 \sigma^2}{\nu^2(1-\lambda)} \nnb
		&\quad +\frac{(96\delta^2 + 24 \nu^2)Ke_1 e_2 \mu^2 \lambda^2 b^2}{\nu^2(1-\lambda)^2} \nnb
		\le&\ \frac{8\mu\sigma^2}{\nu} + \frac{44 K e_1 e_2  \delta^2 \mu^2 \lambda^2 \sigma^2}{\nu^2(1-\lambda)} +\frac{120 K e_1 e_2 \delta^2 \mu^2 \lambda^2 b^2}{\nu^2(1-\lambda)^2} \nnb
		=&\ O \left( \frac{\mu\sigma^2}{\nu} + \frac{\delta^2}{\nu^2}\cdot\frac{K\mu^2 \lambda^2 \sigma^2}{(1-\lambda)} + \frac{\delta^2}{\nu^2}\cdot \frac{K \mu^2 \lambda^2 b^2}{(1-\lambda)^2} \right).
	}
	This leads to \eqref{mu_range_diffusion-0} by dividing $K$ to both sides of \eqref{2378sdyds}.
\section{Proof of Theorem \ref{Thm2}} \label{appendix_MSD_proof}
 The derivation of the MSD expression adjusts the arguments from  \cite[Ch. 11]{sayed2014adaptation} to our case. We start by introducing 
\eq{
	\cC &\define	
	\ba{cc}
	\hspace{-1mm}I_M \hspace{-1mm}-\hspace{-1mm} \frac{\mu}{K}\sum_{k=1}^K H_k & -\frac{c\mu}{K}\cI\tran \cH \cX_{R,u}\hspace{-1mm}\\
	\hspace{-1mm}-\frac{\mu}{c} \cD_1 \cX_L \cT \cR_1 & \cD_1 - \mu \cD_1 \cX_L \cT \cX_R\hspace{-1mm}
	\ea, \label{zeta-definition}\\
	\cG &\define 
	\ba{c}
	\frac{1}{K}\cI\tran \\
	\frac{1}{c}\cD_1 \cX_L \cB_\ell
	\ea, \quad \quad \s_i \define \s_i(\swb_{i-1}),\label{calG_def} 
}
With these definitions, we can rewrite the approximate error dynamics \eqref{approximate-error-dynamics} as $\szb_i' = \cC \szb_{i-1}' + \mu \cG \s_i.$
By squaring and taking conditional expectation over the filtration $\filt_{i-1}$, we have
\eq{
	\hspace{-1mm}\bE[\|\szb_i'\|^2_\Sigma|\filt_{i-1}] = \|\szb_{i-1}'\|^2_{\cC\tran \Sigma \cC} \hspace{-1mm}+\hspace{-1mm} \mu^2 \bE[\|\s_i\|^2_{\cG\tran\Sigma \cG}|\filt_{i-1}]. \hspace{-2mm}
}
where $\Sigma$ is any positive semi-definite matrix to be decided later. By taking expectation again, we have
\eq{\label{b28d9d}
	\hspace{-1mm}\bE\|\szb_i'\|^2_\Sigma = \bE\|\szb_{i-1}'\|^2_{\cC\tran \Sigma \cC} + \mu^2 \bE\|\s_i\|^2_{\cG\tran\Sigma \cG}.
}
Now we analyze the gradient noise term. To do that,  we introduce the network noise quantity
\eq{
\cS &\define \diag\{S_1,S_2,\cdots, S_K\}. \label{calS_defini}
}
where $S_k$ is defined in \eqref{S_k_noise_limit}.  
 Note that $\mu^2 \bE\|\s_i\|^2_{\cG\tran\Sigma \cG} = \mu^2 \Tr\Big(  \Sigma \cG \bE[\s_i \s_i\tran] \cG\tran  \Big).$
By following  \cite[(11.72) -- (11.76)]{sayed2014adaptation}, it holds that $\mu^2\bE\|\s_i\|^2_{\cG\tran\Sigma \cG}$ can be well approximated by $\mu^2\Tr(\Sigma \cG \cS \cG\tran)$. To be more precise, we have
\eq{\label{xbw8}
	\limsup_{i\to \infty} \mu^2 \bE\|\s_i\|^2_{\cG\tran\Sigma \cG} = \mu^2 \Tr(\Sigma \cY) + \Tr(\Sigma)\cdot o(\mu^2),
}
where $\cY\define \cG \cS \cG\tran$ and $o(\mu^2)=O(\mu^{2+\epsilon})$ with $\epsilon>0$. By substituting \eqref{xbw8} into \eqref{b28d9d} and taking the limit, we have
\eq{\label{xn28d7d}
	\limsup_{i\to \infty}\bE\|\szb_i'\|^2_{\Sigma - \cC\tran \Sigma \cC} =&\ \mu^2 \bE\|\s_i\|^2_{\cG\tran\Sigma \cG} \nnb
	=&\ \mu^2 \Tr(\Sigma \cY) + \Tr(\Sigma)\cdot o(\mu^2).
}
Note that from \eqref{bsdkw836dg}, we are interested in $\limsup_{i\to \infty} 
\|\twb_i\|^2= \bE\|\szb_i'\|^2_{\Gamma}$. Thus, we need
\eq{
	\Sigma - \cC\tran \Sigma \cC = \Gamma. \label{lyapunoveqqq}
}
{\color{black} We now recall two block Kronecker product properties that are useful in the following  derivations \cite[Appendix F]{sayed2014adaptation}:
\begin{subequations}
\eq{
\bvec(\cA \cC \cB)&=(\cB\tran \otimes_b \cA)\bvec( \cC ) \label{kron_property1} \\
\Tr(\cA \cB)&= [\bvec(\cB\tran)]\tran \bvec(\cA) \label{kron_property2}
}
\end{subequations}
for any $\cA$, $\cB$, and $\cC$ of appropriate dimensions. 
 To solve for $\Sigma$ in \eqref{lyapunoveqqq}, we apply property \eqref{kron_property1} to both sides of \eqref{lyapunoveqqq} and reach }
\eq{
	\bvec(\Sigma) - (\cC\tran \otimes_b \cC\tran) \bvec(\Sigma) = \bvec(\Gamma),
}
where $\otimes_b$ is block Kronecker operation. Now we define $\cF = \cC\tran \otimes_b \cC\tran \in \RR^{(2K-1)^2 M^2 \times (2K-1)^2 M^2}$.
%
Since $\cC$ is stable for sufficiently small step-sizes, we know $\cF$ is also stable and hence $I-\cF$ is invertible. Therefore, it holds that
\eq{\label{xn2398}
	\bvec(\Sigma) = (I - \cF)^{-1}\bvec(\Gamma).
}
Next we evaluate the right-hand side in \eqref{xn28d7d}. From property \eqref{kron_property2}, we have
\eq{\label{h82kl}
	\mu^2 \Tr(\Sigma \cY) &= \mu^2 [\bvec(\cY\tran)]\tran \bvec(\Sigma)\nnb &\overset{\eqref{xn2398}}{=} \mu^2 [\bvec(\cY\tran)]\tran(I - \cF)^{-1} \bvec(\Gamma).
}
To examine the above quantity, we have to evaluate $(I-\cF)^{-1}$ first. We recall from \eqref{zeta-definition} that
\eq{
\hspace{-2mm}	\cC\tran = 	
	\ba{cc}
	I_M - \frac{\mu}{K}\sum_{k=1}^K H_{k}  & -\frac{\mu}{c} \cR_1\tran \cT\tran \cX_L\tran\cD_1\\
	-\frac{c\mu}{K}\cX_{R,u}\tran \cH \cI & \cD_1 - \mu \cX_R\tran \cT\tran \cX_L\tran\cD_1
	\ea.
}
With definition $\cF = \cC\tran \otimes_b \cC\tran$, we partition $\cF$ into four blocks
\eq{
	\cF = 
	\ba{cc}
	\cF_{11} & \cF_{12} \\
	\cF_{21} & \cF_{22}
	\ea	
}
where 
\eq{
	\cF_{11} = \left(I_M - \frac{\mu}{K}\sum_{k=1}^K H_{k}\right) \otimes \left(I_M - \frac{\mu}{K}\sum_{k=1}^K H_{k}\right)
}
It can be verified that 
\eq{
	\label{23ns0}
	&\ (I - \cF)^{-1} \nnb
	= &\ 
	\ba{cc}
	\hspace{-1.8mm}(I_M \otimes \frac{\mu}{K}\sum_{k=1}^K H_k \hspace{-0.8mm}+\hspace{-0.8mm} \frac{\mu}{K}\sum_{k=1}^K H_k \otimes I_M)^{-1}  & 0 \\
	\hspace{-1.8mm}0 & 0
	\ea	\hspace{-1mm}+\hspace{-1mm} O(1) \nnb
	= &\ 
	\ba{c}
	I_{M^2}\\
	0
	\ea
	Z^{-1}
	\ba{cc}
	I_{M^2} & 0
	\ea + O(1)
}
where $Z \hspace{-0.5mm}=\hspace{-0.5mm}\sum_{k=1}^K \frac{\mu}{K} \left( I_M \hspace{-0.5mm}\otimes\hspace{-0.5mm}  H_k \hspace{-0.5mm}+\hspace{-0.5mm}  H_k \hspace{-0.5mm}\otimes\hspace{-0.5mm} I_M \right)$.
With \eqref{23ns0}, we have
\eq{\label{8shd0}
	&\ (I - \cF)^{-1}\bvec(\Gamma)	\nnb
	=&\ \ba{c}
	I_{M^2}\\
	0
	\ea
	Z^{-1}
	\ba{cc}
	I_{M^2} & 0
	\ea \bvec(\Gamma) + O(1).
}
By substituting 
\eq{
	&\ \ba{cc}
	I_{M^2} & 0
	\ea	
	\bvec(\Gamma) \nnb
	=&\  
	\left(\ba{cc}
	I_M & 0 
	\ea
	\otimes_b
	\ba{cc}
	I_M & 0 
	\ea\right)
	\bvec(\Gamma) \nnb
	=&\ \bvec\left( 
	\ba{cc}
	I_M & 0 
	\ea
	\Gamma
	\ba{c}
	I_M \\
	0
	\ea
	\right) \nnb
	{=}&\ K\bvec(I_M) = K \mathrm{vec}(I_M)
}
into \eqref{8shd0}, we have
\eq{\label{8shd0-1}
	(I - \cF)^{-1}\bvec(\Gamma)	\hspace{-1mm}=\hspace{-1mm} K \ba{c}
	I_{M^2}\\
	0
	\ea Z^{-1} \mathrm{vec}(I_M) \hspace{-0.8mm}+\hspace{-0.8mm} O(1).
}
Next we let 
\eq{\label{P}
 P\define \mbox{unvec}\left(Z^{-1} \mbox{vec}(I_M)\right)=\frac{1}{2}\left(\frac{\mu}{K}\sum_{k=1}^K H_k\right)^{-1}.
}
where the last equality can be verified by following similar arguments to  \cite[Equations (11.123)--(11.129)]{sayed2014adaptation}.
Substituting \eqref{P} into \eqref{8shd0-1}, we have
\eq{\label{bsdh29876}
	(I - \cF)^{-1}\bvec(\Gamma)	=\ K \ba{c}
	I_{M^2}\\
	0
	\ea {\rm bvec}(P) + O(1).
}
Substituting \eqref{bsdh29876} into \eqref{h82kl}, we have
\eq{\label{xn238ds7gh}
	\hspace{-2mm}
	\mu^2 \Tr(\Sigma \cY) \hspace{-1mm}=\hspace{-1mm} \underbrace{\mu^2K[\bvec(\cY\tran)]\tran\hspace{-1mm}
		\ba{c}
		\hspace{-1mm}I_{M^2} \\
		\hspace{-1mm}0
		\ea
		\hspace{-1mm}
		\mbox{bvec}(P)}_{\define a} \hspace{-0.5mm} + \hspace{-0.2mm} O(\mu^2)
} 
To examine $\mu^2\Tr(\Sigma \cY)$ in the previous expression, we need to evaluate $\cY$. Since $\cY = \cG \cS \cG\tran$, we have
\eq{\label{cY}
	\cY &=
	\ba{c}
	\frac{1}{K}\cI\tran \\
	\frac{1}{c}\cX_L \cB_\ell
	\ea	
	\cS
	\ba{cc}
	\frac{1}{K}\cI &  \frac{1}{c}\cB_\ell\tran \cX_L\tran
	\ea	\nnb
	&=
	\ba{cc}
	\frac{1}{K^2}\cI\tran \cS \cI & \frac{1}{K}\cI\tran \cS \cB_\ell\tran \cX_L\tran \\
	\frac{1}{K} \cX_L \cB_\ell \cS \cI & \frac{1}{c^2}\cX_L \cB_\ell \cS \cB_\ell\tran \cX_L\tran
	\ea.
}
Note that from \eqref{calS_defini}, we have $\cI\tran \cS \cI=\sum_{k=1}^K S_k$. 
With the expression of $\cY$ in \eqref{cY}, we have
\eq{
	\label{x82300}
	a &= \mu^2K\Tr\left[\mbox{unbvec}\left\{
	\ba{c}
	I_{M^2} \\
	0
	\ea
	\mbox{bvec}(P)\right\} \cY\right]\nnb
	& \overset{(a)}{=} \mu^2K\Tr[
	\ba{c}
	I_M \\
	0
	\ea
	P
	\ba{cc}
	I_M & 0
	\ea \cY
	] \nnb
	&= \frac{\mu}{2} \Tr\left\{\left(\sum_{k=1}^K H_k\right)^{-1}\left(\sum_{k=1}^K S_k\right)\right\}.
}
where step (a) follows from property \eqref{kron_property1} and in the last step we used \eqref{P} and \eqref{cY}. With the same technique as above, we can also derive that 
\eq{\label{x82300-2}
	\Tr(\Sigma)\cdot o(\mu^2) = o(\mu).
}
Substituting \eqref{xn238ds7gh}--\eqref{x82300-2} into \eqref{xn28d7d}, we have
\eq{\label{xn28d7d-2}
	\hspace{-1mm}\limsup_{i\to \infty}\bE\|\szb_i'\|^2_{\Gamma}  
	\hspace{-0.8mm}=\hspace{-0.8mm} \frac{\mu}{2} \Tr\left\{\hspace{-1.5mm}\left(\sum_{k=1}^K H_k\hspace{-1mm}\right)^{\hspace{-1.5mm}-1}\hspace{-2mm}\left(\sum_{k=1}^K S_k\right)\hspace{-1.5mm}\right\} \hspace{-0.8mm}+\hspace{-0.8mm} o(\mu).
}
With relation \eqref{2378sdbsd} in Lemma \ref{lm-approximation-error}, we also have
\eq{\label{xn28d7d-3}
	&\hspace{-8mm}\limsup_{i\to \infty}\bE\|\szb_i\|^2_{\Gamma}\nnb
	=&\ \frac{\mu}{2} \Tr\left\{\left(\sum_{k=1}^K H_k\right)^{-1}\left(\sum_{k=1}^K S_k\right)\right\} + o(\mu).
}
Recalling the facts that $\bE\|\twb_i\|^2 = \sum_{k=1}^K \bE\|\widetilde{\w}_{k,i}\|^2$ and $\lim_{\mu \to 0}o(\mu)/\mu = 0$, we therefore derive the MSD expression of exact diffusion as follows
\eq{
	{\rm MSD} & = \mu \left(\lim_{\mu \to 0} \limsup_{i\to \infty} \frac{1}{\mu K}\bE\|\twb_i\|^2\right) \nnb
	& \overset{\eqref{bsdkw836dg}}{=} \mu \left(\lim_{\mu \to 0} \limsup_{i\to \infty} \frac{1}{\mu K}\bE\|\szb_i\|_{\Gamma}^2\right) \nnb
	& = \frac{\mu}{2 K} \Tr\left\{\left(\sum_{k=1}^K H_k\right)^{-1}\left(\sum_{k=1}^K S_k\right)\right\}.
}
				
\bibliographystyle{IEEEbib}
\bibliography{reference}

\begin{thebibliography}{10}

\bibitem{shi2015extra}
W.~Shi, Q.~Ling, G.~Wu, and W.~Yin,
\newblock ``{EXTRA}: An exact first-order algorithm for decentralized consensus
  optimization,''
\newblock {\em SIAM Journal on Optimization}, vol. 25, no. 2, pp. 944--966,
  2015.

\bibitem{lorenzo2016next}
P.~D. Lorenzo and G.~Scutari,
\newblock ``{NEXT}: {In}-network nonconvex optimization,''
\newblock {\em IEEE Transactions on Signal and Information Processing over
  Networks}, vol. 2, no. 2, pp. 120--136, 2016.

\bibitem{nedic2017achieving}
A.~Nedic, A.~Olshevsky, and W.~Shi,
\newblock ``Achieving geometric convergence for distributed optimization over
  time-varying graphs,''
\newblock {\em SIAM Journal on Optimization}, vol. 27, no. 4, pp. 2597--2633,
  2017.

\bibitem{yuan2017exact1}
K.~Yuan, B.~Ying, X.~Zhao, and A.~H. Sayed,
\newblock ``Exact dffusion for distributed optimization and learning -- {Part
  I: Algorithm development},''
\newblock {\em IEEE Transactions on Signal Processing}, vol. 67, no. 3, pp. 708
  -- 723, 2019.

\bibitem{rossi2004distributed}
L.~A. Rossi, B.~Krishnamachari, and C.~C.J. Kuo,
\newblock ``Distributed parameter estimation for monitoring diffusion phenomena
  using physical models,''
\newblock in {\em Proc. IEEE Conference on Sensor and Ad Hoc Communications and
  Networks (SECON)}, Santa Clara, CA, 2004, pp. 460--469.

\bibitem{li2002detection}
D.~Li, K.~D. Wong, Y.~Hu, and A.~M. Sayeed,
\newblock ``Detection, classification, and tracking of targets,''
\newblock {\em IEEE Signal Processing Magazine}, vol. 19, no. 2, pp. 17--29,
  2002.

\bibitem{duchi2012dual}
J.~C. Duchi, A.~Agarwal, and M.~J. Wainwright,
\newblock ``Dual averaging for distributed optimization: convergence analysis
  and network scaling,''
\newblock {\em IEEE Transactions on Automatic Control}, vol. 57, no. 3, pp.
  592--606, 2012.

\bibitem{chen2012diffusion}
J.~Chen and A.~H. Sayed,
\newblock ``Diffusion adaptation strategies for distributed optimization and
  learning over networks,''
\newblock {\em IEEE Transactions on Signal Processing}, vol. 60, no. 8, pp.
  4289--4305, 2012.

\bibitem{sayed2014adaptive}
A.~H. Sayed,
\newblock ``Adaptive networks,''
\newblock {\em Proceedings of the IEEE}, vol. 102, no. 4, pp. 460--497, April
  2014.

\bibitem{sayed2014adaptation}
A.~H. Sayed,
\newblock ``Adaptation, learning, and optimization over networks,''
\newblock {\em Foundations and Trends in Machine Learning}, vol. 7, no. 4-5,
  pp. 311--801, 2014.

\bibitem{tu2012diffusion}
S.-Y. Tu and A.~H. Sayed,
\newblock ``Diffusion strategies outperform consensus strategies for
  distributed estimation over adaptive networks,''
\newblock {\em IEEE Transactions on Signal Processing}, vol. 60, no. 12, pp.
  6217--6234, 2012.

\bibitem{nedic2009distributed}
A.~Nedi{\'c} and A.~Ozdaglar,
\newblock ``Distributed subgradient methods for multi-agent optimization,''
\newblock {\em IEEE Transactions on Automatic Control}, vol. 54, no. 1, pp.
  48--61, 2009.

\bibitem{kar2013consensus+}
S.~Kar and J.~M. Moura,
\newblock ``Consensus+ innovations distributed inference over networks:
  cooperation and sensing in networked systems,''
\newblock {\em IEEE Signal Processing Magazine}, vol. 30, no. 3, pp. 99--109,
  2013.

\bibitem{yuan2016convergence}
K.~Yuan, Q.~Ling, and W.~Yin,
\newblock ``On the convergence of decentralized gradient descent,''
\newblock {\em SIAM Journal on Optimization}, vol. 26, no. 3, pp. 1835--1854,
  2016.

\bibitem{chen2013distributed}
J.~Chen and A.~H. Sayed,
\newblock ``Distributed {Pareto} optimization via diffusion strategies,''
\newblock {\em IEEE Journal of Selected Topics in Signal Processing}, vol. 7,
  no. 2, pp. 205--220, 2013.

\bibitem{yuan2017exact2}
K.~Yuan, B.~Ying, X.~Zhao, and A.~H. Sayed,
\newblock ``Exact dffusion for distributed optimization and learning -- {Part
  II: Convergence analysis},''
\newblock {\em IEEE Transactions on Signal Processing}, vol. 67, no. 3, pp. 724
  -- 739, Feb. 2019.

\bibitem{shi2015proximal}
W.~Shi, Q.~Ling, G.~Wu, and W.~Yin,
\newblock ``A proximal gradient algorithm for decentralized composite
  optimization,''
\newblock {\em IEEE Transactions on Signal Processing}, vol. 63, no. 22, pp.
  6013--6023, 2015.

\bibitem{di2016next}
P.~Di~Lorenzo and G.~Scutari,
\newblock ``Next: In-network nonconvex optimization,''
\newblock {\em IEEE Transactions on Signal and Information Processing over
  Networks}, vol. 2, no. 2, pp. 120--136, 2016.

\bibitem{qu2018harnessing}
G.~Qu and N.~Li,
\newblock ``Harnessing smoothness to accelerate distributed optimization,''
\newblock {\em IEEE Transactions on Control of Network Systems}, vol. 5, no. 3,
  pp. 1245--1260, 2018.

\bibitem{pu2018push}
S.~Pu, W.~Shi, J.~Xu, and A.~Nedi{\'c},
\newblock ``A push-pull gradient method for distributed optimization in
  networks,''
\newblock in {\em Proc. IEEE Conference on Decision and Control (CDC)}, Miami,
  FL, 2018, IEEE, pp. 3385--3390.

\bibitem{xin2018linear}
R.~Xin and U.~A. Khan,
\newblock ``A linear algorithm for optimization over directed graphs with
  geometric convergence,''
\newblock {\em IEEE Control Systems Letters}, vol. 2, no. 3, pp. 315--320,
  2018.

\bibitem{xu2015augmented}
J.~Xu, S.~Zhu, Y.~C. Soh, and L.~Xie,
\newblock ``Augmented distributed gradient methods for multi-agent optimization
  under uncoordinated constant stepsizes,''
\newblock in {\em IEEE Conference on Decision and Control (CDC)}, Osaka, Japan,
  2015, pp. 2055--2060.

\bibitem{nedic2016geometrically}
A.~Nedi{\'c}, A.~Olshevsky, W.~Shi, and C.~A. Uribe,
\newblock ``Geometrically convergent distributed optimization with
  uncoordinated step-sizes,''
\newblock in {\em Proc. American Control Conference (ACC)}, Seattle, WA, May
  2017, pp. 3950--3955.

\bibitem{li2017decentralized}
Z.~Li, W.~Shi, and M.~Yan,
\newblock ``A decentralized proximal-gradient method with network independent
  step-sizes and separated convergence rates,''
\newblock {\em IEEE Transactions on Signal Processing}, July 2019,
\newblock early acces. Also available on arXiv:1704.07807.

\bibitem{pu2018distributed}
S.~Pu and A.~Nedi{\'c},
\newblock ``A distributed stochastic gradient tracking method,''
\newblock in {\em IEEE Conference on Decision and Control (CDC)}, Miami, FL,
  Dec. 2018, pp. 963--968.

\bibitem{xin2019distributed}
Ran Xin, Anit~Kumar Sahu, Usman~A Khan, and Soummya Kar,
\newblock ``Distributed stochastic optimization with gradient tracking over
  strongly-connected networks,''
\newblock {\em arXiv preprint:1903.07266}, March 2019.

\bibitem{tang2018d}
H.~Tang, X.~Lian, M.~Yan, C.~Zhang, and J.~Liu,
\newblock ``D2: Decentralized training over decentralized data,''
\newblock in {\em Proc. International Conference on Machine Learning (ICML)},
  Stockholm, Sweden, 2018, pp. 4848--4856.

\bibitem{towfic2015stability}
Z.~J. Towfic and A.~H. Sayed,
\newblock ``Stability and performance limits of adaptive primal-dual
  networks,''
\newblock {\em IEEE Transactions on Signal Processing}, vol. 63, no. 11, pp.
  2888--2903, 2015.

\bibitem{mao2018walkman}
X.~Mao, K.~Yuan, Y.~Hu, Y.~Gu, A.~H. Sayed, and W.~Yin,
\newblock ``Walkman: A communication-efficient random-walk algorithm for
  decentralized optimization,''
\newblock {\em Submitted for publication. {\em Also available on
  arXiv:1804.06568}}, Apr. 2018.

\bibitem{seaman2017optimal}
K.~Seaman, F.~Bach, S.~Bubeck, Y.-T. Lee, and L.~Massouli{\'e},
\newblock ``Optimal algorithms for smooth and strongly convex distributed
  optimization in networks,''
\newblock in {\em Proc. International Conference on Machine Learning (ICML)},
  Sydney, Australia, 2017, vol.~70, pp. 3027--3036.

\end{thebibliography}
\end{document}